\PassOptionsToPackage{table,dvipsnames}{xcolor}
\documentclass{mindlab}

\usepackage{fix-cm}
\usepackage[toc,header]{appendix}
\usepackage{amsmath,amssymb}
\usepackage{array}
\usepackage{enumitem}
\usepackage{float}
\usepackage{multicol}
\usepackage{needspace}
\usepackage{pgfplots}
\usepackage{tabularx}
\usepackage{tikz}
\usepackage{xspace}
\usetikzlibrary{arrows.meta,backgrounds,calc,decorations.pathreplacing,fit,matrix,positioning,shapes.geometric}
\pgfplotsset{compat=1.18}
\hypersetup{
  hidelinks,
  pdftitle={LongStraw: Long-Context RL Beyond 2M Tokens under a Fixed GPU Budget},
  pdfauthor={Changhai Zhou et al.},
  pdfsubject={Fixed-budget long-context GRPO execution}
}

\newcolumntype{C}[1]{>{\centering\arraybackslash}m{#1}}

\newcolumntype{Y}{>{\centering\arraybackslash}X}

\makeatletter
\newcommand{\compacttoc}{\@starttoc{toc}}
\makeatother

\newcommand{\longstraw}{LongStraw\xspace}
\newcommand{\qwenmodel}{Qwen3.6-27B\xspace}
\newcommand{\glmmodel}{GLM-5.2\xspace}

\newcommand{\correspondencemark}{\ast}
\DeclareRobustCommand{\authorcontactlist}{%
  \begin{tabular}[t]{@{}l@{\;}l@{}}
    Changhai Zhou: & \email{chzhou25@m.fudan.edu.cn} \\
    Andrew Chen: & \email{andrew@mindlab.ltd} \\
    Pony Ma: & \email{pony@mindlab.ltd}
  \end{tabular}}

\title{LongStraw: Long-Context RL Beyond 2M Tokens under a Fixed GPU Budget}
\author[1,2]{Changhai Zhou}
\author[1]{Kieran Liu}
\author[1]{Yuhua Zhou}
\author[1]{Qian Qiao}
\author[1]{Jun Gao}
\author[1]{Harry Zhang}
\author[1]{Irvine Lu}
\author[1]{Nolan Ho}
\author[1]{Lucian Li}
\author[1]{Andrew Lei}
\author[1]{Cleon Cheng}
\author[1]{Steven Chiang}
\author[1]{Yihang Zeng}
\author[1]{Di Zhang}
\author[1]{Rio Yang}
\author[1]{Kaijie Chen}
\author[1,\correspondencemark]{Andrew Chen}
\author[1,\correspondencemark]{Pony Ma}
\author[2]{Weizhong Zhang}
\author[2]{Cheng Jin}
\affiliation[1]{MindLab}
\affiliation[2]{Fudan University}
\correspondence{\authorcontactlist}
\code{\url{https://github.com/MindLab-Research/longstraw}}
\date{July 2026}

\abstract{
\beginabstract\small
Long-context RL post-training is constrained by the lifetime of state and
gradients, not attention cost alone. In GRPO, one multi-million-token prompt
must serve old-policy and reference scoring plus multiple policy responses,
while conventional autograd keeps the prompt graph and all response graphs live
alongside model weights, caches, and distributed communication buffers. We
present \textbf{LongStraw}, an objective-aware, architecture-aware system for
resident-state virtualization, response replay, and distributed-gradient
execution. Its transaction captures the shared prompt without autograd, retains
only the architecture-required state on explicitly owned pages, restores that
state for each group member, scores old/reference branches without a graph,
replays one policy response at a time with autograd, and accumulates the
resulting gradients before one distributed finalization and optimizer step.
This schedule bounds the live training graph by the response suffix while
reusing the expensive prompt computation across the complete GRPO group.

We instantiate this design for two incompatible model structures. Qwen3.6-27B
combines 48 recurrent GDN layers with 16 full-attention layers; LongStraw keeps
the compact recurrent state and physically CP8-sharded KV pages, composes
global attention through cross-rank LSE/output merging, and performs blockwise
response replay. GLM-5.2 combines a 78-layer MLA/DSA attention stack with a
256-expert, top-8 MoE tail. Its implementation keeps CP-sharded MLA latent pages
and DSA indexer-key pages in CPU memory, stages one layer at a time,
reconstructs IndexShare-aware global sparse selection over CP32, and dispatches
routed response tokens over EP32. The two paths share one transaction contract
while specializing the retained state, replay operator, and collective
communication to the architecture.

On eight H20 GPUs, Qwen completes exact-attention response-only steps at
2,097,152 positions for \(G=2\) and \(G=8\); increasing the group size from 2
to 8 adds only 0.208 GB of peak allocated memory. At 4,456,448 positions, one
resident prefix supports eight consecutive \(G=8\) optimizer cycles,
comprising 64 response replays at 83.894 GB per rank. On 32 H20 GPUs, GLM-5.2
completes the full exact-2M online GRPO workflow end to end. A Tinker-managed
vLLM TP8/PP4 sampler loads the policy-LoRA adapter projection, generates two
responses for a real DAPO-MATH prompt, and obtains rewards \([-1,+1]\); the same
devices then transition to Megatron TP1/CP32/EP32 training under the shared
position configuration. The training phase composes global cross-CP DSA,
executes two 78-layer backward passes through the MLA/DSA and MoE stack,
finalizes distributed gradients on all 32 ranks, and applies one optimizer
step. These results establish complete million-token RL post-training under
fixed GPU budgets and show that practical context capacity is governed by
resident-state lifetime, response replay, topology handoff, and distributed
ownership rather than attention kernels alone.
}

\begin{document}
\sbox{\mindlablogobox}{%
  \includegraphics[height=12mm]{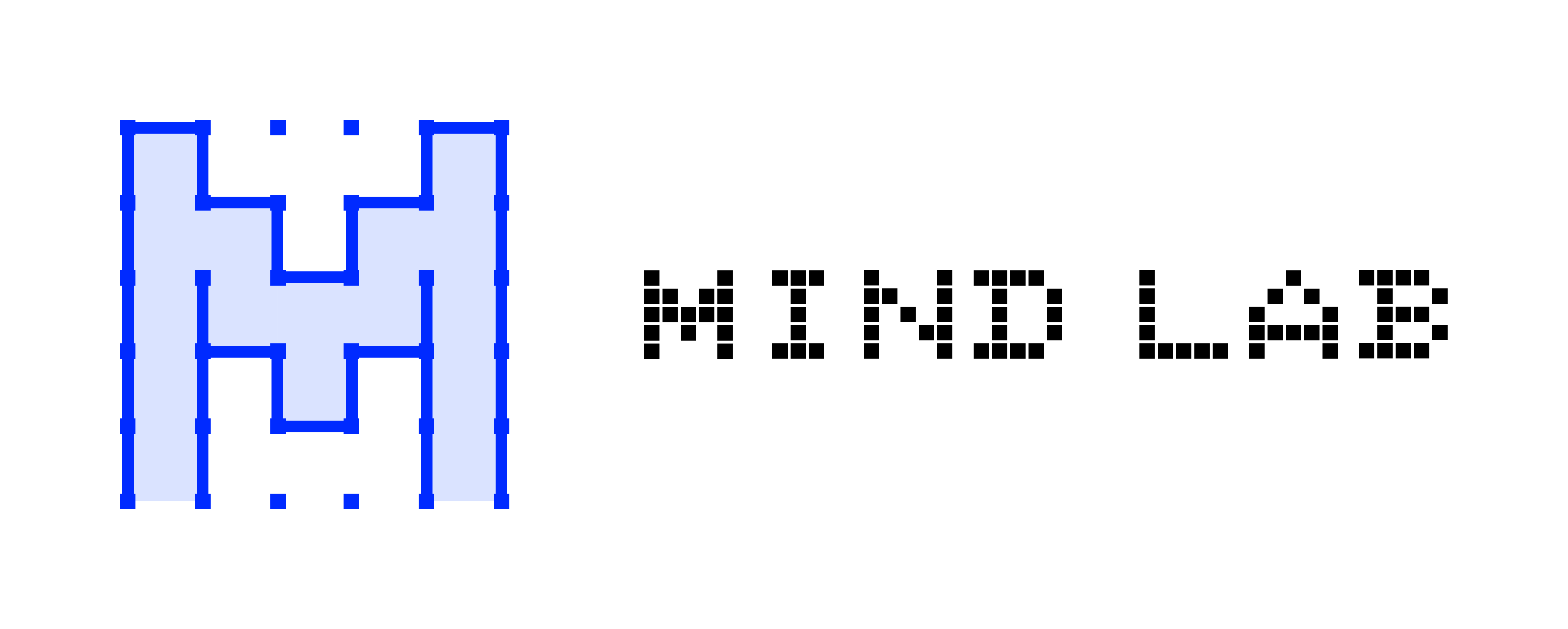}%
}
\maketitle

\newpage
\section*{Contents}
\begingroup
\footnotesize
\setstretch{1.06}
\setlength{\parskip}{0pt}
\setlength{\columnsep}{2em}
\raggedright
\begin{multicols}{2}
\compacttoc
\end{multicols}
\endgroup
\newpage

\section{Introduction}
\label{sec:introduction}

AI agents are moving beyond one-shot answers toward using tools, inspecting
code and documents, and acting over long trajectories. ReAct formalizes this
interaction as a sequence of reasoning, actions, and observations
~\citep{yao2023react}, while LongCat-Flash-Thinking trains agents over long,
multi-turn tool trajectories~\citep{longcatthinking2601}. For these agents,
context carries supporting information, environment observations, tool outputs, and earlier
decisions into the next action.

Long-context inference and post-training use memory differently. An inference
server can prefill a prompt, cache the state used for decoding, and discard the
forward graph~\citep{kwon2023efficient}. Post-training must score several
responses and backpropagate through them. Group Relative Policy Optimization
(GRPO) compares responses that share a prompt through group-relative advantages
~\citep{shao2024deepseekmath}. Each response may be short, but its score still
depends on the full prompt and its cached state.

Existing techniques reduce important parts of this cost, but they do not by
themselves make a fixed-GPU GRPO run fit. Memory-efficient attention reduces the
workspace of an attention layer~\citep{rabe2021self}. FlashAttention improves
the data movement of exact attention~\citep{dao2022flashattention}. LoRA reduces
the number of trainable parameters~\citep{hu2021lora}, while QLoRA also reduces
the storage cost of the base model~\citep{dettmers2023qlora}. The prompt graph,
response graphs, cached state, and distributed communication still compete for
the same device memory.

Large accelerator fabrics can extend sequence length by distributing this work
more widely. Ring Attention reports 4.096M-position training for a 7B model on
32 A100 GPUs~\citep{liu2023ring}. DeepSpeed-Ulysses studies one-million-token
training while scaling to 256 A100 GPUs~\citep{jacobs2023ulysses}. ByteScale
reports a 2M LLaMA-7B case on 1,024 GPUs~\citep{ge2025bytescale}, and USP
combines ring and all-to-all sequence parallelism~\citep{fang2024usp}. These
systems establish the scale-out route. A complementary fixed-budget lineage
includes ZeRO-style partitioning and offload, activation recomputation, and
OOMB's chunk-recurrent full-sequence causal-LM training with paged, offloaded
KV state~\citep{li2026out}. LongStraw addresses a different problem structure:
multiple RL responses share one long prompt, so prompt-state lifetime, role
replay, group accumulation, and cache refresh must be coordinated as one
transaction. The Qwen prototype uses selected OOMB \texttt{chunkoptim} cache
and paged-attention kernels as implementation components; LongStraw is not an
algorithmic extension of OOMB.

\longstraw answers this question with an objective-aware, architecture-aware
resident-state virtualization, response-replay, and distributed-gradient
execution system. It evaluates the shared prompt once without automatic
differentiation, maps only the information needed to condition later tokens to
owned physical pages, and then processes one response at a time. Gradients from
the response members are accumulated before one distributed finalization and
optimizer call. Serial replay increases elapsed time, but it avoids keeping the
prompt graph and every response graph live at the same time.

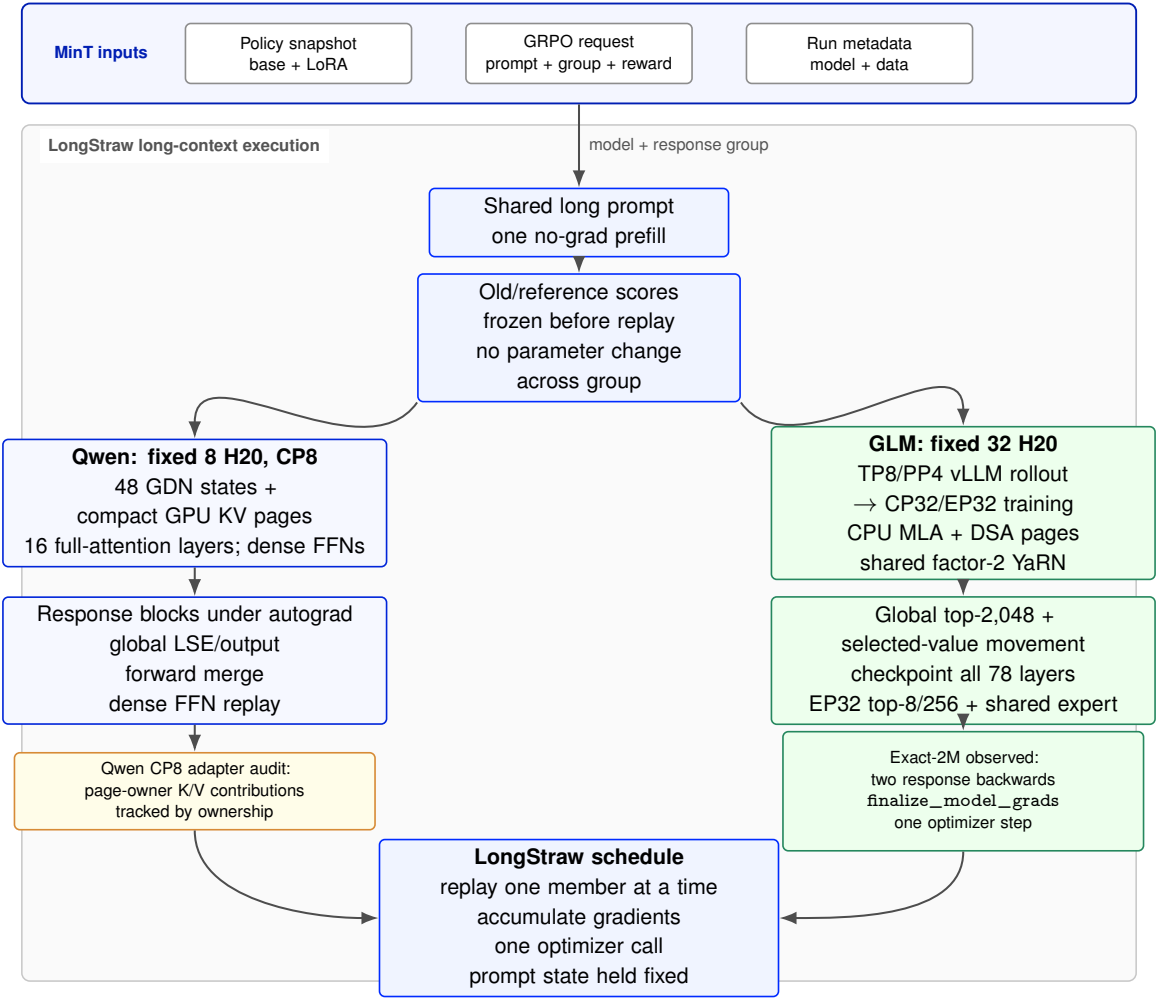
\begin{figure}[!t]
\centering
\resizebox{0.92\textwidth}{!}{%
\begin{tikzpicture}[
    font=\sffamily\small,
    box/.style={draw=mindlabfg!65, rounded corners=2pt, line width=0.6pt,
        fill=white, minimum height=0.82cm, align=center, inner xsep=7pt},
    shared/.style={box, draw=mindlabblue, fill=mindlabbluepale!45},
    qwen/.style={box, draw=mindlabblue!85!black, fill=mindlabbluepale!28,
        text width=4.5cm, minimum height=1.55cm},
    glm/.style={box, draw=ForestGreen!75!black, fill=green!7,
        text width=4.5cm, minimum height=1.55cm},
    warn/.style={box, draw=BurntOrange!85!black, fill=yellow!10,
        text width=4.2cm, font=\sffamily\scriptsize},
    control/.style={box, draw=mindlabfg!48, fill=white,
        text width=2.45cm, minimum height=0.78cm,
        font=\sffamily\scriptsize},
    flow/.style={-Latex, line width=0.75pt, draw=mindlabfg!75},
    local/.style={-Latex, line width=0.7pt, dashed, draw=BurntOrange!85!black}
]
\begin{scope}[on background layer]
  \filldraw[rounded corners=3pt, fill=mindlabfg!2, draw=mindlabfg!24,
    line width=0.55pt] (-7.25,1.42) rectangle (7.25,-9.72);
\end{scope}

\node[draw=mindlabblue!70!black, rounded corners=3pt, line width=0.7pt,
  fill=mindlabbluepale!34, minimum width=14.5cm, minimum height=1.30cm]
  (mint) at (0,2.35) {};
\node[anchor=west, align=left, font=\sffamily\bfseries\scriptsize,
  text=mindlabblue!70!black] at (-6.95,2.35)
  {MinT inputs};
\node[control] (revision) at (-3.65,2.35)
  {Policy snapshot\\base + LoRA};
\node[control] (request) at (0,2.35)
  {GRPO request\\prompt + group + reward};
\node[control] (lifecycle) at (3.65,2.35)
  {Run metadata\\model + data};

\node[anchor=west, fill=white, inner xsep=3pt,
  font=\sffamily\bfseries\scriptsize, text=mindlabfg!72]
  at (-7.02,1.14) {LongStraw long-context execution};
\node[shared, text width=3.4cm] (prompt) at (0,0.15) {Shared long prompt\\one no-grad prefill};
\node[shared, text width=3.7cm] (freeze) at (0,-1.35)
  {Old/reference scores\\frozen before replay\\no parameter change\\across group};

\node[qwen] (qstate) at (-5.0,-3.50)
  {\textbf{Qwen: fixed 8 H20, CP8}\\48 GDN states +\\
   \mbox{compact GPU KV pages}\\16 full-attention layers; dense FFNs};
\node[glm] (gstate) at (5.0,-3.50)
  {\textbf{GLM: fixed 32 H20}\\TP8/PP4 vLLM rollout \(\rightarrow\) CP32/EP32 training\\
   CPU MLA + DSA pages\\shared factor-2 YaRN};

\node[qwen] (qreplay) at (-5.0,-5.55)
  {Response blocks under autograd\\global \mbox{LSE/output}\\
   forward merge\\dense FFN replay};
\node[glm] (greplay) at (5.0,-5.55) {Global top-2,048 + selected-value movement\\checkpoint all 78 layers\\EP32 top-8/256 + shared expert};

\node[warn] (qbound) at (-5.0,-7.25) {Qwen CP8 adapter audit:\\page-owner K/V contributions\\tracked by ownership};
\node[glm, text width=4.2cm, font=\sffamily\scriptsize] (gbound) at (5.0,-7.25)
  {Exact-2M observed:\\two response backwards\\\texttt{finalize\_model\_grads}\\one optimizer step};

\node[shared, text width=4.7cm] (step) at (0,-8.90)
  {\textbf{LongStraw schedule}\\replay one member at a time\\
   accumulate gradients\\one optimizer call\\prompt state held fixed};

\draw[flow] (mint.south) -- node[right, font=\sffamily\scriptsize,
  align=left, text=mindlabfg!72]
  {model + response group} (prompt.north);
\draw[flow] (prompt) -- (freeze);
\draw[flow] (freeze.south west)
  to[out=235,in=90,looseness=1.05] (qstate.north);
\draw[flow] (freeze.south east)
  to[out=305,in=90,looseness=1.05] (gstate.north);
\draw[flow] (qstate) -- (qreplay);
\draw[flow] (gstate) -- (greplay);
\draw[flow] (qreplay) -- (qbound);
\draw[flow] (greplay) -- (gbound);
\draw[flow] (qbound.south)
  to[out=270,in=180,looseness=0.88] (step.west);
\draw[flow] (gbound.south)
  to[out=270,in=0,looseness=0.88] (step.east);
\end{tikzpicture}%
}
\caption{\textbf{LongStraw execution path.} The runtime supplies a model snapshot
and coordinates the response group. LongStraw evaluates the shared prompt once,
stores model-specific state, then replays one response at a time and accumulates
gradients. Qwen keeps recurrent state and sharded KV pages; GLM uses a
Tinker-managed vLLM rollout, keeps CPU MLA/DSA pages, time-multiplexes rollout
and training, composes global DSA, and finalizes gradients before its exact-2M
step. The orange Qwen box marks the separate CP8 replica-finalization audit.}
\label{fig:system-path}
\end{figure}

The stored information follows the model architecture. Qwen3.6-27B combines
recurrent layers with full-attention layers~\citep{qwen36config}; LongStraw keeps
the recurrent state and context-sharded key/value pages needed by later tokens.
GLM-5.2 uses compressed attention state, sparse attention indices, and routed
experts~\citep{glm52config}. Its implementation moves the prompt state to CPU
memory and stages one decoder layer at a time. Figure~\ref{fig:system-path}
shows the shared execution schedule and where the two implementations differ.

LongStraw runs inside the training side of MinT~\citep{mindlab2026mint}. MinT
manages model workers, adapter revisions, and the outer policy transaction;
LongStraw manages the prompt state and response work inside one long-context
transaction. The two systems therefore operate at different levels of the same
training stack.

This execution boundary is motivated by efficiency. Within one GRPO update,
all \(G\) members reuse the same captured prefix before any parameter change.
Response-only replay retains gradients through the response computation while
omitting the prompt-side vector--Jacobian product. Qwen completes eight \(G=8\)
cycles at 4.25M. GLM now completes the full exact-2M online transaction on
32 H20 GPUs: a Tinker-managed vLLM rollout, reward computation, global DSA,
two 78-layer response backwards, distributed gradient finalization, and one
optimizer step. The CP8 page-owner K/V-adapter synchronization is reported as
a companion ownership audit alongside these completed execution receipts.

\paragraph{Contributions.}
This report makes three contributions:
\begin{itemize}
  \item It defines objective-aware resident-state virtualization: state that
  survives the prompt boundary, response work that is recomputed, and
  rank-owned gradient contributions are handled as distinct parts of one GRPO
  step.
  \item It implements that design for two different model structures: a Qwen
  hybrid recurrent/attention stack and a GLM compressed-attention/MoE stack,
  with architecture-specific pages, replay, routing, and ownership.
  \item It reports exact 2M (G=2/G=8) Qwen response-only steps, 4.25M (G=8)
  eight-step prefix reuse, the full exact-2M GLM online GRPO workflow, and its
  vLLM-to-Megatron topology handoff.
\end{itemize}

\FloatBarrier

\subsection{Report Roadmap}

The report first defines the training dependency and model structures, then
describes the Qwen and GLM implementations. The remaining sections present the
fixed-budget measurements, explain the main memory and communication costs, and
state the evaluation scope of the execution paths.

\section{GRPO Training Dependency Graph}
\label{tr:workload}

The systems problem begins with the update graph, not with a particular
attention kernel. Let a prompt contain \(P\) tokens, group member \(i\) contain
\(R_i\) scored response tokens, and the group contain \(G\) responses. For old
policy \(\pi_{\mathrm{old}}\), current policy \(\pi_\theta\), and normalized
advantage \(A_i\), define
\begin{equation}
\rho_{i,t}(\theta)=
\exp\!\left(
  \log\pi_\theta(y_{i,t}\mid x_{1:P},y_{i,<t})-
  \log\pi_{\mathrm{old}}(y_{i,t}\mid x_{1:P},y_{i,<t})
\right).
\label{eq:importance-ratio}
\end{equation}
The clipped policy term is
\begin{equation}
\mathcal{L}_{\mathrm{policy}} =
-\frac{1}{G}\sum_{i=1}^{G}\frac{1}{R_i}\sum_{t=1}^{R_i}
\min\!\left(
  \rho_{i,t}A_i,
  \operatorname{clip}(\rho_{i,t},1-\epsilon,1+\epsilon)A_i
\right),
\label{eq:grpo}
\end{equation}
with a tokenwise reference-policy KL term weighted by \(\beta\). The clipped
ratio surrogate follows PPO~\citep{schulman2017ppo}; group-relative advantages,
member normalization, and the reference-policy penalty follow the GRPO
objective~\citep{shao2024deepseekmath}. This is the member-normalized GRPO term
implemented by the evaluated paths. The
contribution of this report is how its conditional log-probabilities are
executed when the context length exceeds two million positions under a fixed
accelerator allocation without adding devices.

\subsection{Branches and Ordering Constraints}

A grouped update has five logically different phases.

\begin{enumerate}[leftmargin=2.0em,label=\textbf{\arabic*.}]
  \item \textbf{Prompt capture.} Evaluate the shared prompt without autograd
  and retain the architecture-specific conditional state.
  \item \textbf{Pre-step scoring.} Evaluate old-policy and reference-policy
  response log-probabilities before the corresponding policy backward.
  Parameters remain fixed through the group. GLM materializes both old score
  sets first; Qwen performs old/reference scoring and policy replay member by
  member.
  \item \textbf{Advantage construction.} Convert group rewards into the
  advantages used by Equation~\ref{eq:grpo}.
  \item \textbf{Policy replay.} Rebuild one short response graph at a time,
  backpropagate its loss, and accumulate gradients into the same adapter.
  \item \textbf{Optimizer transaction.} Synchronize accumulated gradients,
  step once after all \(G\) members, and clear gradients.
\end{enumerate}

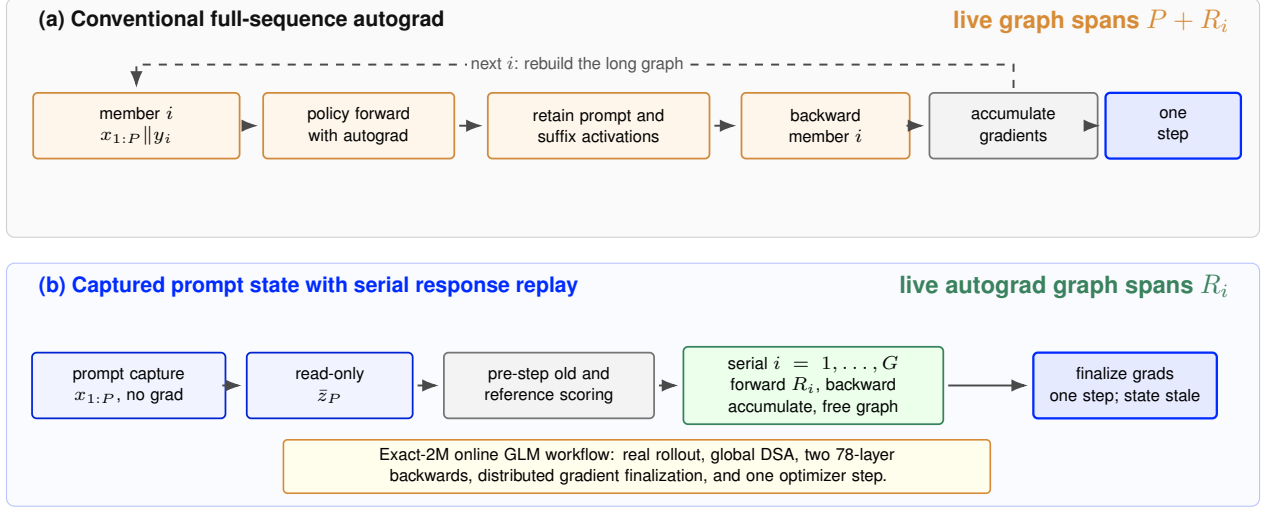
\begin{figure}[t]
\centering
\resizebox{\textwidth}{!}{%
\begin{tikzpicture}[
  font=\sffamily\scriptsize,
  node distance=0.38cm,
  block/.style={draw=mindlabfg!62, rounded corners=1.7pt, line width=0.65pt,
    fill=white, minimum height=0.86cm, align=center, inner xsep=5pt},
  legacy/.style={block, draw=BurntOrange!82!black, fill=BurntOrange!7},
  capture/.style={block, draw=mindlabblue!82!black, fill=mindlabbluepale!42},
  score/.style={block, draw=mindlabfg!58, fill=mindlabfg!5},
  replay/.style={block, draw=ForestGreen!70!black, fill=green!8},
  optstep/.style={block, draw=mindlabblue, fill=mindlabblue!10, line width=0.85pt},
  flow/.style={-Latex, draw=mindlabfg!76, line width=0.72pt, shorten >=1.5pt,
    shorten <=1.5pt},
  warn/.style={draw=BurntOrange!85!black, fill=yellow!9, rounded corners=1.5pt,
    line width=0.6pt, align=center, inner xsep=5pt, inner ysep=3pt}
]

\filldraw[rounded corners=3pt, fill=mindlabfg!2, draw=mindlabfg!20]
  (-0.25,3.42) rectangle (16.15,6.55);
\filldraw[rounded corners=3pt, fill=mindlabbluepale!15, draw=mindlabblue!28]
  (-0.25,-0.10) rectangle (16.15,3.08);

\node[anchor=west, font=\sffamily\bfseries\small, text=mindlabfg]
  at (0.05,6.25) {(a) Conventional full-sequence autograd};
\node[anchor=east, font=\sffamily\bfseries, text=BurntOrange!85!black]
  at (15.88,6.25) {live graph spans \(P+R_i\)};

\node[legacy, text width=2.35cm] (seq) at (1.45,4.88)
  {member \(i\)\\\(x_{1:P}\Vert y_i\)};
\node[legacy, text width=2.15cm] (fullfwd) at (4.36,4.88)
  {policy forward\\with autograd};
\node[legacy, text width=2.55cm] (retain) at (7.50,4.88)
  {retain prompt and\\suffix activations};
\node[legacy, text width=1.85cm] (fullbwd) at (10.47,4.88)
  {backward\\member \(i\)};
\node[score, text width=1.85cm] (fullacc) at (12.93,4.88)
  {accumulate\\gradients};
\node[optstep, text width=1.42cm] (fullstep) at (15.02,4.88)
  {one\\step};

\draw[flow] (seq) -- (fullfwd);
\draw[flow] (fullfwd) -- (retain);
\draw[flow] (retain) -- (fullbwd);
\draw[flow] (fullbwd) -- (fullacc);
\draw[flow] (fullacc) -- (fullstep);
\draw[flow, dashed] (fullacc.north) -- (12.93,5.68) -- (1.45,5.68) -- (seq.north);
\node[font=\sffamily\scriptsize, text=mindlabfg!72,
  fill=mindlabfg!2, inner sep=1.3pt]
  at (7.22,5.68) {next \(i\): rebuild the long graph};

\node[anchor=west, font=\sffamily\bfseries\small, text=mindlabblue]
  at (0.05,2.78) {(b) Captured prompt state with serial response replay};
\node[anchor=east, font=\sffamily\bfseries, text=ForestGreen!68!black]
  at (15.88,2.78) {live autograd graph spans \(R_i\)};

\node[capture, text width=2.18cm] (prefill) at (1.35,1.48)
  {prompt capture\\\(x_{1:P}\), no grad};
\node[capture, text width=1.82cm] (state) at (3.98,1.48)
  {read-only\\\(\bar z_P\)};
\node[score, text width=2.40cm] (freeze) at (6.85,1.48)
  {pre-step old and\\reference scoring};
\node[replay, text width=3.05cm] (serial) at (10.31,1.48)
  {serial \(i=1,\ldots,G\)\\forward \(R_i\), backward\\accumulate, free graph};
\node[optstep, text width=2.05cm] (onestep) at (14.39,1.48)
  {finalize grads\\one step; state stale};

\draw[flow] (prefill) -- (state);
\draw[flow] (state) -- (freeze);
\draw[flow] (freeze) -- (serial);
\draw[flow] (serial) -- (onestep);
\node[warn, text width=8.9cm, anchor=north] at (8.00,0.78)
  {Exact-2M online GLM workflow: real rollout, global DSA, two 78-layer
   backwards, distributed gradient finalization, and one optimizer step.};

\end{tikzpicture}%
}
\caption{\textbf{Changing the graph boundary, not the GRPO objective.}
(a) Conventional autograd retains prompt-dependent activations in each member
graph. (b) The reported schedule captures a read-only prompt state, computes
old/reference scores under unchanged pre-step parameters, and replays and frees one response graph at a time,
accumulates \(G\) local gradients, finalizes them across distributed ownership,
and applies one optimizer step. Serial
replay bounds live policy-autograd activations by one response, although
group-indexed inputs and frozen scores still grow with the supplied group.}
\label{fig:training-graph}
\end{figure}

The ordering is load-bearing. Stepping between group members changes both the
importance ratio and the prompt state. Holding all response graphs makes
activation memory scale with \(G\), whereas serial replay makes group
cardinality primarily a scheduling and time dimension. This is a systems
statement, measured here as the scheduling behavior; \(G\) still defines
reward normalization and the GRPO advantage set. Figure~\ref{fig:training-graph}
contrasts the full-sequence and captured-state graphs.

The Qwen million-token measurements use supplied responses and rewards to
isolate the long-context training graph, with a separate 192-token online
canary. The GLM measurement instead closes the complete online chain at exact
2M: policy-LoRA sampling, DAPO reward, old log-probabilities, two response-only
backwards, distributed gradient finalization, and one optimizer step.

\subsection{Conditional Replay as an Efficiency Boundary}

Let \(z_P(\theta)\) denote the complete model state after processing the prompt.
A conventional full-sequence loss differentiates both its explicit response
computation and the parameter dependence of the prompt state:
\begin{equation}
\nabla_\theta \ell(\theta,z_P(\theta)) =
\left.\frac{\partial\ell}{\partial\theta}\right|_{z_P}+
\frac{\partial\ell}{\partial z_P}
\frac{\partial z_P}{\partial\theta}.
\label{eq:gradient-boundary}
\end{equation}
LongStraw stores
\(\bar z_P=\operatorname{stopgrad}(z_P(\theta))\) and computes the first term.
This retains the direct gradient of every scored response token and removes the
long prompt backward graph, which is the central memory saving. The omitted
second term is the prompt-side vector--Jacobian product. Exact response
attention preserves the conditional response operator while avoiding the
prompt graph's dominant memory lifetime.

State validity has two timescales. Within update \(k\), every group member sees
the same \(\theta_k\), so sharing one captured prefix is exact for the stated
response-only computation. After the optimizer produces \(\theta_{k+1}\), an
exact replay recaptures \(z_P(\theta_{k+1})\); an explicit multi-step
prefix-reuse mode can instead retain the resident boundary across optimizer
cycles. The 4.25M run demonstrates eight consecutive reuse-and-update cycles,
and the report keeps this efficiency mode separate from the exact within-update
response-only contract.

\subsection{Three Levels of Validation}

We use three tests because a correct distributed forward may still leave
replicated gradients rank-local.

\paragraph{Execution capacity.}
Did every requested score, backward, collective, and optimizer event complete
on every rank with finite values? This is established by per-rank logs and
rank-local traces.

\paragraph{Response-operator fidelity.}
Does distributed replay compute the model-defined conditional response
operator? Qwen reaches this level for full-attention layers through a global
CP8 merge with BF16 numerator reduction. The current exact-2M GLM run exchanges
candidate positions and selected MLA values across CP32 and composes the global
DSA output before the MoE tail.

\paragraph{Distributed-update consistency.}
Are all sharded gradient contributions reduced to the correct parameter owner,
and do replicated adapters remain identical after the optimizer call? The Qwen
implementation completes scoring, response backward, and optimizer application
for the evaluated response-only procedure. In the CP8 prototype, custom
backward all-reduces
\(dQ\) but leaves page-owner \(dK/dV\) contributions to replicated projection
adapters local. The eight AdamW instances therefore establish local optimizer
application, not cross-rank replica equivalence. This is a distributed
optimizer-finalization audit, separate from the completed response-gradient path.
The current GLM exact-2M run calls
\texttt{finalize\_model\_grads} after both backwards and before one optimizer
step on every rank, closing the distributed update within the complete exact-2M
workflow.

We count an execution run as complete when old scores are frozen, every group
member produces a live backward, local gradients accumulate, each worker issues
exactly one optimizer call, values remain finite, and all ranks terminate. This
defines the execution-completeness criterion used in these receipts; the
ownership audits above report the numerical composition fields separately.

\section{Architecture Anatomy and Bottleneck Sources}
\label{tr:architecture}

The two models differ along two independent axes. The feed-forward axis is
dense versus MoE. It determines parameter residency, token routing, and the
shape of activation buffers. The token-mixing axis is GDN/full attention versus
MLA/DSA; it determines retained prompt state and response-time collectives.
Model-level labels such as ``dense'', ``MoE'', or ``sparse'' hide this
separation.

\begin{table}[!htbp]
\centering
\caption{The two model axes and the state that crosses the detached prompt
boundary. Dense/MoE describes the FFN; full/GDN versus MLA/DSA describes
attention.}
\label{tab:model-contracts}
\footnotesize
\setlength{\tabcolsep}{3pt}
\begin{tabular}{@{}C{0.13\textwidth}C{0.17\textwidth}C{0.22\textwidth}C{0.22\textwidth}C{0.18\textwidth}@{}}
\toprule
Model & FFN topology & Attention topology & Durable prompt state & Response-time collectives \\
\midrule
\rowcolor{mindlabbluepale!22}
Qwen3.6-27B & 64 dense gated FFNs & 48 GDN + 16 full-attention layers & GPU GDN state + compact CP-sharded KV pages & Global CP8 forward merge; response-only objective complete; K/V-adapter synchronization is tracked in a companion audit \\
\rowcolor{green!7}
GLM-5.2 & 3 dense + 75 MoE; 256 routed, top-8 + 1 shared & 78 MLA/DSA; 21 index + 57 IndexShare layers & CPU CP-sharded MLA latent pages + index-layer DSA key pages & Global CP32 candidate/value composition; EP32 dispatch/combine; gradient finalization before one exact-2M step \\
\bottomrule
\end{tabular}
\end{table}

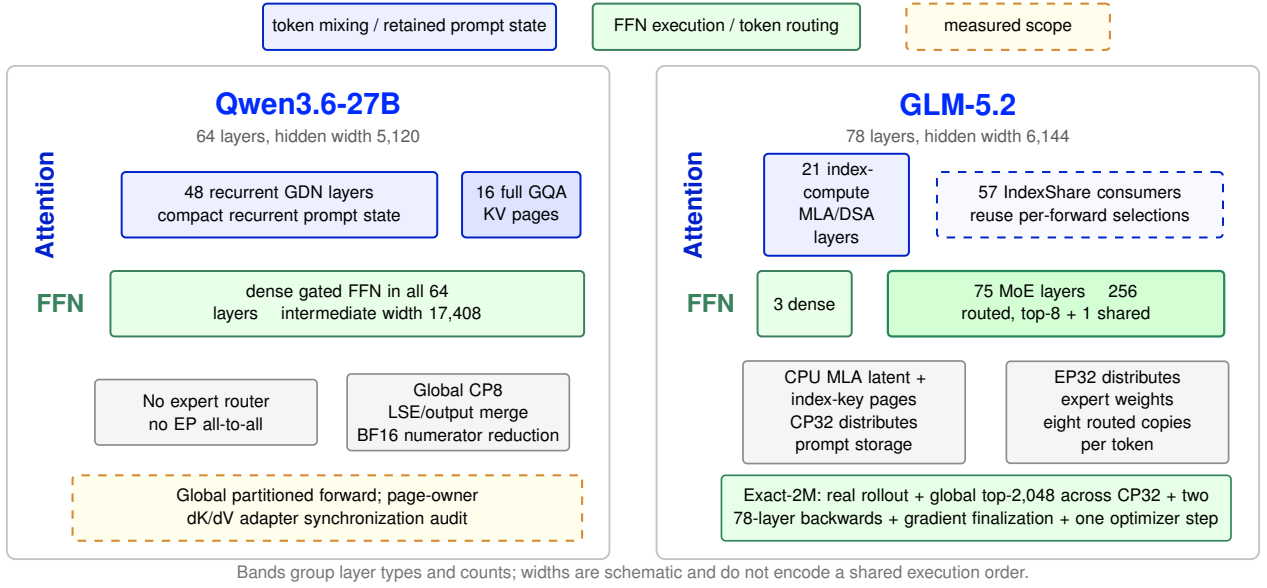
\begin{figure}[t]
\centering
\resizebox{\textwidth}{!}{%
\begin{tikzpicture}[
  font=\sffamily\scriptsize,
  panel/.style={draw=mindlabfg!25, rounded corners=3pt, fill=white,
    line width=0.65pt},
  attn/.style={draw=mindlabblue!78!black, fill=mindlabbluepale!58,
    rounded corners=1.5pt, line width=0.7pt, align=center, minimum height=0.82cm},
  attnshare/.style={attn, fill=mindlabbluepale!24, dashed},
  dense/.style={draw=ForestGreen!67!black, fill=green!9, rounded corners=1.5pt,
    line width=0.7pt, align=center, minimum height=0.82cm},
  moe/.style={draw=ForestGreen!77!black, fill=green!17, rounded corners=1.5pt,
    line width=0.8pt, align=center, minimum height=0.82cm},
  state/.style={draw=mindlabfg!48, fill=mindlabfg!4, rounded corners=1.5pt,
    line width=0.6pt, align=center, minimum height=0.82cm},
  boundary/.style={draw=BurntOrange!85!black, fill=yellow!9, rounded corners=1.5pt,
    dashed, line width=0.75pt, align=center, minimum height=0.82cm}
]

\node[attn, minimum width=3.65cm, minimum height=0.58cm] at (5.15,7.15)
  {token mixing / retained prompt state};
\node[dense, minimum width=3.35cm, minimum height=0.58cm] at (9.12,7.15)
  {FFN execution / token routing};
\node[boundary, minimum width=2.55cm, minimum height=0.58cm] at (12.65,7.15)
  {measured scope};

\node[panel, minimum width=7.55cm, minimum height=6.18cm] (qpanel) at (3.88,3.57) {};
\node[panel, minimum width=7.55cm, minimum height=6.18cm] (gpanel) at (12.02,3.57) {};

\node[font=\sffamily\bfseries\large, text=mindlabblue] at (3.88,6.18)
  {Qwen3.6-27B};
\node[font=\sffamily\bfseries\large, text=mindlabblue] at (12.02,6.18)
  {GLM-5.2};
\node[text=mindlabfg!68] at (3.88,5.76) {64 layers, hidden width 5,120};
\node[text=mindlabfg!68] at (12.02,5.76) {78 layers, hidden width 6,144};

\node[font=\sffamily\bfseries, text=mindlabblue!80!black, rotate=90]
  at (0.58,4.92) {Attention};
\node[attn, text width=3.72cm] (qgdn) at (3.52,4.92)
  {48 recurrent GDN layers\\compact recurrent prompt state};
\node[attn, text width=1.25cm, fill=mindlabblue!13] (qfull) at (6.55,4.92)
  {16 full GQA\\KV pages};

\node[anchor=west, font=\sffamily\bfseries, text=ForestGreen!68!black]
  at (0.35,3.68) {FFN};
\node[dense, text width=5.70cm] at (4.37,3.68)
  {dense gated FFN in all 64 \mbox{layers} \quad intermediate width 17,408};

\node[state, text width=2.55cm] at (2.60,2.32)
  {No expert router\\no EP all-to-all};
\node[state, text width=2.55cm] at (5.76,2.32)
  {Global CP8 LSE/output merge\\BF16 \mbox{numerator} reduction};
\node[boundary, text width=6.15cm] at (4.12,1.12)
  {Global partitioned forward; page-owner dK/dV adapter synchronization audit};

\node[font=\sffamily\bfseries, text=mindlabblue!80!black, rotate=90]
  at (8.70,4.92) {Attention};
\node[attn, text width=1.58cm] at (10.50,4.92)
  {21 index-compute\\MLA/DSA \mbox{layers}};
\node[attnshare, text width=3.35cm] at (13.55,4.92)
  {57 IndexShare consumers\\reuse per-forward selections};

\node[anchor=west, font=\sffamily\bfseries, text=ForestGreen!68!black]
  at (8.50,3.68) {FFN};
\node[dense, minimum width=1.18cm] at (10.10,3.68)
  {3 dense};
\node[moe, text width=3.98cm] at (13.25,3.68)
  {75 MoE layers \quad 256 routed, top-8 + 1 shared};

\node[state, text width=2.55cm] at (10.72,2.32)
  {CPU MLA latent + index-key pages\\CP32 distributes prompt storage};
\node[state, text width=2.55cm] at (14.02,2.32)
  {EP32 distributes\\expert weights\\eight routed copies\\per token};
\node[state, draw=ForestGreen!72!black, fill=green!10, text width=6.15cm] at (12.25,1.12)
  {Exact-2M: real rollout + global top-2,048 across CP32 + two 78-layer backwards + gradient finalization + one optimizer step};

\node[font=\sffamily\scriptsize, text=mindlabfg!58] at (7.95,0.30)
  {Bands group layer types and counts; widths are schematic and do not encode a shared execution order.};

\end{tikzpicture}%
}
\caption{\textbf{Two independent architecture axes.} Blue bands encode
token mixing and retained prompt state; green bands encode FFN execution and
routing. Qwen combines 48 recurrent GDN and 16 full-attention layers with
dense FFNs. GLM combines 21 index-computing and 57 IndexShare layers with three
dense and 75 MoE FFNs. The orange boxes state the measured scope of each
path: Qwen establishes global forward partitioning and completes the stated
response-only objective, while page-owner K/V adapter finalization requires a
separate replica-equivalence test. GLM composes global DSA and finalizes one
exact-2M online GRPO update after two 78-layer backwards.}
\label{fig:architecture-blueprint}
\end{figure}

Figure~\ref{fig:architecture-blueprint} opens both decoder stacks at the level
needed by the training runtime. The diagram shows counts and ownership rather
than implying that the two models share one layer template.

\subsection{Qwen: Dense FFNs with Hybrid Token Mixing}

The inspected \qwenmodel configuration has hidden width 5,120, 64 decoder
layers, 48 \texttt{linear\_attention} entries, and 16
\texttt{full\_attention} entries~\citep{qwen36config}. The implementation
instantiates the former with its recurrent GDN module. Gated DeltaNet supplies
that module's gated delta rule~\citep{yang2025gateddeltanet}, while
grouped-query attention shares fewer
KV heads across a larger set of query heads~\citep{ainslie2023gqa}. Every layer
ends in a dense gated feed-forward network with
intermediate width 17,408. A gated dense FFN applies the same three matrices to
every token in the SwiGLU form~\citep{shazeer2020glu},
\begin{equation}
F_{\mathrm{dense}}(h)=
W_2\!\left(\operatorname{SiLU}(W_1h)\odot W_3h\right).
\label{eq:dense-ffn}
\end{equation}
There is no token router and no expert all-to-all. Once the long-prompt FFN
graph is detached, a response replay only invokes the same dense matrices for
the short suffix.

The two token mixers expose different prompt-state scaling. A GDN layer carries
a fixed-shape recurrent state across the prompt boundary. A full-attention
layer carries key and value pages whose storage grows with \(P\). Full attention
also has a global semantic requirement: every response query must attend to
pages owned by all context-parallel ranks. Qwen thus places most of the
long-context storage problem in 16 layers, while the other 48 layers contribute
compact recurrent state.

The reported Qwen implementation uses NF4 QLoRA with 116,727,808 trainable
parameters. Quantized base weights reduce persistent model storage, but they do
not by themselves remove prompt pages or the response activation graph. The
working design is therefore built around physical page compaction and
conditional replay, not only parameter quantization~\citep{dettmers2023qlora}.

\subsection{GLM Attention: MLA, DSA, and IndexShare}

Multi-head latent attention (MLA) compresses per-token KV content into a latent
representation~\citep{deepseekv2}. The inspected \glmmodel configuration has
hidden width 6,144 and 78 decoder layers. Its KV and query latent widths are 512
and 2,048, respectively~\citep{glm52config}. Its sparse indexer has 32
heads of dimension 128 and selects at most 2,048 preceding positions for each
query~\citep{glm52config}. Following the DSA definition~\citep{deepseekv32}, an index score can be
written as
\begin{equation}
I_{t,s}=\sum_{j=1}^{H_I}w^I_{t,j}
\operatorname{ReLU}\!\left(\left\langle q^I_{t,j},k^I_s\right\rangle\right),
\qquad H_I=32,
\label{eq:dsa-index}
\end{equation}
and the sparse response output is
\begin{equation}
u_t=\operatorname{Attn}\!\left(
q_t,\{c_s:s\in\operatorname{TopK}(I_{t,:},2048)\}
\right).
\label{eq:dsa-attention}
\end{equation}
The 2,048 selected positions define the operator, not merely its memory layout.
Faithful distributed replay requires global selection over the logical context
and correct composition of the selected values.

Computing a new sparse index in every layer would repeat similar work. The
inspected GLM configuration instead yields 21 index-computing layers and 57
IndexShare layers that consume a selection published by a nearby source
layer~\citep{glm52config}. This cross-layer reuse resembles the mechanism
analyzed by IndexCache~\citep{bai2026indexcache}. It changes the
response implementation in two ways. First, only the 21 compute layers need durable
prompt indexer-key pages. Second, the response forward must maintain a
per-forward producer/consumer holder whose index schedule cannot span response
branches or parameter versions.

\subsection{GLM Feed-Forward Path: Dense then MoE}

The inspected configuration assigns dense FFNs to the first three GLM decoder
layers and 256 routed experts with top-8 routing plus one shared expert to the
remaining 75 layers~\citep{glm52config}. For token \(t\),
\begin{equation}
F_{\mathrm{MoE}}(h_t)=F_{\mathrm{shared}}(h_t)+
\sum_{e\in\operatorname{Top8}(g(h_t))}p_{t,e}F_e(h_t).
\label{eq:moe-ffn}
\end{equation}
MoE sparsity reduces the number of experts evaluated for one token, but it
creates a distributed parameter and data-movement problem. EP ranks hold
different experts. The router expands one token into eight expert assignments,
dispatches those rows to their owners, evaluates the selected expert paths,
and combines the outputs
~\citep{shazeer2017moe,lepikhin2020gshard,fedus2021switch}.

The parameter and activation scales explain why full-sequence GLM autograd is
not made cheap by sparsity. An expert has intermediate width 2,048, so its
three gated-FFN matrices contain
\begin{equation}
3\times6144\times2048=37{,}748{,}736
\end{equation}
weights. The 256 routed experts in one sparse layer contain approximately
9.66 billion weights, even though a token activates only eight of them. At a
balanced 65,536-token CP shard, top-8 routing produces
\(65{,}536\times8=524{,}288\) expert-token rows. One BF16 buffer with shape
\([524{,}288,6144]\) occupies exactly 6~GiB before outputs, permutations,
LoRA intermediates, or routing skew. A conventional beyond-2M-context graph can hold several
such tensors across layers and backward phases.

Expert parallelism distributes the parameter working set, but it does not
change the number of routed token copies. Context parallelism distributes
attention state, but it does not place experts. Prior MoE folding work studies
heterogeneous mappings across these parallel dimensions
~\citep{liu2025moefolding}. In our run, folding CP32 and EP32 over the same ranks
reduces separate process groups and matches the available 32 GPUs; the two
dimensions still solve different ownership and communication problems.

\subsection{What the Long Prompt Must and Need Not Retain}

Both models still execute every prompt token through every layer. The next
layer needs the output even when the current layer is sparse or recurrent. The
capacity gain comes from tensor lifetime: dense FFN intermediates, MoE routes,
expert-token permutations, attention scratch, and adapter activations from the
prompt are allowed to die immediately. Only the conditional state required by
future response tokens survives. For Qwen this is compact GDN and KV state. For
GLM it is MLA latent pages, selected indexer-key pages, position/page metadata,
and the rules needed to reconstruct IndexShare during each short replay.

\section{LongStraw: Long-Context Execution Design}
\label{sec:method}

LongStraw is the execution stack
developed in this report. Its design follows one rule: retain a tensor across the prompt
boundary only when a later response token depends on it. The rule applies to
physical allocations, not just logical tensor views. It also separates
state required by the model operator from scratch and activations required only
while producing that state.

Throughout this report, \emph{budget-constrained} means a fixed accelerator
count together with each device's H20 memory limit: eight H20 GPUs for the Qwen
path and 32 H20 GPUs for the GLM path. We report elapsed time, allocated GPU
memory, and phase-level event traces where the receipts provide them. The
resource axis is fixed-device execution for the stated transaction; the tables
keep the available fields separate from phase-specific timing and allocation
diagnostics for every reported run.

\paragraph{System scope.}
LongStraw is more than a resident-state manager. It is an
\emph{objective-aware, architecture-aware resident-state virtualization,
response-replay, and distributed-gradient execution system}. ``Objective-aware''
means that state lifetime follows the downstream training objective: old and
reference scores need values but no backward graph, policy replay needs the
response graph, and an optimizer step decides whether a prompt state can remain
resident or must be recaptured. ``Architecture-aware'' means that the retained
state and its owner are derived from the model operator--Qwen keeps GDN state
and dense KV pages, whereas GLM keeps MLA/DSA state and re-enters the MoE tail.
``Virtualization'' means that one logical prompt boundary is mapped to physical
GPU or CPU pages with explicit ownership, append/pop transactions, and a
version check performed at every replay and restore boundary.

\paragraph{One-step procedure.}
For exact-2M GLM, Tinker manages a vLLM rollout at TP8/PP4 before DAPO reward
computation. The same 32 GPUs then
switch to the CP32/EP32 training layout for replay and optimization.
For prompt state \(S_P\) and group member \(i\), the transaction is
\begin{equation}
\begin{aligned}
S_P &\leftarrow \operatorname{no\_grad\_capture}(x_{1:P};\theta_k),\\
\text{for }i=1,\ldots,G:\quad
&\operatorname{restore}(S_P);\quad
  \operatorname{score}_{\mathrm{old/ref}}(y_i)\ [\text{no graph}];\\[-0.2ex]
&\operatorname{replay}_{\mathrm{policy}}(y_i)\ [\text{response graph}];\quad
  \operatorname{backward}(\ell_i);\quad
  \operatorname{pop\_response\_state},\\
&\operatorname{finalize\_distributed\_gradients};\quad
  \operatorname{optimizer.step}();\quad
  \operatorname{zero\_grad}().
\end{aligned}
\label{eq:longstraw-transaction}
\end{equation}
The final state action is objective-dependent: exact replay computes
\(S_P(\theta_{k+1})\) again, while a resident multi-step run retains the
resident boundary across optimizer cycles to amortize prefill.
State retention decides what persists; replay decides what is recomputed; and
gradient aggregation combines rank-owned contributions into one distributed
optimizer update according to parameter ownership.

The corresponding live-memory boundary is
\begin{equation}
M_{\mathrm{live}}\approx M_{\mathrm{weights}}
 +M_{\mathrm{resident\ state}}(P)
 +M_{\mathrm{one\ response\ graph}}(B_R)
 +M_{\mathrm{response\ gradient\ pages}},
\label{eq:longstraw-memory-boundary}
\end{equation}
instead of retaining a full prompt graph for each old/reference/policy role and
all \(G\) responses. Prompt computation remains real, and architecture-specific
resident state can still grow with length; LongStraw bounds graph lifetime and
physical ownership rather than eliminating compute.

\subsection{Capture Once, Replay the Suffix}

For update \(k\), the runtime performs the following transaction.

\begin{enumerate}[
  label=\textbf{Phase \arabic*:},
  wide=0pt,
  leftmargin=0pt,
  labelsep=0.65em,
  align=left,
  widest=4,
  itemsep=0.20\baselineskip
]
  \item Run \(x_{1:P}\) under \(\theta_k\) with autograd disabled. At each
  layer, save the model-specific prompt state and release
  transient hidden tensors, attention scratch, FFN activations, and MoE routing
  buffers.
  \item Treat the completed prompt state as read-only. Materialize each
  response's old/reference log-probabilities before its policy backward,
  without updates between group members. GLM freezes both old score sets before
  replay; Qwen uses a member-serial
  old/reference/policy schedule.
  \item For each group member, rebuild the short current-policy response path
  under autograd. Reuse the same read-only prompt state, backpropagate the
  member loss, and immediately release that member's graph.
  \item After all \(G\) backwards, retain the accumulated local gradients and
  issue one optimizer call per worker. Exact response-only replay recaptures the
  prefix under the updated parameters before another step. Resident reuse keeps
  the prior cache across cycles to amortize prefill.
\end{enumerate}

This schedule changes the dominant activation scale from \(P+R\) to \(R\), but
prompt compute remains. Phase~1 still sends the full prompt through every
decoder layer and retains architecture-specific state: full-attention pages,
recurrent state, and DSA latent values plus index keys. The multi-step mode
reuses that expensive state across cycles; its execution is reported separately
from the exact within-update response-only contract.

\begin{figure}[t]
\centering
\resizebox{\textwidth}{!}{%
\begin{tikzpicture}[
  font=\sffamily\scriptsize,
  phase/.style={draw=mindlabfg!35, fill=mindlabfg!4, line width=0.6pt,
    minimum width=2.28cm, minimum height=0.83cm, align=center,
    font=\sffamily\small},
  rowlabel/.style={anchor=east, align=right,
    font=\sffamily\bfseries\scriptsize},
  gpu/.style={draw=mindlabblue!78!black, fill=mindlabbluepale!58,
    rounded corners=1.2pt, line width=0.65pt, align=center},
  cpu/.style={draw=BurntOrange!80!black, fill=yellow!12,
    rounded corners=1.2pt, line width=0.65pt, align=center},
  transient/.style={draw=mindlabfg!48, fill=mindlabfg!7,
    rounded corners=1.2pt, line width=0.6pt, align=center},
  autograd/.style={draw=ForestGreen!72!black, fill=green!12,
    rounded corners=1.2pt, line width=0.7pt, align=center},
  gradient/.style={draw=mindlabblue!88!black, fill=mindlabblue!14,
    rounded corners=1.2pt, line width=0.75pt, align=center, inner xsep=2pt},
  stale/.style={draw=BurntOrange!90!black, fill=BurntOrange!10,
    rounded corners=1.2pt, dashed, line width=0.75pt, align=center}
]

\def\xzero{3.45}
\def\xone{5.55}
\def\xtwo{7.65}
\def\xthree{9.75}
\def\xfour{11.85}
\def\xfive{13.95}

\node[phase, minimum width=2.06cm] at (4.50,6.02) {Capture\\\scriptsize no grad};
\node[phase, minimum width=2.06cm] at (6.60,6.02) {Old + reference\\\scriptsize frozen scoring};
\node[phase, minimum width=2.06cm] at (8.70,6.02) {Policy forward\\\scriptsize autograd};
\node[phase, minimum width=2.06cm] at (10.80,6.02) {Backward\\\scriptsize autograd};
\node[phase, minimum width=2.06cm] at (12.90,6.02) {Finalize + step\\\scriptsize once};

\draw[mindlabfg!22, line width=0.45pt] (\xzero,0.42) rectangle (\xfive,5.48);
\foreach \x in {\xone,\xtwo,\xthree,\xfour}
  \draw[mindlabfg!18, line width=0.45pt] (\x,0.42) -- (\x,5.48);
\foreach \y in {1.42,2.42,3.42,4.42}
  \draw[mindlabfg!18, line width=0.45pt] (\xzero,\y) -- (\xfive,\y);

\node[rowlabel, text=mindlabblue!82!black] at (3.00,4.92)
  {Qwen durable prompt state\\GPU resident};
\node[rowlabel, text=BurntOrange!88!black] at (3.00,3.92)
  {GLM durable prompt state\\CPU resident};
\node[rowlabel, text=mindlabfg!72] at (3.00,2.92)
  {Transient prompt work\\GPU, no autograd};
\node[rowlabel, text=ForestGreen!68!black] at (3.00,1.92)
  {Response activations\\GPU / autograd};
\node[rowlabel, text=mindlabblue!88!black] at (3.00,0.92)
  {Trainable gradients\\GPU};

\node[gpu, minimum width=10.25cm, minimum height=0.66cm] at (8.70,4.92)
  {captured under \(\theta_k\) \quad read-only for every score and response branch};
\node[cpu, minimum width=10.25cm, minimum height=0.66cm] at (8.70,3.92)
  {CPU pages remain live \quad one layer is staged to GPU during scoring and replay};

\foreach \x in {6.18,6.73,8.28,8.83,10.38,10.93}
  \filldraw[fill=mindlabblue!48, draw=mindlabblue!75!black, line width=0.35pt]
    (\x,3.63) rectangle +(0.28,0.12);
\node[anchor=north, font=\sffamily\scriptsize,
  text=mindlabblue!80!black, fill=yellow!12, inner sep=1pt]
  at (8.70,3.61) {blue ticks: transient per-layer GPU stage};

\node[transient, minimum width=1.90cm, minimum height=0.66cm] at (4.50,2.92)
  {layer-local scratch\\released immediately};
\draw[mindlabfg!32, dashed] (\xone,2.92) -- (\xfive,2.92);
\node[fill=white, inner sep=1.5pt, text=mindlabfg!55] at (9.75,2.92)
  {no prompt activation graph survives};

\node[transient, minimum width=1.90cm, minimum height=0.66cm] at (6.60,1.92)
  {score transient\\not saved};
\node[autograd, minimum width=1.90cm, minimum height=0.66cm] at (8.70,1.92)
  {save member \(i\)\\response graph};
\node[autograd, minimum width=1.90cm, minimum height=0.66cm, fill=green!6]
  at (10.80,1.92) {consume graph\\then free};
\draw[-Latex, draw=ForestGreen!72!black, line width=0.65pt]
  (11.55,1.56) to[out=315,in=225,looseness=1.05]
  node[pos=0.72, below, font=\sffamily\scriptsize,
  text=ForestGreen!68!black, fill=white, inner sep=0.8pt]
  {serial \(i=1,\ldots,G\)} (7.95,1.56);

\node[gradient, text width=1.75cm, minimum width=1.85cm,
  minimum height=0.66cm] at (10.80,0.92)
  {sum \(G\) local\\gradients};
\node[gradient, text width=1.55cm, minimum width=1.75cm,
  minimum height=0.66cm, fill=mindlabblue!8]
  at (12.90,0.92) {finalize owners\\step + clear};

\draw[BurntOrange!90!black, dashed, line width=1.0pt]
  (\xfive,0.30) -- (\xfive,6.47);
\node[stale, text width=1.35cm, minimum height=0.66cm, anchor=west]
  at (14.08,4.92) {STALE\\under \(\theta_{k+1}\)};
\node[stale, text width=1.35cm, minimum height=0.66cm, anchor=west]
  at (14.08,3.92) {STALE\\recapture required};
\node[stale, text width=1.35cm, minimum height=0.66cm, anchor=west,
  font=\sffamily\bfseries\scriptsize]
  at (14.08,0.92) {next loop\\requires recapture};

\end{tikzpicture}%
}
\caption{\textbf{State lifetime across one grouped update.} Durable Qwen
state remains GPU-resident; GLM pages remain CPU-resident and only one layer is
staged at a time. Prompt activations are released during no-grad capture,
whereas only the current response graph is live under autograd. Group-indexed
inputs and frozen scores may still grow with \(G\). Worker-local gradients
accumulate over \(G\) serial backwards, are finalized across their ownership
groups, and feed one optimizer step. The optimizer
changes \(\theta_k\) to \(\theta_{k+1}\), making every captured prompt state
stale; a repeated loop would require recapture.}
\label{fig:state-lifetime}
\end{figure}
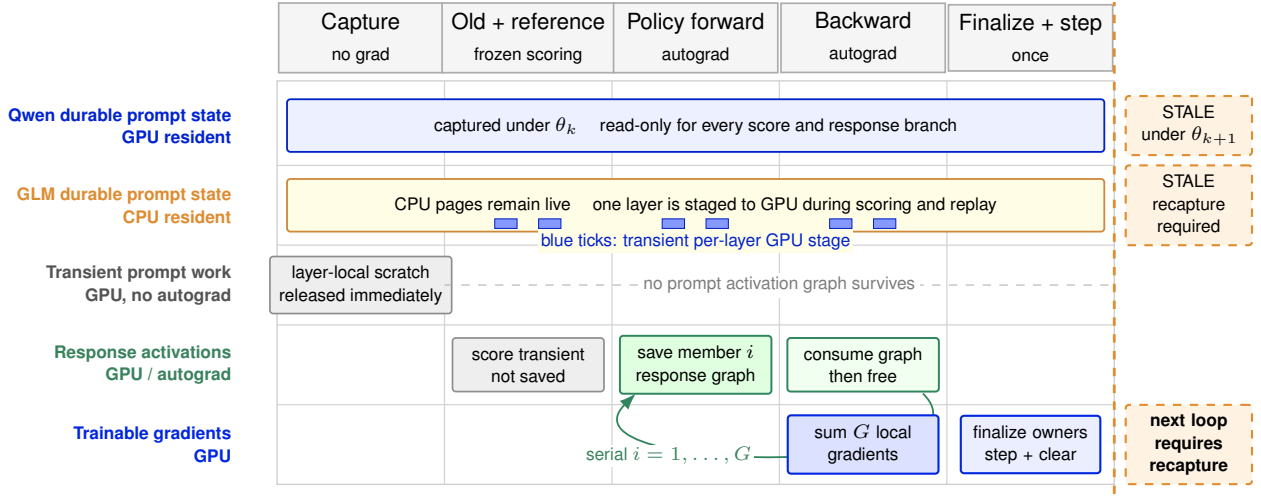

Nothing in the transaction is specialized to \(G=2\) or \(G=8\). For a
member-serial group, the leading resource accounting is
\begin{equation}
\begin{aligned}
M_{\mathrm{live}} &\approx M_{\mathrm{fixed}}+M_{\mathrm{prompt}}(P)
 +M_{\mathrm{grad}}+\max_i M_{\mathrm{branch}}(R_i)
 +M_{\mathrm{score}}\!\left(\sum_i R_i\right), \\
T_{\mathrm{update}} &= T_{\mathrm{prompt}}(P)
 +\sum_{i=1}^{G}T_{\mathrm{score+replay}}(R_i).
\end{aligned}
\label{eq:serial-group-scaling}
\end{equation}
Equation~\ref{eq:serial-group-scaling} bounds the live policy graph by the
largest member while input/label objects, rewards, reports, and frozen
old/reference scores scale with \(G\) or \(\sum_i R_i\). The runners accept a
configured list of group members rather than hard-coding \(G=2\) or \(G=8\).
The Qwen receipts use \(G=2\) and \(G=8\), and the GLM exact-2M receipt uses a
nondegenerate \(G=2\) reward pair; \(G=1\) is retained as the single-member
control in the staged route.

\noindent\begin{minipage}{\textwidth}
A coherent distributed update routes every sharded gradient contribution to its
parameter owner and keeps replicated adapters synchronized after the optimizer
call. This ownership contract is independent of response-only execution: the
Qwen runs complete the evaluated response-only procedure, while the CP8
prototype all-reduces \(dQ\) and records page-owner \(dK/dV\) contributions in
the companion adapter audit. The current GLM exact-2M run performs global
cross-CP DSA and calls
\texttt{finalize\_model\_grads} after both response backwards and before its
single optimizer step, closing the complete exact-2M distributed update.
\end{minipage}

\Needspace{5\baselineskip}
\subsection{State Inventory and Ownership}

Table~\ref{tab:prefix-state-inventory} distinguishes durable prompt state from
transient prompt work. The distinction is the basis for both memory accounting
and correctness during layerwise response replay.

\begin{table}[!htbp]
\centering
\caption{Prompt-boundary state inventory. Durable rows survive no-grad prefix
capture and are read by response replay; transient rows are released after
capture and recomputed only for the short response under autograd. All replay
uses a detached prompt state.}
\label{tab:prefix-state-inventory}
\scriptsize
\setlength{\tabcolsep}{3pt}
\renewcommand{\arraystretch}{1.16}
\begin{tabularx}{\textwidth}{@{}C{0.20\textwidth}
    C{0.20\textwidth}
    Y
    Y@{}}
\toprule
State & Placement and ownership & Scaling with prompt length $P$ & Use during response replay \\
\midrule
\rowcolor{mindlabbluepale!20}
Qwen GDN recurrent state & GPU; retained per GDN layer on each CP8 rank & Fixed-size recurrent boundary state; does not grow linearly with $P$ & Restores the recurrent state for the response transition; prompt-state gradients remain detached \\
\rowcolor{mindlabbluepale!20}
Qwen full-attention KV pages & Compact GPU pages physically owned across CP8 for 16 layers & $O(P/8)$ KV storage per rank & Each response query reads every shard through the global CP8 LSE/output merge \\
\midrule
\rowcolor{green!7}
GLM MLA latent pages & CPU; Megatron-zigzag CP32 shards for all 78 layers & $O(P/32)$ per rank; at $P=2{,}097{,}152$, each rank owns 1,024 pages and 65,536 tokens & Staged layer by layer; selected values move from CP owners into the global DSA output \\
\rowcolor{green!7}
GLM DSA indexer-key pages & CPU; CP32 shards retained only for 21 index-computing layers & $O(P/32)$ at those 21 layers & Supplies local candidates with global positions; CP32 forms global top-2048, reused by 57 IndexShare layers \\
\rowcolor{green!7}
GLM page and position metadata & Host page IDs and valid-token counts; replay constructs device-side positions & $O(P/(32\mathbin{\times}64))$ page records at page size 64 & Restores Megatron page order, global token positions, and causal bounds for local replay \\
\midrule
Qwen transient response work & GPU attention scratch, dense-FFN activations, and temporary response pages & Prompt work is not retained; autograd storage follows the response block & Recomputed blockwise for policy backward, then released \\
GLM transient attention/MoE work & GPU DSA scores/top-k holder, attention scratch, router decisions, permuted rows, and selected-expert intermediates & Prompt tensors exist during no-grad capture but do not survive it; replay storage follows response length and routing & Recomputed inside whole-layer checkpointing; IndexShare state is per forward and EP32 dispatches selected response rows \\
\bottomrule
\end{tabularx}
\end{table}

Logical sharding is insufficient when a retained tensor is a view into a
larger allocation. The allocator cannot release the parent while any view is
alive. The Qwen implementation therefore copies every retained page shard into
a right-sized physical allocation. GLM applies the same rule on CPU: a restored
layer's page table, local positions, latent pages, and indexer pages must all
describe one local shard.

Stored state is immutable within one grouped update. Response branches may
append suffix pages or build temporary IndexShare selections, but cannot mutate
the shared prefix. Branch-local state is released after scoring or backward so
every response observes identical conditioning state.

\subsection{Device Placement and Layer Staging}

Qwen retains compact GDN and KV state on GPU because eight-way context
parallelism makes the per-rank state fit and the response attention merge reads
all shards repeatedly. GLM retains its CP-sharded MLA and indexer-key pages on
CPU. During response replay, it stages the pages needed by one layer, executes
the short response layer, and releases or returns the staged copy before
advancing. A shared RoPE cache avoids rebuilding position tensors for every
layer and branch.

CPU placement trades transfer time for bounded residency. The trade is useful
only because replay is layerwise. Copying the complete 78-layer prefix back to
GPU would recreate the storage peak. Conversely,~staging tensors without
preserving their page order and global positions would change the attention
operator. Device placement and logical ownership are therefore a joint
implementation requirement.

\subsection{Whole-Layer Checkpointing}

Activation checkpointing trades retained tensors for recomputation, either at
a whole block or at selected operations within it
~\citep{chen2016training,korthikanti2022recomputation}.
For our native MoE replay, however, the required boundary is the complete layer.
Attention-only checkpointing does not bound a GLM response graph if the MoE
tail retains router outputs, dispatch permutations, expert inputs, LoRA
intermediates, and concatenation buffers. The working path checkpoints the
complete decoder layer. Forward saves the short layer input and minimal
metadata; backward recomputes attention projection, sparse selection,
IndexShare publication or consumption, output projection, router decisions,
expert dispatch/combine, and the selected dense or expert LoRA paths without
retaining the complete layer graph across backward or any prompt activation.
Checkpointing is applied to the short response graph, not the full 2,097,152-position prompt. The
prompt was already evaluated without autograd. This distinction explains why
checkpointing succeeds here while conventional full-sequence checkpointing
still exposes large routed-token and attention workspaces during backward
recomputation through the complete decoder stack.

\Needspace{5\baselineskip}
\subsection{Parallel Layout}

The parallel dimensions distribute different objects. Table~\ref{tab:parallel-layout}
lists the layouts used by the evaluated runs.

\begin{table}[!htbp]
\centering
\caption{Parallel layouts and their ownership contracts. Context parallelism
partitions prompt state, whereas expert parallelism partitions routed experts;
each path records the collectives used by its measured update.}
\label{tab:parallel-layout}
\scriptsize
\setlength{\tabcolsep}{3pt}
\renewcommand{\arraystretch}{1.18}
\begin{tabularx}{\textwidth}{@{}C{0.17\textwidth}
    Y
    Y
    Y@{}}
\toprule
Path and topology & State or expert ownership & Response-time collectives & Measured semantics and update path \\
\midrule
\rowcolor{mindlabbluepale!20}
Qwen3.6-27B\newline 8 H20, CP8 & Compact full-attention KV pages and GDN boundary state are distributed across eight context ranks; dense FFNs have no expert ownership & Forward response attention uses the global CP8 max/normalizer/value-sum LSE/output merge; the stated response-only backward and local optimizer calls complete & $dQ$ is all-reduced; page-owner $dK/dV$ and replicated-adapter synchronization are recorded in a companion CP8 audit \\
\midrule
\rowcolor{green!7}
GLM-5.2\newline 32 H20\newline vLLM rollout TP8/PP4;\newline training\newline TP1/CP32/EP32/\newline ETP1/PP1 & The same GPUs are reused by phase. CP32 owns 1,024 CPU prefix pages (65,536 tokens) per rank; EP32 nominally owns eight of 256 routed experts per rank & The vLLM rollout samples at TP8/PP4. Training exchanges global DSA candidates/values across CP32, then dispatches and combines routed response rows across EP32 & At exact 2M, all ranks report two 78-layer backwards, \texttt{finalize\_model\_grads}, and one optimizer step \\
\bottomrule
\end{tabularx}
\end{table}

Qwen uses CP8 to distribute full-attention pages and recurrent state over eight
GPUs. Its dense FFNs do not require expert dispatch. GLM training uses TP1/CP32/EP32/
ETP1/PP1. CP32 distributes long attention state. EP32 distributes 256 routed
experts, nominally eight per rank. The same 32 ranks participate in both groups,
but CP collectives cannot replace EP collectives: response attention needs a
cross-context selection or reduction, while MoE needs token dispatch and
combine across expert owners.

The exact-2M online transaction cannot colocate its rollout and training
topologies within the fixed budget. LongStraw releases the Megatron training
actor, starts a Tinker-managed multi-node vLLM policy sampler at TP8/PP4 on the
same 32 H20s, records
the two completions and old log-probabilities, then releases the sampler and
restores Megatron at TP1/CP32/EP32/ETP1/PP1. The policy-LoRA content hash is
checked across this handoff. The temporal handoff makes topology phase-specific
and avoids simultaneous residency of the two runtimes on the fixed device inventory
for this run.

The checkpoint declares 1,048,576 native positions. For the 2M transaction,
rollout and training both receive an explicit YaRN configuration with factor
2.000015 and \texttt{max\_position\_embeddings=2097168}. Sharing this position
contract keeps sampling and training aligned throughout the exact-2M workflow.

\subsection{Adapter and Optimizer Scope}

The final GLM run uses rank-8 LoRA over the configured attention projections,
dense FFNs, routed and shared expert FFNs, and the output head. Base weights,
embeddings, normalization parameters, router parameters, and the DSA indexer
remain frozen. The Qwen path uses NF4 QLoRA over its configured dense model
targets. These are parameter-efficient adapter updates; neither set of
experiments studies full-parameter model
training~\citep{hu2021lora,dettmers2023qlora}. Rank allocation, adapter sharing,
mixed-precision fine-tuning, and pruning are complementary design axes
~\citep{zhou2025rankadaptor,zhou2025bslora,zhou2026balancing,zhou2026autoqra,zhou2025largecompression,zhou-etal-2025-qpruner};
LongStraw holds these choices fixed to isolate execution lifetime and
distributed state management.

All old-policy values are frozen before policy replay. In the exact-2M GLM
transaction, rollout supplies the old log-probabilities and the old snapshot is
also the reference policy. With \(\epsilon=0.2\) and \(\beta=0\), the measured
importance ratios include both one lower-clipped and one upper-clipped token
across the two members. Thus the update exercises a nondegenerate reward path
and both sides of the clipped objective.

\section{Qwen: Dense Hybrid Replay within an Eight-H20 Budget}
\label{tr:qwen-path}

The Qwen path is the cleaner of the two architecture cases because it has no
expert router, no expert-parallel token exchange, and no sparse index whose
meaning changes under context sharding.  It is nevertheless not a conventional
data-parallel training job.  The implementation replicates the dense model and
response query computation across eight ranks while distributing the long
key/value sequence of each full-attention layer.  This section describes that
implementation at the tensor level.  The global conditional forward operator
runs within the fixed eight-H20 envelope, while page-owner K/V gradient
composition is evaluated separately.

\subsection{Hybrid Layer Anatomy and the Prompt Boundary}

The examined model snapshot contains 64 decoder layers at hidden width
5,120, with a repeating pattern of three \texttt{linear\_attention} entries and
one \texttt{full\_attention} entry~\citep{qwen36config}. The implementation
maps the former to its gated-delta-network (GDN) module, giving 48 GDN layers
and 16 full-attention layers. The recurrent mechanism follows Gated DeltaNet
~\citep{yang2025gateddeltanet}. In the pinned snapshot, each full-attention
layer has 24 query heads, four KV heads, head dimension 256, an output gate,
and a partial rotary factor of 0.25~\citep{qwen36config}. The 24-to-4 query/KV
arrangement follows grouped-query attention~\citep{ainslie2023gqa}, while its
rotary position mechanism follows RoPE~\citep{su2021roformer}.
Every layer uses the
same dense gated FFN shape, with intermediate width 17,408.  Ignoring biases,
one FFN therefore contains
\begin{equation}
3\times 5{,}120\times 17{,}408 = 267{,}386{,}880
\label{eq:qwen-dense-ffn-weights}
\end{equation}
base weights.  Every token activates all three projections; there is no
conditional expert capacity to distribute.

The two token mixers leave different state at the end of a no-grad prompt.
For a GDN layer, the boundary consists of a recurrent matrix and the final
three-token convolution tail.  Its shape is independent of prompt length.  For
a full-attention layer, the boundary is every prompt key and value, so storage
is linear in prompt length.  Prompt capture evaluates the complete dense stack,
including every FFN, but releases prompt hidden states, FFN intermediates, and
temporary mixer work after each chunk.  Only 48 compact GDN boundaries and 16
full-attention KV page sets survive.  Thus pages serve the 16 quadratic-history
layers, while recurrence removes length-dependent state from the remaining 48
GDN layers.

The NF4 QLoRA path uses rank 16, scaling 32, learning rate
\(2\times10^{-4}\), and zero weight decay.  LoRA targets full-attention,
GDN, and dense-FFN projections, giving 116,727,808 trainable parameters.  NF4
reduces persistent base-weight storage, but it does not reduce KV length or
response activation lifetime~\citep{dettmers2023qlora}.  Those two terms are
handled by physical context parallelism and blockwise replay.

\subsection{From Logical Shards to Physical Pages}

Let page size be \(S=64\), global page index be \(p\), and the CP world size be
\(C=8\).  Page ownership is block-cyclic,
\begin{equation}
\operatorname{owner}(p)=p\bmod C.
\label{eq:qwen-page-owner}
\end{equation}
Each rank retains only its owned pages but keeps the global logical KV length.
Thus RoPE positions, query start indices, and causal masks use original
sequence coordinates, not compacted local ones.

This distinction was initially only logical.  A page obtained by slicing a
4,096-token prefix chunk could be contiguous as a tensor while still retaining
the storage of the complete parent chunk.  Discarding the unowned slice did not
release that parent allocation.  The manager instead allocates right-sized page
tensors and copies owned slices.  PagedAttention manages serving KV through
logical-to-physical block tables over~fixed-size physical blocks
~\citep{kwon2023efficient}.  At our training boundary, copying owned slices
makes allocator ownership agree with page-table ownership, so releasing a
logical page also releases its physical storage.

The arithmetic beyond two million context positions is explicit. The reported sequence has
2,088,960 prompt tokens followed by 8,192 response-input tokens, for an exact
context length of 2,097,152. The prompt contains
\begin{equation}
N_{\mathrm{page}}=\frac{2{,}088{,}960}{64}=32{,}640,
\qquad
N_{\mathrm{page/rank}}=\frac{32{,}640}{8}=4{,}080.
\label{eq:qwen-page-count}
\end{equation}
For BF16 K and V with four KV heads and head dimension 256, the raw prompt KV
payload retained by one rank across the 16 full-attention layers is
\begin{align}
B_{\mathrm{KV/rank}}
&=16\times4{,}080\times 2_{\{K,V\}}
  \times64\times4\times256\times2\ \text{bytes} \\
&=17{,}112{,}760{,}320\ \text{bytes}
 \approx 15.94\ \text{GiB}.
\label{eq:qwen-kv-bytes}
\end{align}
The measured allocator footprint is larger because it includes quantized
weights, adapters, recurrent state, page metadata, response pages, and
temporary kernels. Equation~\ref{eq:qwen-kv-bytes} exposes the ownership slope:
each additional prompt page resides on one CP rank per full-attention layer,
not on all eight ranks.

\subsection{Global Full-Attention Forward Composition}

For response query \(t\), let \(\mathcal{K}_r\) be the keys owned by rank \(r\)
and let \(s_{t,j}=q_t^\top k_j/\sqrt{d}\).  A local paged-attention kernel
returns a normalized local output and its row log-normalizer,
\begin{equation}
\ell_{r,t}=\log\!\sum_{j\in\mathcal{K}_r}e^{s_{t,j}},
\qquad
o_{r,t}=\frac{\sum_{j\in\mathcal{K}_r}e^{s_{t,j}}v_j}
                   {e^{\ell_{r,t}}}.
\label{eq:qwen-local-attn}
\end{equation}
The global output is reconstructed without moving the KV pages.

\begin{align}
m_t &= \max_r \ell_{r,t}, \\
a_t &= \sum_r e^{\ell_{r,t}-m_t}, \\
n_t &= \sum_r e^{\ell_{r,t}-m_t}o_{r,t}.
\label{eq:qwen-lse-parts}
\end{align}
Then
\begin{equation}
o_t=\frac{n_t}{a_t},
\qquad
\ell_t=m_t+\log a_t.
\label{eq:qwen-global-attn}
\end{equation}
This requires one MAX all-reduce and two SUM all-reduces per composed output.
It is the same stable log-sum-exp identity used by exact tiled attention
~\citep{rabe2021self,dao2022flashattention}.  It reconstructs the dense
conditional response-attention operator from disjoint KV partitions.

Production reduces \(n_t\) in BF16 to lower communication payload.  The maximum,
denominator, and global LSE remain FP32.  Here ``exact'' means exact partition
composition of the specified dense operator under that finite-precision
reduction, not bitwise equality with unsharded FP32 execution.

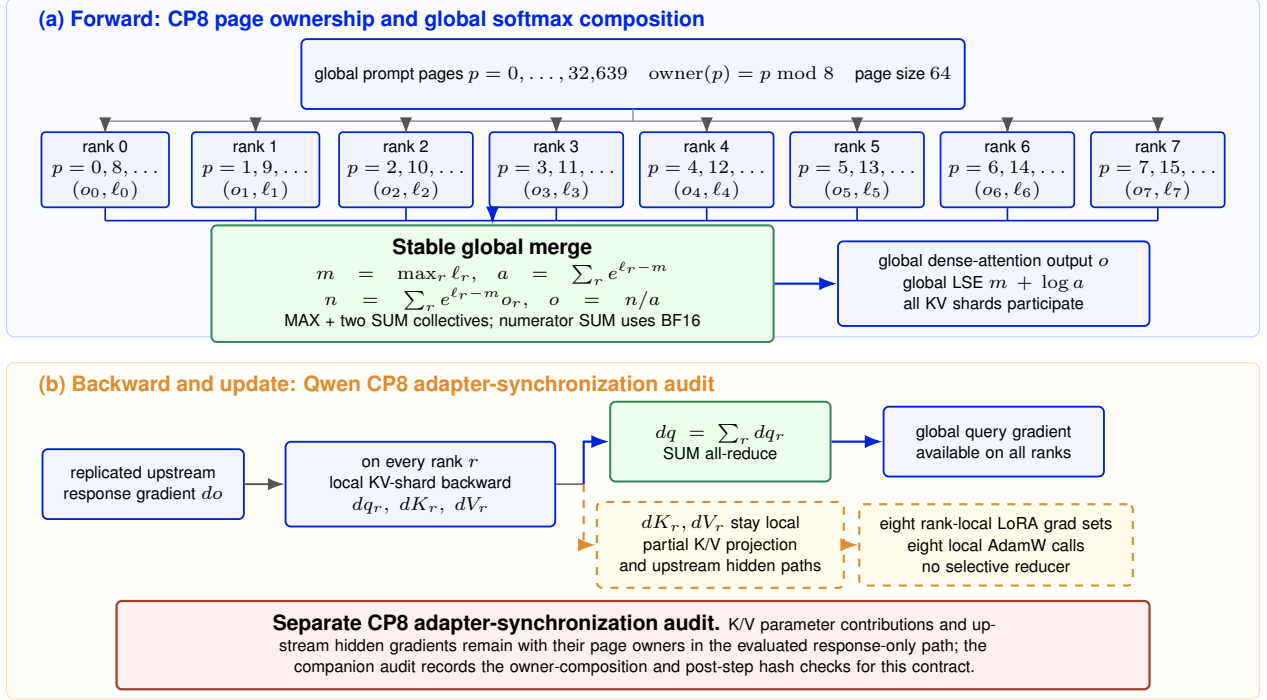
\begin{figure}[t]
\centering
\resizebox{\textwidth}{!}{%
\begin{tikzpicture}[
  font=\sffamily\scriptsize,
  page/.style={draw=mindlabblue!78!black, fill=mindlabbluepale!45,
    rounded corners=1.5pt, line width=0.65pt, align=center,
    minimum width=1.68cm, minimum height=0.90cm, inner sep=3pt},
  merge/.style={draw=ForestGreen!70!black, fill=green!8,
    rounded corners=2pt, line width=0.75pt, align=center,
    minimum height=1.05cm, inner sep=5pt},
  valid/.style={draw=mindlabblue!82!black, fill=mindlabbluepale!35,
    rounded corners=2pt, line width=0.7pt, align=center,
    minimum height=0.92cm, inner sep=5pt},
  risk/.style={draw=BurntOrange!88!black, fill=yellow!10,
    rounded corners=2pt, dashed, line width=0.8pt, align=center,
    minimum height=0.92cm, inner sep=5pt},
  warning/.style={draw=BrickRed!80!black, fill=red!6,
    rounded corners=2pt, line width=0.85pt, align=center,
    minimum height=0.86cm, inner sep=5pt},
  wire/.style={draw=mindlabfg!62, line width=0.62pt},
  flow/.style={-Latex, draw=mindlabfg!72, line width=0.72pt},
  goodflow/.style={-Latex, draw=mindlabblue!85!black, line width=0.82pt},
  riskflow/.style={-Latex, draw=BurntOrange!88!black, dashed, line width=0.82pt}
]

\filldraw[rounded corners=3pt, fill=mindlabbluepale!12, draw=mindlabblue!24]
  (-0.25,4.72) rectangle (16.25,9.18);
\filldraw[rounded corners=3pt, fill=yellow!3, draw=BurntOrange!28]
  (-0.25,-0.05) rectangle (16.25,4.38);

\node[anchor=west, font=\sffamily\bfseries\small, text=mindlabblue!88!black]
  at (0.05,8.88) {(a) Forward: CP8 page ownership and global softmax composition};

\node[valid, minimum width=7.15cm] (globalpages) at (8.00,8.18)
  {global prompt pages \(p=0,\ldots,32{,}639\) \quad
   \(\operatorname{owner}(p)=p\bmod 8\) \quad page size \(64\)};

\draw[draw=mindlabfg!46, line width=0.55pt] (8.00,7.72) -- (8.00,7.56);
\draw[draw=mindlabfg!46, line width=0.55pt] (1.05,7.56) -- (14.95,7.56);

\node[page] (r0) at (1.05,6.92) {rank 0\\\(p=0,8,\ldots\)\\\((o_0,\ell_0)\)};
\node[page] (r1) at (3.03,6.92) {rank 1\\\(p=1,9,\ldots\)\\\((o_1,\ell_1)\)};
\node[page] (r2) at (5.01,6.92) {rank 2\\\(p=2,10,\ldots\)\\\((o_2,\ell_2)\)};
\node[page] (r3) at (6.99,6.92) {rank 3\\\(p=3,11,\ldots\)\\\((o_3,\ell_3)\)};
\node[page] (r4) at (8.97,6.92) {rank 4\\\(p=4,12,\ldots\)\\\((o_4,\ell_4)\)};
\node[page] (r5) at (10.95,6.92) {rank 5\\\(p=5,13,\ldots\)\\\((o_5,\ell_5)\)};
\node[page] (r6) at (12.93,6.92) {rank 6\\\(p=6,14,\ldots\)\\\((o_6,\ell_6)\)};
\node[page] (r7) at (14.91,6.92) {rank 7\\\(p=7,15,\ldots\)\\\((o_7,\ell_7)\)};

\foreach \x in {1.05,3.03,5.01,6.99,8.97,10.95,12.93,14.91}{
  \draw[flow] (\x,7.56) -- (\x,7.39);
}

\node[fit=(r0)(r1)(r2)(r3)(r4)(r5)(r6)(r7), inner sep=1pt] (rankset) {};
\node[merge, text width=7.05cm] (lsemerge) at (6.15,5.42)
  {\textbf{Stable global merge}\\
   \(m=\max_r\ell_r,\quad a=\sum_r e^{\ell_r-m}\)\\
   \(n=\sum_r e^{\ell_r-m}o_r,\quad o=n/a\)\\
   MAX + two SUM collectives; numerator SUM uses BF16};
\node[valid, text width=3.75cm] (globalout) at (12.75,5.42)
  {global dense-attention output \(o\)\\global LSE \(m+\log a\)\\all KV shards participate};

\coordinate (rankbus) at (6.15,6.25);
\draw[wire, draw=mindlabblue!76!black] (1.05,6.25) -- (14.91,6.25);
\foreach \r in {r0,r1,r2,r3,r4,r5,r6,r7}{
  \draw[wire, draw=mindlabblue!76!black]
    (\r.south) -- (\r.south |- rankbus);
}
\draw[goodflow] (rankbus) -- (lsemerge.north);
\draw[goodflow] (lsemerge) -- (globalout);

\node[anchor=west, font=\sffamily\bfseries\small, text=BurntOrange!88!black]
  at (0.05,4.08) {(b) Backward and update: Qwen CP8 adapter-synchronization audit};

\node[valid, text width=2.30cm] (upstream) at (1.55,2.78)
  {replicated upstream\\response gradient \(do\)};
\node[valid, text width=3.20cm] (localbwd) at (5.20,2.78)
  {on every rank \(r\)\\local KV-shard backward\\\(dq_r,\ dK_r,\ dV_r\)};

\node[merge, text width=2.55cm] (dqsum) at (9.15,3.34)
  {\(dq=\sum_r dq_r\)\\SUM all-reduce};
\node[valid, text width=2.55cm] (globaldq) at (12.75,3.34)
  {global query gradient\\available on all ranks};

\node[risk, text width=2.90cm] (localkv) at (9.15,1.98)
  {\(dK_r,dV_r\) stay local\\partial K/V projection\\and upstream hidden paths};
\node[risk, text width=3.25cm] (localopt) at (12.78,1.98)
  {eight rank-local LoRA grad sets\\eight local AdamW calls\\no selective reducer};

\draw[flow] (upstream) -- (localbwd);
\coordinate (gradsplit) at (7.35,2.78);
\draw[wire] (localbwd.east) -- (gradsplit);
\draw[goodflow] (gradsplit) |- (dqsum.west);
\draw[goodflow] (dqsum) -- (globaldq);
\draw[riskflow] (gradsplit) |- (localkv.west);
\draw[riskflow] (localkv) -- (localopt);

\node[warning, text width=13.25cm] at (8.00,0.66)
  {\textbf{Separate CP8 adapter-synchronization audit.}
   K/V parameter contributions and upstream \mbox{hidden} gradients remain with
   their page owners in the evaluated response-only path; the companion audit
   records the owner-composition and post-step hash checks for this contract.};

\end{tikzpicture}%
}
\caption{\textbf{Qwen CP8 forward composition and adapter-synchronization audit.}
The full-attention forward partitions physical KV pages and recomposes the
global softmax over all shards, subject to the BF16 numerator reduction. Backward
all-reduces the query gradient, but K/V-derived parameter and hidden-gradient
contributions remain rank local before eight local optimizer calls. The receipt
therefore records global conditional-forward, direct response gradients, and
the evaluated response-only procedure, with replicated K/V-adapter agreement
kept as a separate ownership property.}
\label{fig:qwen-cp-merge}
\end{figure}

\subsection{Four-Block Response Replay}

The shared prompt is captured once in 510 chunks of 4,096 tokens.  A response
branch is then split into four 2,048-token blocks.  The branch first runs a
no-grad suffix pass.  At each block boundary it clones the input GDN states and
appends that block's KV pages.  The policy pass traverses the four blocks in
reverse.  For one block it restores the corresponding GDN input state with
gradient tracking, recomputes all 64 decoder layers under whole-layer
checkpointing, forms causal selected-token log-probabilities, runs backward,
propagates the recurrent/conv-state gradient to the preceding block, and pops
the temporary KV update.  Only one 2,048-token response graph is live at a
time, so block count increases replay work without retaining all block graphs
together across the complete suffix on each device.

For each group member the observed ordering is old scoring, reference scoring,
policy suffix forward, and policy reverse.  Members are serialized after the
shared prompt.  Their parameter gradients accumulate until one optimizer call
is issued after the last member.  This schedule explains why, for this
workload, increasing group size mainly increases time rather than peak memory.

The workload configuration is explicit.  First, the inputs are synthetic
random tokens and rewards are deterministic functions of group index.  Second,
the old and reference scores are produced by the same current model before the
step; the run uses \(\beta=0\), so it does not exercise an independent reference
policy or an active KL term.  The live policy expression is the unclipped ratio
term.  Because old and current scores coincide at the first step, the ratio is
one and clipping would be inactive even if present.  Third, 8,192 is the number
of response-input tokens.  The causal selector scores labels from response
index one onward, yielding at most 8,191 suffix targets per branch under the
evaluated implementation.  The run logs record the response length but do not
serialize this scored-token count.

\subsection{Lazy Gradient Pages and the Fixed Eight-H20 Envelope}

The detached prompt changes which gradients need physical storage.  Prompt KV
pages still participate in every response attention forward and therefore
contribute to the global \(dQ\).  They do not require prompt-side \(dK/dV\)
pages when the prompt boundary is treated as a stop-gradient state.  The page
manager consequently allocates gradient pages lazily when a backward actually
touches them, prunes detached-prefix K/V gradient pages, and retains K/V
gradient pages only for the response suffix. This changes allocation, not
attention: prefix state remains in the forward operator, while storage for
unreachable prompt gradients is never materialized for the detached prefix or
its paged KV history.

The physical-page fix was a separate turning point.  In the early 2M prefix
capture, logical slices retained their parent allocation and the run stopped
after 352 of 510 chunks at about 136.068~GB allocator memory.  Copying each
owned slice into a right-sized allocation completed all 510 chunks at a
58.655~GB prefix peak in the compact-page measurement.  The later full GRPO runs
include model, adapter, response-page, and kernel overhead; these prefix-only
numbers isolate the allocator failure that motivated physical compaction.

The live response activation is bounded by the response block, not the full
suffix: with a 2,048-token response block, reverse replay keeps
\(O(\text{response-block-size})\) activations while the 4,096-token prompt
chunks and page tables remain resident.  The evaluated configuration uses
page size 64, prefix blocks of 4,096, response blocks of 2,048, and 512-token
backward MLP microblocks.  Stage-1 capture/scoring may use a larger 4,096-token
MLP microblock, while response backward keeps 512 to bound live gradients.

\subsection{The Fixed Eight-H20 Execution Envelope}

Both runs use the fixed eight-NVIDIA-H20 budget, CP8, page size 64, a 4,096-token prefix
chunk, a 2,048-token response block, and 512-token FFN microblocks during
backward.  CUDA peak counters are reset after model, adapter, optimizer, and
input construction.  Reported memory is decimal GB from
\texttt{max\_memory\_allocated}, not reserved memory or process memory from
\texttt{nvidia-smi}.  Timings exclude model loading.

\paragraph{Group size two.}
The reported rank records 5,198.780 seconds end to end and a 97.503 GB peak.
The shared prefix consumes 4,656.225 seconds, about 89.6\% of the reported
time.  Across the eight per-rank records, total time ranges from 5,196.750 to
5,200.684 seconds; local optimizer calls take 0.155--0.200 seconds.

\paragraph{Group size eight.}
The reported run completes in 6,785.225 seconds at a 97.711 GB peak.  Prefix
capture takes 4,653.420 seconds.  Group zero completes at 4,931.993 seconds;
each of the seven additional serialized members adds a median 264.739 seconds.
After the first member, a typical branch spends about 32.0 seconds in old
scoring, 32.0 seconds in reference scoring, 31.2 seconds in suffix forward,
and 169.1--169.6 seconds in reverse replay.  Completion times across ranks range
from 6,783.943 to 6,785.225 seconds, while the corresponding local
optimizer-call timers range from 0.163 to 0.186 seconds.

The increase from \(G=2\) to \(G=8\) is therefore 1,586.445 seconds but only
0.208 GB (0.213\%) at the reported peak. Removing the shared prefix, the
post-prefix cost per member is 271.278 versus 266.476 seconds. Amortizing the
prefix reduces mean wall time per supplied response from 2,599.390 to
848.153 seconds. These derived values are the two reported group endpoints.
They establish complete single-step execution of the evaluated response-only
program: scoring, four-block policy backwards, gradient materialization, and
local optimizer calls all fit at a context length of 2,097,152. Separate 4.25M
measurements extend shared-prefix execution to eight consecutive \(G=8\) steps;
Section~\ref{tr:analysis} and Appendix~\ref{app:qwen-frontier} report those
multi-step results.

\subsection{CP8 Adapter Synchronization Audit}

The evaluated response-only procedure completes global CP8 forward composition,
response backwards, and local AdamW calls. A coherent replicated
K/V-adapter update additionally requires cross-rank gradient composition. Each
attention rank computes a local query-gradient contribution
\(dq_r\) and local
gradients for its K/V tokens.  The correct query gradient is
\begin{equation}
dq=\sum_{r=0}^{7}dq_r,
\label{eq:qwen-dq-reduce}
\end{equation}
and the custom backward performs this all-reduce.  In contrast, its \(dK_r\)
and \(dV_r\) remain rank local.  That is valid for sharded KV storage, but the
K/V projection LoRA weights are replicated.  Their correct parameter
contributions require composition across the disjoint token owners,
\begin{equation}
\nabla W_K=\sum_r \nabla W_K^{(r)},
\qquad
\nabla W_V=\sum_r \nabla W_V^{(r)},
\label{eq:qwen-kv-param-reduce}
\end{equation}
with K/V hidden-state contributions composed before differentiating earlier
replicated layers.

The evaluated runner initializes only NCCL: it has no DDP wrapper,
parameter-gradient reducer, or selective model-parallel composition required by
Equation~\ref{eq:qwen-kv-param-reduce}. Each rank owns an AdamW instance
~\citep{loshchilov2019adamw} and
steps locally.  All-reducing every completed gradient would still be wrong:
query contributions are already global but K/V contributions are not, so the
two paths require reductions at their distinct parameter-ownership boundaries
for replicated adapters.

The receipt focuses on execution events rather than gradient-norm, parameter-
delta, or post-step hash fields. The implementation completes the full evaluated
response-only transaction, including global full-attention response forwards,
response backwards, and local AdamW calls, within the fixed eight-H20 envelope
at 2,097,152 positions. The page-owner K/V terms are tracked as a distinct CP8
adapter-synchronization audit alongside the complete response-only execution
and prefix-reuse receipts.

\section{GLM: Paged MLA/DSA Replay within a 32-H20 MoE Budget}
\label{tr:glm-path}

The GLM path is architecturally more demanding.  Its long-range token
mixer is neither dense attention nor a prompt-length-independent recurrence.
Each decoder layer combines multi-head latent attention (MLA) with a dynamic
sparse-attention (DSA) index, while most feed-forward blocks are routed MoE
layers. Consequently, a useful prompt boundary must preserve both attention
representations, reproduce IndexShare across layers, and re-enter the
expert-parallel tail under autograd. The complete path traverses 78 layers,
composes DSA across CP ranks, and finalizes distributed gradients before the
exact-2M optimizer step.

\subsection{Layer Anatomy: MLA/DSA Before Dense or Routed FFNs}

The inspected configuration has hidden width $H=6{,}144$ and 78 decoder
layers.  Attention uses 64 MLA query heads, query and KV latent ranks 2,048 and
512, and a 64-dimensional positional channel.  The absorbed key retained by
the runtime therefore has width $512+64=576$.  The DSA indexer uses 32 heads
of dimension 128 and selects 2,048 key positions per response query.  These
dimensions follow the released GLM architecture and the inspected runtime
configuration~\citep{glm52config}.

Not every layer recomputes the index.  With index frequency four and skip
offset three, the zero-based index-computing set is
\begin{equation}
\mathcal{I}_{\mathrm{compute}}
=\{0,1,2\}\cup\{6,10,14,\ldots,74\},
\qquad |\mathcal{I}_{\mathrm{compute}}|=21.
\label{eq:glm-index-schedule}
\end{equation}
The other 57 layers are \mbox{IndexShare} consumers.  A compute layer publishes its
top-$k$ tensor to a carrier scoped to one decoder forward; consumers reuse it
instead of retaining prompt-length index state.  This implements cross-layer
index reuse, not a per-layer cache~\citep{bai2026indexcache}.

The FFN schedule is similarly nonuniform.  Layers 0--2 are dense gated FFNs.
The remaining 75 layers have 256 routed experts, top-8 routing, and one shared
expert.  With intermediate width 2,048, one routed expert contains, ignoring
biases,
\begin{equation}
B_{\mathrm{expert}}
=3\times6{,}144\times2{,}048
=37{,}748{,}736
\label{eq:glm-expert-weights}
\end{equation}
base weights.  A sparse layer contains approximately 9.66 billion routed
expert weights, but one token evaluates eight experts, or about 302 million
routed weights, plus the shared branch.  MoE reduces activated parameter count;
it does not remove router, permutation, all-to-all, expert-input, and combine
tensors from the layer execution~\citep{lepikhin2020gshard,fedus2021switch,
deepseekv2,deepseekv3}.

The 2M run uses rank-8 LoRA on eight target categories: the query and KV
down/up projections, the attention output projection, both FFN projections,
and the output head.  The FFN patterns match dense, routed-expert, and shared-
expert modules.  Base weights, embeddings, normalization parameters, DSA
indexer projections, and router parameters remain frozen.  The router's
\texttt{expert\_bias} is a non-parameter state buffer; its transition is not
recorded in the 2M run logs.

\subsection{TP1/CP32/EP32 Assigns Two Different Kinds of Ownership}

The reported 2M topology is TP1/CP32/EP32/ETP1/PP1 on 32 H20 GPUs.  TP1 leaves
each attention and dense projection structurally intact.  CP32 distributes the
prompt token axis and therefore the retained MLA/DSA state.  EP32 distributes
the 256 routed experts, nominally eight experts per rank.  CP and EP contain
the same workers in this run, but they are not interchangeable dimensions:
CP answers which rank owns a prompt position, whereas EP answers which rank
owns an expert parameter shard~\citep{liu2025moefolding}.

The distinction appears directly in the data movement.  During prefix capture,
one rank processes 65,536 prompt tokens.  Top-8 routing expands that shard to
$65{,}536\times8=524{,}288$ routed rows before load skew.  A single balanced
BF16 hidden buffer of shape $[524{,}288,6{,}144]$ occupies
\begin{equation}
524{,}288\times6{,}144\times2
=6{,}442{,}450{,}944\ \text{bytes}=6\ \text{GiB}.
\label{eq:glm-routed-buffer}
\end{equation}
This is a derived lower-level buffer size, not a measured whole-layer peak.
The live MoE path also needs dispatch metadata, output storage, inverse
permutations, shared-expert work, and LoRA intermediates.  They are released
during capture because autograd is disabled; a conventional beyond-2M graph cannot.
Successive full-sequence runs therefore moved from DSA scratch OOMs to
expert-LoRA and expert-output OOMs.

\subsection{Zigzag Context Pages within the 32-H20 Route}

Let the prompt length be $P=2{,}097{,}152$, the physical page size be
$S=64$, and the CP size be $C=32$.  Megatron's context-parallel layout
first partitions the global sequence into $2C=64$ contiguous chunks.  Each
chunk contains
\begin{equation}
N_{\mathrm{page}}=\frac{P}{S}=32{,}768\ \text{pages},
\qquad
N_{\mathrm{chunk}}=\frac{N_{\mathrm{page}}}{2C}=512\ \text{pages}.
\label{eq:glm-page-count}
\end{equation}
Rank $r\in\{0,\ldots,31\}$ owns chunk $r$ and its mirrored chunk
$63-r$.  Its ordered global page set is therefore
\begin{equation}
\mathcal{P}_r=
\{512r,\ldots,512r+511\}
\mathbin{\Vert}
\{512(63-r),\ldots,512(63-r)+511\},
\label{eq:glm-tr-zigzag-pages}
\end{equation}
where $\Vert$ denotes concatenation in the local tensor order.  Every rank
stores 1,024 pages, or 65,536 tokens.  For example, rank 0 stores pages
0--511 followed by 32,256--32,767; rank 31 stores 15,872--16,383 followed by
16,384--16,895.  The 2M run log records all 32 page samples and agrees with
this formula.

The page manager copies into right-sized CPU pages, not views of larger GPU
capture tensors.  PagedAttention manages serving KV through fixed-size physical
blocks and a logical-to-physical block table~\citep{kwon2023efficient}.  In our
training page manager, memory is bounded only when allocator ownership matches
page-table ownership.  Global page IDs stay attached to local tensor order, so
DSA masks and selections retain their original global coordinates throughout
capture and replay.

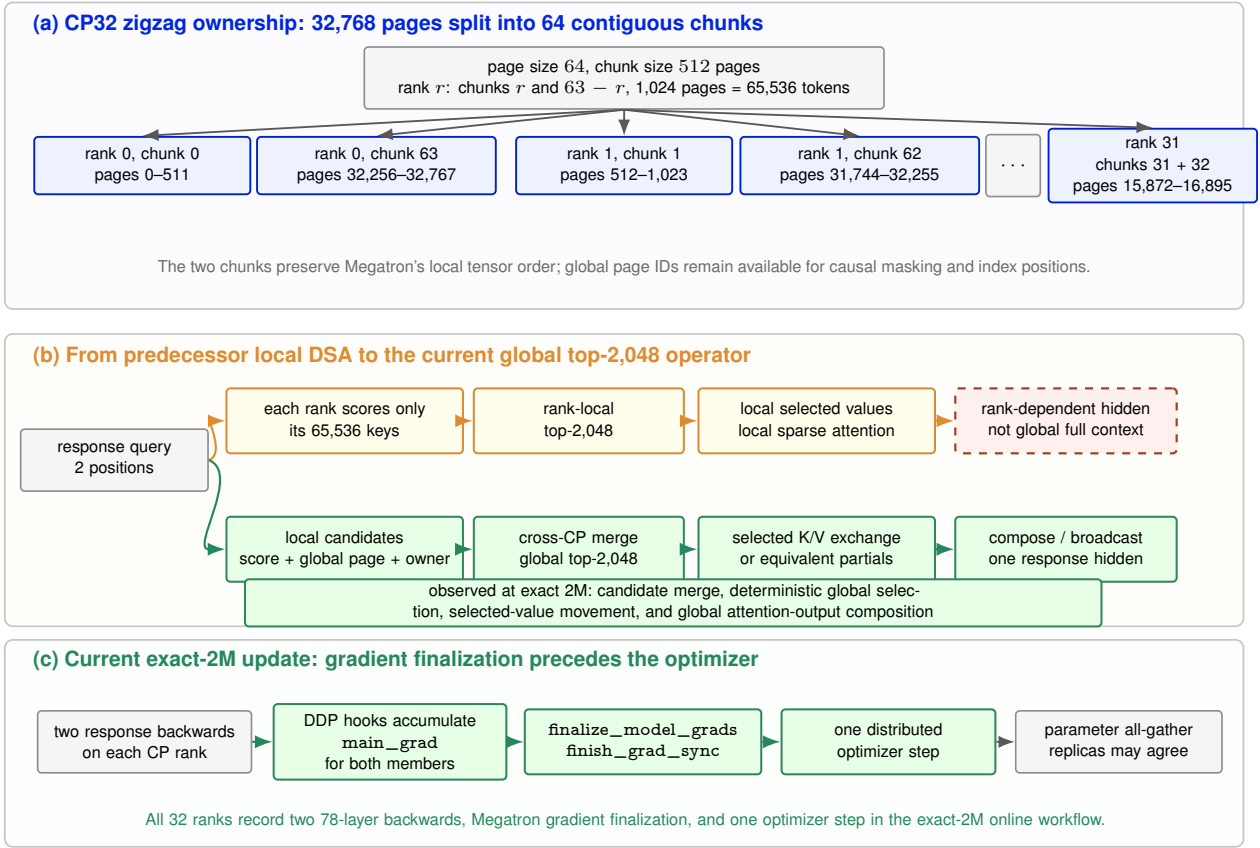
\begin{figure}[t]
\centering
\resizebox{\textwidth}{!}{%
\begin{tikzpicture}[
  font=\sffamily\scriptsize,
  panel/.style={draw=mindlabfg!24, fill=white, rounded corners=3pt,
    line width=0.65pt},
  chunk/.style={draw=mindlabblue!78!black, fill=mindlabbluepale!50,
    rounded corners=1.5pt, line width=0.72pt, align=center,
    minimum height=0.76cm, inner sep=3pt},
  local/.style={draw=BurntOrange!86!black, fill=yellow!9,
    rounded corners=1.5pt, line width=0.76pt, align=center,
    minimum height=0.86cm, inner sep=4pt},
  required/.style={draw=ForestGreen!72!black, fill=green!10,
    rounded corners=1.5pt, line width=0.76pt, align=center,
    minimum height=0.86cm, inner sep=4pt},
  missing/.style={draw=BrickRed!82!black, fill=red!6, dashed,
    rounded corners=1.5pt, line width=0.84pt, align=center,
    minimum height=0.86cm, inner sep=4pt},
  neutral/.style={draw=mindlabfg!48, fill=mindlabfg!4,
    rounded corners=1.5pt, line width=0.62pt, align=center,
    minimum height=0.82cm, inner sep=4pt},
  flow/.style={-Latex, draw=mindlabfg!70, line width=0.70pt},
  currentflow/.style={-Latex, draw=BurntOrange!88!black,
    line width=0.80pt},
  reqflow/.style={-Latex, draw=ForestGreen!75!black,
    line width=0.82pt},
  missflow/.style={-Latex, draw=BrickRed!82!black, dashed,
    line width=0.84pt}
]

\filldraw[panel, fill=mindlabbluepale!10]
  (-0.20,6.80) rectangle (16.20,10.86);
\filldraw[panel, fill=yellow!2]
  (-0.20,2.60) rectangle (16.20,6.47);
\filldraw[panel, fill=green!1]
  (-0.20,-0.32) rectangle (16.20,2.42);

\node[anchor=west, font=\sffamily\bfseries\small,
  text=mindlabblue!88!black] at (0.05,10.56)
  {(a) CP32 zigzag ownership: 32,768 pages split into 64 contiguous chunks};

\node[neutral, text width=6.60cm] (formula) at (8.00,9.86)
  {page size $64$, chunk size $512$ pages\\
   rank $r$: chunks $r$ and $63-r$, 1,024 pages = 65,536 tokens};

\node[chunk, text width=2.65cm] (r0a) at (1.62,8.70)
  {rank 0, chunk 0\\pages 0--511};
\node[chunk, text width=2.95cm] (r0b) at (4.72,8.70)
  {rank 0, chunk 63\\pages 32,256--32,767};
\node[chunk, text width=2.65cm] (r1a) at (8.00,8.70)
  {rank 1, chunk 1\\pages 512--1,023};
\node[chunk, text width=2.95cm] (r1b) at (11.12,8.70)
  {rank 1, chunk 62\\pages 31,744--32,255};
\node[neutral, minimum width=0.72cm] (dots) at (13.15,8.70) {\(\cdots\)};
\node[chunk, text width=2.55cm] (r31) at (15.00,8.70)
  {rank 31\\chunks 31 + 32\\pages 15,872--16,895};

\draw[flow] (formula.south) -- (r0a.north);
\draw[flow] (formula.south) -- (r0b.north);
\draw[flow] (formula.south) -- (r1a.north);
\draw[flow] (formula.south) -- (r1b.north);
\draw[flow] (formula.south) -- (r31.north);

\node[font=\sffamily\scriptsize, text=mindlabfg!62, align=center]
  at (8.00,7.35)
  {The two chunks preserve Megatron's local tensor order; global page IDs remain available for causal masking and index positions.};

\node[anchor=west, font=\sffamily\bfseries\small,
  text=BurntOrange!88!black] at (0.05,6.17)
  {(b) From predecessor local DSA to the current global top-2,048 operator};

\node[neutral, text width=2.20cm] (query) at (1.25,4.80)
  {response query\\2 positions};
\node[local, text width=2.85cm] (localscore) at (4.30,5.32)
  {each rank scores only\\its 65,536 keys};
\node[local, text width=2.50cm] (localtopk) at (7.40,5.32)
  {rank-local\\top-2,048};
\node[local, text width=2.85cm] (localattn) at (10.55,5.32)
  {local selected values\\local sparse attention};
\node[missing, text width=2.65cm] (localout) at (13.85,5.32)
  {rank-dependent hidden\\not global full context};

\draw[currentflow] (query.east) to[out=20,in=180] (localscore.west);
\draw[currentflow] (localscore) -- (localtopk);
\draw[currentflow] (localtopk) -- (localattn);
\draw[currentflow] (localattn) -- (localout);

\node[required, text width=2.85cm] (candidates) at (4.30,3.62)
  {local candidates\\score + global page + owner};
\node[required, text width=2.50cm] (globaltopk) at (7.40,3.62)
  {cross-CP merge\\global top-2,048};
\node[required, text width=2.85cm] (exchange) at (10.55,3.62)
  {selected K/V exchange\\or equivalent partials};
\node[required, text width=2.65cm] (compose) at (13.85,3.62)
  {compose / broadcast\\one response hidden};

\draw[reqflow] (query.east) to[out=-22,in=180] (candidates.west);
\draw[reqflow] (candidates) -- (globaltopk);
\draw[reqflow] (globaltopk) -- (exchange);
\draw[reqflow] (exchange) -- (compose);

\node[required, font=\sffamily\scriptsize, minimum height=0cm, inner sep=2pt,
  text width=11.20cm] at (8.65,2.91)
  {observed at exact 2M: candidate merge, deterministic global selection,
   selected-value movement, and global attention-output composition};

\node[anchor=west, font=\sffamily\bfseries\small,
  text=ForestGreen!75!black] at (0.05,2.14)
  {(c) Current exact-2M update: gradient finalization precedes the optimizer};

\node[neutral, text width=2.55cm] (localbwd) at (1.65,1.07)
  {two response backwards\\on each CP rank};
\node[required, text width=2.80cm] (maingrad) at (4.90,1.07)
  {DDP hooks accumulate\\\texttt{main\_grad}\\for both members};
\node[required, text width=2.85cm] (finalize) at (8.25,1.07)
  {\texttt{finalize\_model\_grads}\\\texttt{finish\_grad\_sync}};
\node[required, text width=2.55cm] (diststep) at (11.50,1.07)
  {one distributed\\optimizer step};
\node[neutral, text width=2.45cm] (allgather) at (14.55,1.07)
  {parameter all-gather\\replicas may agree};

\draw[reqflow] (localbwd) -- (maingrad);
\draw[reqflow] (maingrad) -- (finalize);
\draw[reqflow] (finalize) -- (diststep);
\draw[flow] (diststep) -- (allgather);

\node[font=\sffamily\scriptsize, text=ForestGreen!75!black, align=center,
  text width=15.20cm]
  at (8.00,0.02)
  {All 32 ranks record two 78-layer backwards, Megatron gradient finalization, and one optimizer step in the exact-2M online workflow.};

\end{tikzpicture}%
}
\caption{\textbf{GLM CP32 ownership: predecessor failure and current repair.}
Megatron zigzag partitioning gives each rank two mirrored 512-page chunks.
The predecessor DSA fallback selected top-2,048 independently inside each local
65,536-token shard. The current exact-2M path exchanges global candidates and
selected values, composes one response hidden state, and calls Megatron
gradient finalization before the optimizer, closing the full exact-2M online
GRPO workflow.}
\label{fig:glm-cp-dsa}
\end{figure}

\subsection{Stored Tensor Layout and Derived Residency}

Every layer stores one absorbed MLA page component.  Its runtime key is
\texttt{mla\_latent\_kv\_pages}.  The 21 index-computing layers additionally
store DSA index-key pages under \texttt{dsa\_indexer\_key\_pages}; an
\mbox{IndexShare} consumer stores no duplicate index key.  Earlier live shape
measurements record both components as BF16 and yield the following layouts:
\begin{align}
\text{MLA page} &: [1,64,1,576],
&\text{local materialization} &: [1,65{,}536,1,576],\\
\text{DSA key page} &: [1,64,128],
&\text{local materialization} &: [1,65{,}536,128].
\label{eq:glm-page-shapes}
\end{align}
The 2M summary records component names, page counts, and CPU placement but
omits dtype and shape; we therefore use the earlier live trace for the layouts
above.

The corresponding retained-state arithmetic follows directly from these
shapes.  Per rank, one layer's MLA pages occupy
\begin{equation}
1{,}024\times64\times576\times2=72\ \text{MiB},
\end{equation}
and one compute layer's index pages occupy
\begin{equation}
1{,}024\times64\times128\times2=16\ \text{MiB}.
\end{equation}
Thus the CPU-resident prompt state is
\begin{align}
B_{\mathrm{CPU/rank}}
&=78\times72\ \text{MiB}+21\times16\ \text{MiB}\\
&=5{,}952\ \text{MiB}=5.8125\ \text{GiB},
\label{eq:glm-prefix-bytes}
\end{align}
or 186 GiB across 32 ranks.  Layerwise staging uses 72 MiB for an IndexShare
layer and 88 MiB for an index-computing layer.  These payloads exclude response
activations and kernel workspaces, so they are not whole-step peaks.  The 2M
run log reads the retained CUDA peak after prefix capture, before policy replay.

\subsection{Predecessor Layer Trace: State, Index, Attention, and MoE}

The 2026-07-13 predecessor response path consumes two scored positions; its decoder input has shape
$[2,1,6{,}144]$.  The runner materializes that rank's
65,536 prompt positions, recomputes the response-side projections, and forms a
65,538-position local attention problem.  An index-computing layer evaluates
the response queries against its local DSA key and publishes indices with
shape $[1,2,2{,}048]$.  An IndexShare layer consumes the previously published
indices.  Both paths run the runtime's unfused absorbed sparse attention,
the live output projection and bias/dropout/add path, and the dense-or-MoE
\texttt{\_forward\_mlp} tail.

For a routed layer, each local response row is expanded to eight assignments,
permuted by expert owner, exchanged through the EP all-to-all, evaluated by
the owner's expert FC1/FC2 LoRA path, and inverse-combined with router
probabilities.  The shared expert branch is evaluated in parallel and added to
the routed result.  All 75 MoE tails therefore execute under the native EP
autograd graph rather than an attention-only or synthetic FFN surrogate.  The
trace omits routing counts and all-to-all measurements.

\subsection{Whole-Layer Checkpointing and the Recorded Backward}

Checkpoint placement controls which intermediates survive and which operators
are recomputed~\citep{chen2016training,korthikanti2022recomputation}.
Attention-only checkpointing left the routed tail live, so the policy path
checkpoints whole decoder layers reentrantly with RNG
preservation.  Forward retains only the short
$[2,1,6{,}144]$ input and metadata.  Reverse re-enters layers 77 through 0,
restages CPU pages, recomputes sparse attention plus the dense/MoE tail, and
releases workspace; saved sparse-attention tensors may be offloaded to CPU.

The trace shapes make this lifecycle observable.  A representative policy
trace records:
\begin{itemize}[leftmargin=1.7em]
  \item decoder input and every layer output: $[2,1,6{,}144]$ BF16;
  \item local top-$k$: $[1,2,2{,}048]$, with 65,536 prefix positions;
  \item logits: $[1,2,154{,}880]$, followed by a BF16 logits-gradient event;
  \item layer and attention-projection gradients: $[2,1,6{,}144]$ BF16;
  \item sparse-attention output gradients: $[2,1,16{,}384]$ BF16.
\end{itemize}

In that predecessor run, policy and old-policy traces contain 396 and 160 events, respectively. Two
members, two phases, and 32 ranks produce 128 files with 35,584 JSONL events.
Each policy rank records 78 checkpointed layer ends and 78 each of layer,
attention-projection, and sparse-attention backwards. These traces show that
the architecture-shaped path completed twice; they contain no parameter-
gradient tensors or cross-rank reductions. The current exact-2M audit instead
contains 64 trace files and 25,536 events.

\begin{figure}[t]
\centering
\pgfdeclarelayer{checkpointlayer}
\pgfsetlayers{background,checkpointlayer,main}
\resizebox{0.90\textwidth}{!}{%
\begin{tikzpicture}[
  font=\sffamily\scriptsize,
  panel/.style={draw=mindlabfg!24, fill=white, rounded corners=3pt,
    line width=0.65pt},
  cpu/.style={draw=mindlabfg!55, fill=mindlabfg!5, rounded corners=1.5pt,
    line width=0.65pt, align=center, minimum height=0.88cm, inner sep=4pt},
  attn/.style={draw=mindlabblue!80!black, fill=mindlabbluepale!52,
    rounded corners=1.5pt, line width=0.75pt, align=center,
    minimum height=0.92cm, inner sep=4pt},
  share/.style={attn, fill=mindlabbluepale!22, dashed},
  dense/.style={draw=ForestGreen!70!black, fill=green!9,
    rounded corners=1.5pt, line width=0.72pt, align=center,
    minimum height=0.92cm, inner sep=4pt},
  moe/.style={draw=ForestGreen!78!black, fill=green!17,
    rounded corners=1.5pt, line width=0.78pt, align=center,
    minimum height=0.92cm, inner sep=4pt},
  checkpoint/.style={draw=BurntOrange!88!black, dashed, rounded corners=3pt,
    line width=0.9pt, inner sep=7pt},
  trace/.style={draw=BurntOrange!76!black, fill=yellow!8,
    rounded corners=1.5pt, line width=0.72pt, align=center,
    minimum height=0.82cm, inner sep=4pt},
  flow/.style={-Latex, draw=mindlabfg!70, line width=0.72pt},
  cpuflow/.style={-Latex, draw=mindlabblue!80!black, line width=0.78pt,
    preaction={draw=white, line width=1.8pt}},
  backflow/.style={-Latex, draw=BurntOrange!88!black, dashed,
    line width=0.82pt}
]

\begin{pgfonlayer}{background}
  \filldraw[panel, fill=mindlabbluepale!10]
    (-0.20,2.92) rectangle (16.75,8.72);
  \filldraw[panel, fill=yellow!2]
    (-0.20,-0.65) rectangle (16.75,2.64);
\end{pgfonlayer}

\node[anchor=west, font=\sffamily\bfseries\small,
  text=mindlabblue!88!black, fill=mindlabbluepale!10, inner sep=1.2pt]
  at (0.05,8.58)
  {(a) One rank, one response layer: CPU prompt state enters the native layer tail};

\node[cpu, text width=3.10cm] (mlapages) at (1.70,7.38)
  {all 78 layers\\MLA BF16 pages\\\([1,64,1,576]\)\\1,024 pages / rank};
\node[cpu, text width=3.10cm] (dsapages) at (1.70,5.72)
  {21 compute layers only\\DSA-key BF16 pages\\\([1,64,128]\)\\1,024 pages / rank};
\node[attn, text width=3.40cm] (stage) at (5.55,6.56)
  {stage one layer to GPU\\MLA \([1,65{,}536,1,576]\)\\DSA \([1,65{,}536,128]\) if needed};

\node[attn, text width=2.60cm] (response) at (5.55,8.00)
  {response hidden\\\([4,1,6{,}144]\)};
\node[attn, text width=2.75cm] (compute) at (9.00,7.45)
  {index-compute layer\\local candidates\\+ global positions};
\node[share, text width=2.75cm] (indexshare) at (9.00,5.82)
  {IndexShare layer\\consume producer's\\global top-\(k\)};
\node[attn, text width=2.45cm] (sparse) at (12.20,6.56)
  {global top-2,048\\selected-value movement\\CP32 composition};
\node[attn, text width=1.90cm] (proj) at (15.18,6.56)
  {output projection\\bias/dropout/add};

\draw[flow] (response.south) -- (stage.north);
\draw[flow] (stage.north east)
  to[out=-10,in=190,looseness=0.78] (compute.south west);
\draw[flow] (stage.south east)
  to[out=10,in=170,looseness=0.78] (indexshare.north west);
\draw[flow] (compute.south east)
  to[out=10,in=170,looseness=0.78] (sparse.north west);
\draw[flow] (indexshare.north east)
  to[out=-10,in=190,looseness=0.78] (sparse.south west);
\draw[flow] (sparse) -- (proj);

\node[dense, text width=2.55cm] (dense) at (6.35,3.98)
  {layers 0--2\\dense gated FFN\\FC1 / FC2 LoRA};
\node[moe, text width=2.85cm] (router) at (9.55,3.98)
  {layers 3--77\\top-8 router + shared\\permute by expert owner};
\node[moe, text width=2.20cm] (a2a) at (12.85,3.98)
  {EP32 all-to-all\\8 experts / rank\\expert FC1 / FC2};
\node[dense, text width=1.45cm] (out) at (15.28,3.98)
  {inverse\\combine\\\([2,1,6{,}144]\)};

\draw[flow] (proj.south west)
  .. controls (13.90,5.15) and (8.45,5.05) .. (dense.north east);
\draw[flow] (proj.south east)
  .. controls (15.95,4.88) and (11.55,4.78) .. (router.north east);
\draw[flow] (router) -- (a2a);
\draw[flow] (a2a.east) -- (out.west);
\draw[flow] (dense.south)
  .. controls (6.35,3.14) and (15.28,3.14) .. (out.south);

\node[anchor=west, font=\sffamily\bfseries\small,
  text=BurntOrange!88!black, fill=yellow!2, inner sep=1.2pt] at (0.05,2.40)
  {(b) Reverse replay: restage, recompute, release};

\node[trace, text width=3.15cm] (logits) at (1.90,0.65)
  {logits \([1,4,154{,}880]\)\\BF16 logits-gradient event};
\node[trace, text width=3.20cm] (reverse) at (5.45,0.65)
  {layers 77 \(\rightarrow\) 0\\reentrant checkpoint\\RNG state preserved};
\node[trace, text width=3.35cm] (grads) at (9.20,0.65)
  {layer / projection grad\\\([4,1,6{,}144]\) BF16\\sparse grad \([4,1,16{,}384]\)};
\node[cpu, text width=3.10cm] (release) at (13.00,0.65)
  {release staged layer state\\retain CPU prompt pages\\next layer or branch};

\draw[backflow] (logits) -- (reverse);
\draw[backflow] (reverse) -- (grads);
\draw[backflow] (grads) -- (release);

\begin{pgfonlayer}{checkpointlayer}
  \node[checkpoint,
    fit=(stage)(compute)(indexshare)(sparse)(proj)(dense)(router)(a2a)(out)(reverse)(grads)(release)]
    (ckpt) {};
\end{pgfonlayer}
\node[font=\sffamily\bfseries\scriptsize,
  text=BurntOrange!88!black, fill=mindlabbluepale!10, inner sep=1.2pt]
  at (13.30,8.12) {whole decoder layer: reentrant checkpoint boundary};

\draw[cpuflow] (mlapages.south east)
  to[bend left=7] (stage.north west);
\draw[cpuflow] (dsapages.north east)
  to[bend right=7] (stage.south west);

\node[font=\sffamily\scriptsize, text=mindlabfg!62, align=center]
  at (8.00,-0.40)
  {Derived page payload: 72 MiB for an IndexShare layer, 88 MiB for an index-compute layer; response and kernel workspaces are additional.};

\end{tikzpicture}%
}
\caption{\textbf{GLM resident layer replay and whole-layer checkpointing.}
CPU pages retain CP-sharded prompt state. A compute layer stages its local MLA
and DSA-key pages, then global CP32 composition selects top-2,048 positions and
moves their values; an IndexShare layer consumes the matching per-forward
global selection. The live
attention projection and dense or EP32 MoE tail execute inside one checkpoint
boundary.  Backward traverses layers in reverse, restages and recomputes one
layer, emits the recorded tensor shapes, and releases its workspace.  Page
shapes and byte counts are derived from those shapes; they are not whole-step peak
measurements.}
\label{fig:glm-layer-replay}
\end{figure}
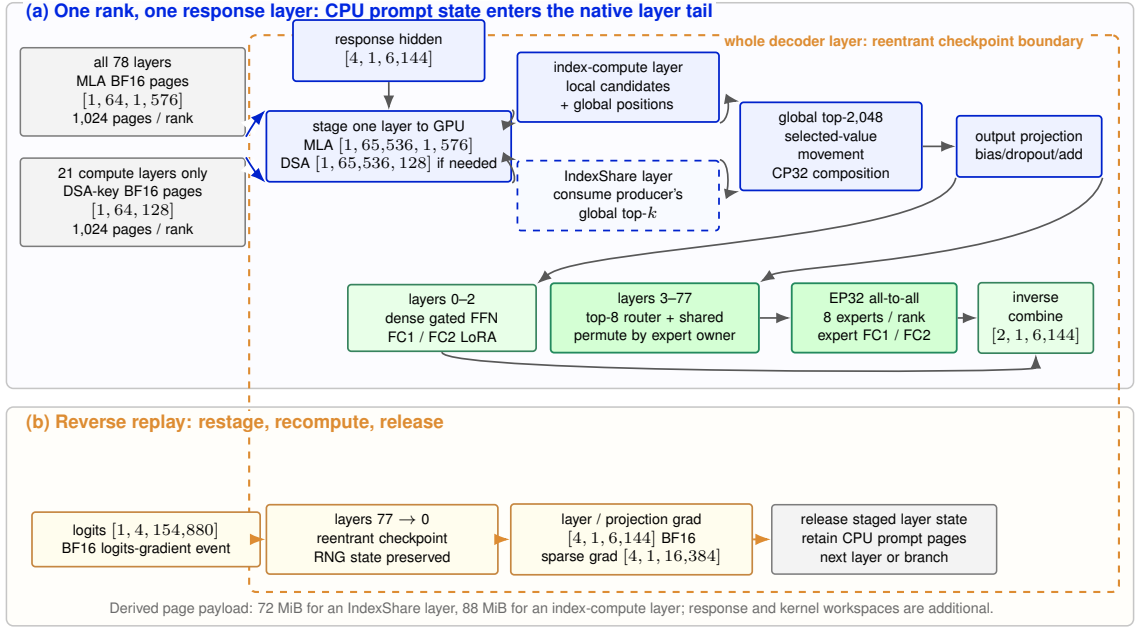

\subsection{Global DSA, Position Extension, and Temporal Topology Reuse}

For each response query, every CP rank computes local DSA candidate scores with
global positions. The adapter exchanges those candidates, performs a
deterministic global top-$k$, moves the selected MLA values from their owners,
and composes the global response output. The 57 IndexShare layers consume the
selection published by their source layer rather than searching again. This
candidate/value movement is observed on all rank groups in the exact-2M online
transaction audit.

The released checkpoint declares 1,048,576 native positions. The online run
uses an explicit YaRN configuration with factor 2.000015 and
\texttt{max\_position\_embeddings=2097168}. Both the Tinker-managed vLLM sampler
and the Megatron replay receive the same RoPE parameters, so the experimental
position extension is shared across rollout and training.

Rollout and training time-share the same 32 H20 GPUs. The runtime releases the
Megatron TP1/CP32/EP32 actor, samples with the Tinker-managed vLLM sampler at
TP8/PP4, releases the sampler, and restores Megatron at
TP1/CP32/EP32/ETP1/PP1. The policy-LoRA hash matches across the handoff.

The sampler receives a vLLM-compatible projection of the exported rank-8 policy
LoRA. GLM-5.2's vLLM interface does not expose \texttt{lm\_head} as a supported
LoRA module, so the projection removes only its LoRA tensors after verifying
that \(B\) is elementwise zero. Because the omitted update is \(BA\), it has
zero contribution to the rollout logits; all remaining adapter tensors are
loaded unchanged. The exact-2M run loads this projected adapter before sampling
begins in the active vLLM worker.

Each replay begins by restoring the per-layer prefix boundary and ends by
appending or popping response pages after metadata, storage-pointer, and
version checks. A temporary IndexShare holder or expert permutation therefore
cannot overwrite the resident prompt state seen by the next group member.
Response rows have one designated CP owner; the owner-broadcast path makes the
response hidden state and its autograd contribution available exactly once
before the EP router. Native Megatron routing expands each live row to top-8
assignments, dispatches them through EP32, and inverse-combines the expert
outputs.

The successful run disables parameter gradients for all 1,394 parameters during
prefix capture, stores the retained prefix on CPU, and replaces 98 of 99 RoPE
owners with one shared cache instance. Whole-layer checkpointing bounds the
response graph. The implementation also contains chunked output-head paths, but
the exact-2M run records \texttt{chunk\_token\_count=null}; we therefore do not
attribute this result to output-head or prefix chunking.

After both response backward passes, the implementation applies
\begin{equation}
\texttt{finalize\_model\_grads}
\;\longrightarrow\;
\texttt{optimizer.step()}
\;\longrightarrow\;
\texttt{zero\_grad}.
\label{eq:glm-finalize-contract}
\end{equation}
This order composes CP/EP-owned gradients according to Megatron's ownership
rules before one optimizer update. All 32 ranks record both backwards,
finalization, and the optimizer step in the exact-2M transaction.

\subsection{Predecessor CP-Local 2M Diagnostic}

For response query $q_t$, rank $r$ in the archived 2M run computes index
scores only
against its page set $\mathcal{P}_r$, then selects
\begin{equation}
I_{r,t}=\operatorname{TopK}_{j\in\mathcal{P}_r}
\left[\sum_{h=1}^{32}w_{t,h}\,
\phi\!\left(q_{t,h}^{\top}k_j\right)\right],
\qquad |I_{r,t}|=2{,}048,
\label{eq:glm-local-dsa}
\end{equation}
with causal masking in global page coordinates.  Selected values yield
rank-local hidden states; CP ranks merge neither candidates nor outputs.
Disjoint prompt ownership makes $I_{r,t}$ rank-specific.  Each layer therefore
realizes local top-2,048 over 65,536 tokens, not model-global top-2,048 over the
full 2,097,152-position prompt.

A faithful distributed operator first needs sufficient local candidates with
global scores, positions, and owner IDs; a deterministic cross-rank merge must
then compute
\begin{equation}
I_t=\operatorname{TopK}_{j\in\bigcup_r\mathcal{P}_r}s_{t,j},
\qquad |I_t|=2{,}048.
\label{eq:glm-global-dsa}
\end{equation}
The selected K/V rows must be exchanged from their owners, or their attention
numerators and normalizers must be composed with an equivalent distributed
operator.  Finally, all ranks that continue the replicated decoder must agree
on the composed response hidden state.  Candidate merge, selected-value
exchange, and output composition are all absent from the archived 2M run, which
therefore evaluates CP-local rather than model-global DSA over the full prompt
for every response query.

\subsection{Predecessor Gradient-Finalization Path}

A predecessor implementation used a separate cross-rank gradient-reduction path.
LoRA transformation is
applied before Megatron DDP construction, so trainable adapter parameters are
registered in DDP gradient buffers.  The inspected DDP configuration has
\texttt{overlap\_grad\_reduce=false}.  Its parameter hooks therefore add each
local autograd gradient into \texttt{param.main\_grad}, but they do not launch
an all-reduce or reduce-scatter during backward.  In the conventional Megatron
schedule, the post-backward \texttt{finalize\_model\_grads} call invokes
\texttt{finish\_grad\_sync} on every model chunk.  This is the required
DP$\times$CP gradient reduction for non-expert replicas.

The predecessor resident path does not call that schedule. It runs two local
\texttt{loss.backward()} calls and then invokes
\texttt{engine.optimizer\_step()} directly.  The concrete engine method calls
the distributed optimizer's \texttt{step()} without first finalizing model
gradients.  The distributed optimizer then copies the shard it owns from the
unreduced full gradient buffer, updates that parameter shard, and all-gathers
the updated parameter shards.  In symbols, the required non-expert gradient is
\begin{equation}
g^{\star}=\sum_{r=0}^{31}g_r,
\qquad
\text{but the observed path supplies shard owner }r(s)
\text{ with }\left.g_{r(s)}\right|_s.
\label{eq:glm-missing-grad-finalize}
\end{equation}
The subsequent parameter all-gather can make complete parameter replicas look
mutually consistent even though different parameter shards were updated from
different local objectives.  Parameter equality after all-gather would not
repair the missing gradient sum.  In this run the local gradients are expected
to differ because each rank's DSA forward sees a different 65,536-token shard.

The affected class is the CP-replicated, non-expert LoRA state: attention
projections, the three dense FFNs, the output head, and any other adapter placed
in the non-expert DP$\times$CP buffer.  Routed experts require a different
interpretation.  With world size 32, EP32, ETP1, and PP1, expert data-parallel
size is one.  Routed-expert adapter weights are marked
\texttt{allreduce=false}, are uniquely owned, and receive token contributions
through the differentiable EP dispatch/combine path; no replica average is
required for those unique expert shards.  Shared experts have an additional
PEFT EP gradient hook in the inspected runtime, but the run trace does not
record per-module hook coverage or hashes, so the run does not determine
whether their gradients were synchronized.  The router \texttt{expert\_bias}
state is also unrecorded.

\begingroup
\looseness=-1
The predecessor 2M logs contain no CP finalization event, gradient norm,
parameter delta, pre/post adapter hash, or next-forward equality test. They
therefore establish neither a CP32-reduced GRPO gradient nor a synchronized
model-global GLM update across the complete adapter state.
\par
\endgroup

\paragraph{What the predecessor established.}
The predecessor 2M run demonstrates fixed-budget full-model resident replay and
backward capacity: all 32 workers executed the 78-layer MLA/IndexShare/MoE path
twice, completed two local backward passes, and invoked the distributed
optimizer. Its CP-local DSA and unfinalized non-expert gradients define the
historical receipt's measurement scope; the current exact-2M transaction uses
the global operator and finalized update described below.

\subsection{Complete Exact-2M Online Training Workflow}

The current run completes the finalized path. A Tinker-managed vLLM
sampler, loaded with the vLLM-compatible rank-8 policy-LoRA projection, produces
two exact-2M conditioned completions from one DAPO-MATH prompt,
with rewards \([-1,+1]\) and stored old log-probabilities. Global DSA supplies
the prompt-wide response operator, and \texttt{finalize\_model\_grads} precedes
one optimizer step on every rank. The run records 64 rank/group trace files and
25,536 events without non-finite values or execution errors.

\FloatBarrier

\Needspace{0.12\textheight}
\section{Making the GLM GRPO Path Fit a Fixed 32-H20 Budget}
\label{sec:experiments}

\begin{figure}[!t]
\centering
\resizebox{\textwidth}{!}{%
\begin{tikzpicture}[
  font=\sffamily\scriptsize,
  stage/.style={draw=mindlabfg!65, line width=0.6pt, rounded corners=1.5pt,
    minimum width=2.35cm, minimum height=1.45cm, text width=2.08cm,
    align=center, inner sep=4pt},
  fail/.style={stage, fill=red!6, draw=BrickRed!75!black},
  build/.style={stage, fill=mindlabbluepale!35, draw=mindlabblue!78!black},
  pass/.style={stage, fill=green!8, draw=ForestGreen!75!black},
  arrow/.style={-Latex, line width=0.7pt, draw=mindlabfg!70},
  note/.style={align=center, text width=2.15cm, text=mindlabmuted,
    font=\sffamily\small}
]
\node[fail] (s1) at (0,0) {\textbf{I. Full graph}\\32K passes\\2.097M OOM migrates};
\node[build] (s2) at (2.8,0) {\textbf{II. Prefix state}\\128K $\rightarrow$ 2.097M\\all 78, no grad};
\node[build] (s3) at (5.6,0) {\textbf{III. Layer 0}\\1M and 2.097M\\backward tests};
\node[build] (s4) at (8.4,0) {\textbf{IV. All layers}\\32K then 64K\\IndexShare integration};
\node[build] (s5) at (11.2,0) {\textbf{V. Ownership}\\CPU pages\\CP32 + EP32};
\node[build] (s6) at (14.0,0) {\textbf{VI--VII. Predecessor}\\2.097M, \(G=1\to2\)\\CP-local execution};
\node[pass] (s7) at (16.8,0) {\textbf{VIII. Online}\\\mbox{exact-2M} \(G=2\)\\global DSA + finalize};

\draw[arrow] (s1) -- (s2);
\draw[arrow] (s2) -- (s3);
\draw[arrow] (s3) -- (s4);
\draw[arrow] (s4) -- (s5);
\draw[arrow] (s5) -- (s6);
\draw[arrow] (s6) -- (s7);

\node[note] at (0,-1.65) {DSA scratch $\rightarrow$ expert LoRA $\rightarrow$ MoE concatenate};
\node[note] at (2.8,-1.65) {Prompt storage fits;\\replay follows};
\node[note] at (5.6,-1.65) {One restored layer completes};
\node[note] at (8.4,-1.65) {Checkpoint complete layer};
\node[note] at (11.2,-1.65) {Bound GPU residency};
\node[note] at (14.0,-1.65) {Predecessor all layers;\\current global path follows};
\node[note] at (16.8,-1.65) {Two 78-layer backwards;\\one all-rank step};

\draw[decorate,decoration={brace,amplitude=4pt},draw=mindlabblue]
  (-1.15,1.25) -- (9.55,1.25)
  node[midway,above=4pt,text=mindlabblue,font=\sffamily\small]
  {separate the failing dependency classes};
\draw[decorate,decoration={brace,amplitude=4pt},draw=ForestGreen!75!black]
  (10.05,1.25) -- (17.95,1.25)
  node[midway,above=4pt,text=ForestGreen!75!black,font=\sffamily\small]
  {assemble the target-scale transaction};
\end{tikzpicture}%
}
\caption{\textbf{The staged route to an exact-2M online GLM update in 32 H20s.} Each stage adds one
dependency class only after the narrower test passes. Prefix capacity,
differentiable replay, all-layer integration, grouped ordering, global DSA,
gradient finalization, and optimizer application form the measured stages.}
\label{fig:scaling-timeline}
\end{figure}
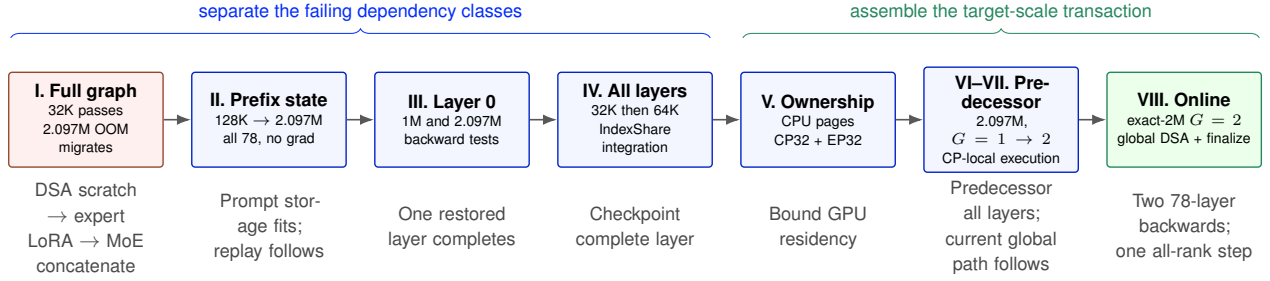

The final grouped run fixes the 32-H20 allocation while separating TP8/PP4
rollout from TP1/\allowbreak CP32/\allowbreak EP32/\allowbreak ETP1/\allowbreak PP1
training. Each stage isolates one dependency
class: memory, prefix state, replay, architecture state, placement, or group
ordering. Table~\ref{tab:optimization-summary} summarizes the progression and
Figure~\ref{fig:scaling-timeline} shows its bottlenecks.

\begin{table}[!t]
\centering
\caption{The progression from a conventional GLM full-sequence path to one
complete exact-2M online \(G=2\) transaction.}
\label{tab:optimization-summary}
\scriptsize
\setlength{\tabcolsep}{3pt}
\begin{tabular}{@{}C{0.15\textwidth}C{0.23\textwidth}C{0.31\textwidth}C{0.23\textwidth}@{}}
\toprule
Stage & Limiting observation & Modification & Observed result \\
\midrule
\rowcolor{red!6}
Full sequence & 32K passes; 2.097M OOM moves from DSA scratch to expert LoRA and MoE concatenate & Stop retaining a full-context autograd graph & Establishes the need for a detached prompt boundary \\
Prefix capture & No durable GLM prompt representation & Capture MLA latent + DSA index-key pages, no grad & 128K, 256K, 512K, 1M, then 2.097M prefix \\
Layer-0 replay & 1M unchunked MoE replay OOM & Chunk/offload prefix state; recompute a short suffix & Layer-0 1M and 2.097M backward + optimizer-call tests \\
All-layer replay & Missing IndexShare holder, CP guard, in-place views & Publish/reuse top-k holder; repair layer tail & Bounded 32K multi-layer tests \\
Topology/state & All-layer working set remains too large & TP1/CP32/EP32, CPU pages, layer checkpointing & 32K all-layer, then 64K all-layer \\
2.097M single member & Group semantics untested & All 78 layers, response replay, one backward & \(G=1\) backward + optimizer-call test \\
\rowcolor{yellow!12}
Predecessor grouped run & Old values and local group gradients must remain ordered & Freeze both old scores; run two serial backwards & \(G=2\), 32/32 ranks terminate; CP-local DSA and no gradient finalization \\
\rowcolor{green!8}
\textbf{Exact-2M online} & Rollout and training need different topologies; DSA and gradients must be global & Reuse 32 GPUs by phase; shared YaRN; global DSA; finalize then step & Real rollout, rewards \([-1,+1]\), two backwards, one finalized step on 32/32 ranks \\
\rowcolor{mindlabbluepale!20}
Full workflow closure & Rollout, global DSA, replay, and the distributed update must compose in one run & Execute the complete transaction across 32 H20 GPUs & Real rollout, rewards, two 78-layer backwards, gradient finalization, and one optimizer step \\
\bottomrule
\end{tabular}
\end{table}

\subsection{Stage I: Full-Graph Bottleneck Localization}

A conventional full-sequence LoRA path passed at 32K. Extending the same graph
to a 2,097,152-position prompt did not reveal one dominant allocation that could be optimized in
isolation. Reducing or bypassing one peak moved the OOM to the next part of the
layer. The observed sequence was DSA attention scratch, expert-LoRA scale/add
work, and finally MoE output concatenation near the H20 memory limit.

This movement is consistent with the GLM layer anatomy. DSA avoids dense core
attention over every key, but its indexer must still process the long context.
MoE evaluates a subset of experts per token, but full-sequence backward retains
routing, permutations, selected expert inputs, and adapter intermediates. At a
65,536-token shard, one fully expanded routed hidden buffer can already occupy
6~GiB, as derived in Section~\ref{tr:architecture}. Optimizing sparse attention
alone therefore cannot bring the complete beyond-2M autograd graph within the
per-rank H20 memory budget for this workload.

The conclusion from this stage was structural: the long prompt could not remain
inside the differentiable graph. The next tests separated prompt-state capacity
from response replay.

\begingroup
\addtolength{\parskip}{3.5pt}
\subsection{Stage II: Prefix-State Scaling}

The prefix-only path evaluated all 78 layers without autograd and retained MLA
latent pages plus DSA indexer-key pages. It was scaled through 128K, 256K,
512K, 1M, and 2,097,152. The final pass established that the full decoder could process
the prompt, CP-sharded pages could cover it, and prompt MoE work could be
released per layer without retaining a backward graph.

Prefix capture established the prompt-state capacity and exposed the interfaces
that the later training stages add: state restoration, response logits,
IndexShare lifetime, expert routing under autograd, and the optimizer event.
The next replay stages compose these pieces into the training transaction.

\subsection{Stage III: Single-Layer Differentiable Replay}

Layer-0 replay provided the first small differentiable slice. The runtime
restored one layer's prompt pages, appended a short response, produced a live
loss, and ran backward. Early 32K and 128K tests checked tensor shapes and
interfaces.
At 1M, GPU-resident state and MoE work became costly. CPU-resident pages and
chunked staging then enabled layer-0 backward plus optimizer-call tests at
1M and 2,097,152.

This stage separated two questions that are often conflated: whether the prompt
state fits, and whether a response query can consume it under autograd. The
layer-0 result answered both for one layer, but it could not exercise
cross-layer IndexShare or accumulate activation lifetime through the 78-layer
stack.

\subsection{Stage IV: All-Layer Architecture Integration}

Short 32K and 64K all-layer runs exposed bugs invisible to prefix-only and
layer-0 tests.

\paragraph{IndexShare lifetime.}
An index-computing layer must publish a top-k selection to a per-forward holder,
and dependent layers must consume the matching selection. The holder cannot be
global across response branches, and checkpoint recomputation must reproduce
the same producer/consumer order.

\paragraph{Resident DSA interface.}
The stock non-packed DSA path expected query and key layouts compatible with
its CP all-gather guard. A short resident query over long saved pages did not
match that interface. The predecessor runner used the runtime's unfused
absorbed MLA sparse fallback over materialized local pages. It restored
execution with CP-local selection; the final exact-2M runner in
Section~\ref{tr:glm-path} composes the global DSA path and finalizes gradients
before the optimizer step.

\paragraph{Views and in-place operations.}
Restored-page views tolerated in no-grad capture became invalid under autograd.
Explicit ownership and removal of in-place mutation fixed the layer tail.

\paragraph{Activation lifetime.}
Checkpointing only attention retained MoE routing and expert activations. A
complete-layer boundary made saved activations scale with response length.

The resulting 32K and 64K tests executed all 78 attention and FFN tails,
including 21 index producers, 57 IndexShare consumers, three dense FFNs, 75
MoE FFNs, and their backward paths.

\subsection{Stage V: Parallel Ownership and CPU Pages}

The final topology uses TP1/CP32/EP32/ETP1/PP1. CP32 assigns a portion of
the long prompt pages to every rank. EP32 assigns the 256 routed experts,
nominally eight experts per rank. CPU page storage and one-layer staging bound
attention-state residency; complete-layer checkpointing bounds response
activations. A shared RoPE cache avoids repeated layer-local position
allocation.

This topology is a budget choice, not a generic recipe for every MoE model. It
reuses the same 32 ranks for CP and EP rather than scaling a second device group;
attention state uses all ranks for CP, and the routed expert set is large enough
to use those ranks for EP. Both communication patterns still execute.

\endgroup
\subsection[Stage VI: Fixed-Budget Optimizer-Call Test]{Stage VI: A 2.097M Single-Member Test under the Fixed Budget}

Before introducing group accumulation, a \(G=1\) test ran the 2,097,152-position prompt,
all 78 response layers, one backward, and one optimizer call per worker under
the same 32-H20 allocation. This run combined state restoration, checkpoint
recomputation, local gradient materialization, and the optimizer path within the
target per-rank H20 storage envelope.

This \(G=1\) member is the staged control before grouped execution: it exercises
all-layer replay and the optimizer path, while the subsequent \(G=2\) run adds
frozen old scores and cross-response gradient accumulation in a fresh process.

\subsection{Stage VII: Predecessor Grouped Execution}

The predecessor run uses one 2,097,152-position prompt and two deterministic responses,
each with three input and two scored tokens. Rewards \([0,1]\) produce
advantages \([-1,1]\). The run captures the prompt, materializes both old-policy
score sets, then executes two serial 78-layer policy backwards into rank-8 LoRA
adapters. After the second backward, all 32 workers issue one optimizer call
and one gradient clear.

PyTorch allocation during prefix capture is not uniform across ranks. The
capture-window \texttt{max\_memory\_allocated} ranges from 112.571 to
145.148~GB per rank. A direct read of dsw-6601 reports NVIDIA H20-3e devices
with 143,771~MiB (140.401~GiB, or 150.755~GB in decimal) each. The 32-GPU
GLM log uses the recorded capture readings for the resource diagnostic; if those
workers expose the same per-device total, the readings correspond to
74.7--96.3\% of device total. This 32.577~GB spread reveals
substantial rank nonuniformity and motivates testing better placement and load
balance. A historical Qwen log also reports a PyTorch allocator capacity of
139.73~GiB (about 150.0~GB decimal); that process-level limit is not the H20
hardware specification. The counter is read before response replay, so it labels
the capture-window resource field. The completed 2M run is the measured grouped
operating point within the fixed budget.

This historical stage establishes all-rank grouped execution and two full
78-layer backwards under the fixed budget. Its historical DSA is CP-local, and
CP-replicated non-expert adapter gradients bypass Megatron finalization;
supplied responses and a detached prompt isolate capacity from online rollout.
Separate short-context integrations execute vLLM sampling, DAPO reward, and the
Tinker/Megatron response-only update on all 32 ranks, with finite values and
complete layer traces throughout.

\subsection{Stage VIII: Exact-2M Online Transaction}

The current run closes the stages that the predecessor separated. One real
DAPO-MATH example supplies an exactly 2,097,152-token prompt. A Tinker-managed
multi-node vLLM sampler at TP8/PP4 loads a vLLM-compatible projection of the
rank-8 policy LoRA and generates two completions,
\texttt{Answer: 2} and \texttt{Answer: 4}, and records their old
log-probabilities. Ground-truth scoring returns rewards \([-1,+1]\), so the
\(G=2\) update has a nonzero advantage in both directions.

The same 32 H20 GPUs are reused rather than expanded. After rollout, the sampler
is released and Megatron is restored at TP1/CP32/EP32/ETP1/PP1. Rollout and
training share an opt-in factor-two YaRN configuration over the checkpoint's
native 1,048,576-position range. The training path performs global cross-CP DSA,
two 78-layer response-only backwards, \texttt{finalize\_model\_grads}, and one
optimizer step on all 32 ranks. Its 64 trace files contain 25,536 events with no
non-finite values or execution errors.

The first exact-2M rollout candidate takes 1867.580 seconds, while a second
candidate using the cached prompt takes 30.798 seconds. These are candidate
latencies within one run, not complete-transaction time or matched throughput;
the run reports neither total wall time nor peak allocation.

\section{Execution Results and Runtime Traces}
\label{tr:evidence}

We report completed operations together with their measured scope. A row enters
the main table only when every participating worker reaches the requested
terminal boundary. The table distinguishes a local optimizer call, a finalized
distributed step, and the ownership contract used by each path.

\begin{table}[!htbp]
\centering
\caption{Fixed-budget long-context results. Qwen's 2M rows complete one
response-only step; its 4.25M row completes eight prefix-reuse steps. GLM
completes the full exact-2M online GRPO workflow with a finalized distributed
step. Hardware and suffix workloads differ, so wall times are not comparable.}
\label{tab:main-exact-results}
\scriptsize
\setlength{\tabcolsep}{2.7pt}
\begin{tabularx}{\textwidth}{@{}C{0.15\textwidth}C{0.16\textwidth}c C{0.15\textwidth}Y c c@{}}
\toprule
Path & Prompt / response & Observed \(G\) & Hardware/layout & Completed operations on every worker & Wall (s) & Peak GB \\
\midrule
\rowcolor{mindlabbluepale!20}
Qwen global forward & 2,088,960 / 8,192 input\(^*\) & 2 & 8 H20, CP8 & old + ref + backward + local AdamW call & 5198.780 & 97.503 \\
\rowcolor{mindlabbluepale!20}
Qwen global forward & 2,088,960 / 8,192 input\(^*\) & 8 & 8 H20, CP8 & eight serial members + local AdamW call & 6785.225 & 97.711 \\
\rowcolor{green!8}
Qwen prefix reuse & 4,448,256 / 8,192 input\(^*\) & 8 & 8 H20, CP8 & 8 steps; 64 member replays; AdamW/rank & \(8\!\times\!\approx4051\)\(^\ddagger\) & 83.894 \\
\midrule
\rowcolor{green!8}
GLM exact-2M online & 2,097,152 prompt / 5 generated (4 scored) per member & 2 & 32 H20; vLLM rollout TP8/PP4; train CP32/EP32 & policy rollout + rewards + \(2\!\times\!78\)-layer backward + global DSA + finalized step & n/r\(^\dagger\) & n/r \\
\bottomrule
\end{tabularx}
\vspace{2pt}

\raggedright\scriptsize
\(^*\)The Qwen scorer drops the first response label, so 8,192 response input tokens
yield at most 8,191 scored positions; the run log does not record the realized
count. Peaks are \texttt{max\_memory\_allocated}/\(10^9\) over the measured run.
\(^\dagger\)The GLM artifact reports 1867.580~s for the first rollout candidate
and 30.798~s for the cached candidate, not a complete transaction time. No valid
whole-run peak is reported. All rows condition on a detached prompt state;
Qwen CP8 replicated-adapter consistency remains a separate audit. The
observed \(G\) values are measured settings, not loop limits: both paths are
driven by configured member lists rather than hard-coded to those values, while
payload and score storage still grows with group size. Only the listed settings
have completed runs. \(^\ddagger\) is post-prefix time per cycle after one
resident prefix capture; the eight cycles intentionally reuse that cache after
parameter updates. Section~\ref{tr:analysis} documents the companion
prefix-comparison experiment.
\end{table}

\subsection{Runtime and Implementation Versions}

In the current rerun protocol, MinT Runtime supplies the model/session control
plane and managed Megatron trainer groups~\citep{mindlab2026mint}. The pinned
local stack implements its asynchronous request lifecycle with Ray-resident
Megatron workers.
LongStraw is an opt-in long-context execution extension on that substrate. It adds
GLM prefix capture, CPU-resident MLA/DSA state, CP/EP
ownership, response-only replay, serial GRPO accumulation, and version-pinned
validation. The 2M runner creates the model
through MinT and then invokes a LongStraw-installed method on the resident Megatron
actor rather than the stock MinT forward/backward path. MinT manages workers;
LongStraw performs long-context replay and backward.

The current implementation package at commit \texttt{2ec76d9}, tagged
\texttt{v0.2.2}, contains the exact-2M validation summary. It packages the
global-DSA replay, rollout-to-training topology handoff, distributed gradient
finalization, and validation artifacts used by the completed workflow.

Table~\ref{tab:reproduction-stack} lists the maintained source stack and this
provenance boundary.

\begin{table}[H]
\centering
\caption{Exact source versions for the maintained GLM exact-2M rerun stack.}
\label{tab:reproduction-stack}
\scriptsize
\setlength{\tabcolsep}{3pt}
\renewcommand{\arraystretch}{1.16}
\begin{tabularx}{\textwidth}{@{}C{0.13\textwidth}C{0.27\textwidth}C{0.25\textwidth}Y@{}}
\toprule
Component & Exact revision & Role & Source version \\
\midrule
MinT Runtime &
\texttt{12c83d90\allowbreak{}4df5faf3\allowbreak{}e2cd6063\allowbreak{}3b448a83\allowbreak{}17d84ee0} &
Model/session control plane and Ray-resident Megatron worker substrate &
Clean exact checkout; the exact-2M run activates the validated LongStraw extension \\
Megatron-LM &
\texttt{03db8324\allowbreak{}007ed7b3\allowbreak{}3edffc14\allowbreak{}7160bebf\allowbreak{}9552846c} &
Distributed model, parallelism, gradient, and optimizer runtime &
Clean exact checkout \\
Megatron-Bridge &
\texttt{22edeb2a\allowbreak{}487d6a9c\allowbreak{}c0dcea56\allowbreak{}7827826c\allowbreak{}c76427c2} &
GLM-5.2 model and LoRA configuration bridge &
Exact base plus the bundled GLM-5.2 integration patch \\
verl &
\texttt{d2916f5a\allowbreak{}0ed34646\allowbreak{}4d8999e0\allowbreak{}40e0ebb0\allowbreak{}5bb8fadf} &
Training-datum conversion and MCore integration used by MinT &
Exact base plus the bundled MCore compatibility patch \\
LongStraw-alpha &
\texttt{2ec76d91\allowbreak{}19936822\allowbreak{}8e3d455f\allowbreak{}a5cec3f8\allowbreak{}5ebf735d} &
Global DSA, response replay, topology handoff, validation evidence &
Clean local \texttt{v0.2.2} validation package \\
GLM-5.2 &
\texttt{b4734de4\allowbreak{}facf877f\allowbreak{}85769a91\allowbreak{}1abafc52\allowbreak{}83eab3d9} &
Base weights, tokenizer, and model configuration &
Exact model snapshot revision \\
\bottomrule
\end{tabularx}
\end{table}

\subsection{Qwen Results}

The Qwen \(G=2\) and \(G=8\) probes share one 2,088,960-position prompt and use
8,192 response-input tokens per member, producing a context length of exactly
2,097,152.
Both finish on eight H20 workers. The
reported whole-run allocated-memory peaks are 97.503 and 97.711 decimal GB per
rank, and wall times are 5198.780 and 6785.225 seconds. The near-flat peak from
\(G=2\) to \(G=8\), together with the recorded member ordering, validates the
serial response-graph lifetime at these endpoints. Post-prefix work is 271.278
versus 266.476 seconds per member, while mean total wall time per response falls
from 2,599.390 to 848.153 seconds because the same prefix is amortized over four
times as many members. These endpoints expose the serial group-scheduling
behavior directly.

Every worker records old/reference scoring, full policy-response backward, and
a local AdamW call, completing the stated single-step response-only program.
This completes that objective: response gradients are materialized
and an optimizer call executes on every rank. Forward response attention
composes all CP8 KV partitions. The backward implementation separately narrows the
distributed-update claim: \(dQ\) is
all-reduced, while page-owner \(dK/dV\) contributions to replicated adapters
are not. The companion CP8 adapter audit keeps this ownership path separate from
the completed response-only execution and prefix-reuse receipts.

At 4,456,448 positions, the resident prefix-reuse route adds repeated training
results. It completes eight \(G=8\) accumulation-and-step cycles, 64 member
replays in total, with all eight ranks reporting each applied optimizer step.
The peak is 83.894~GB per rank. The one-time prefix capture takes 17,729.8
seconds, while each post-prefix cycle takes about 4,051 seconds. Keeping that
cache across optimizer steps is an intentional resident-reuse mode that avoids
repeated multi-hour prefills.

A separate 1M probe compares this mode with a freshly recomputed prefix on real
DAPO prompt tokens. Recaptured-versus-resident GRPO loss differs by only
0.1236\% and 0.0378\% after steps one and two, with mean absolute policy-
log-probability differences of 0.0137 and 0.0250. At steps four and eight, loss
differences grow to 22.81\% and 9.33\% in magnitude and the log-probability
differences to 3.136 and 3.774. The prompt and response segments are drawn from
real DAPO text, but they are not model-sampled completions and the
rewards/advantages remain synthetic. For this measured workload, the values
provide a direct comparison between recaptured and resident prefix-state
execution.

The workload is synthetic. Old and reference scores come from the same current
runner, \(\beta=0\), and the implemented live policy term is unclipped. At the
first step, old and current scores coincide and the importance ratio is one.
The probes do not exercise an independent reference model, nonzero KL, or the
clipped-min branch of the GRPO objective in Equation~\ref{eq:grpo}.

\subsection{GLM Exact-2M Online Transaction}

The exact-2M run starts from one real DAPO-MATH example and the active rank-8
policy LoRA. A Tinker-managed multi-node vLLM sampler at TP8/PP4 loads its
vLLM-compatible projection, processes an exactly 2,097,152-token prompt, and
produces two five-token completions, each with four scored log-probabilities.
Ground-truth evaluation assigns rewards \([-1,+1]\), after which Megatron
resumes on the same 32 H20 GPUs.

The sampler log records \texttt{MultiNodeVLLMEngine} initialization and
successful LoRA loading before exact-2M generation. The policy checkpoint
content matches across the sampler handoff.

Every rank records two live response-only backwards, global cross-CP DSA,
\texttt{finalize\_model\_grads}, one optimizer step, and one gradient clear.
The recorded importance ratios activate one lower and one upper clipping event
at \(\epsilon=0.2\); the reference coefficient is \(\beta=0\). The policy
checkpoint content matches across the sampler handoff. No rank records a
non-finite value or execution error.

Rollout and training use the same factor-two YaRN position configuration over a
checkpoint whose native context is 1,048,576. The first rollout candidate takes
1867.580 seconds and the cached candidate 30.798 seconds. These are
candidate-latency fields in the exact-2M receipt; the main table keeps
whole-step time and peak as \mbox{n/r} because those fields are not part of this
artifact.

\Needspace{5\baselineskip}
\subsection{Runtime Trace Coverage}

The exact-2M GLM run emits separate rank/group JSONL traces. Table~\ref{tab:trace-audit}
summarizes the inventory.

\begin{table}[!htbp]
\centering
\caption{GLM exact-2M online trace inventory. The audit covers all rank groups,
global DSA execution, both response backwards, gradient finalization, and the
single optimizer step.}
\label{tab:trace-audit}
\scriptsize
\setlength{\tabcolsep}{4pt}
\renewcommand{\arraystretch}{1.16}
\begin{tabularx}{\textwidth}{@{}C{0.22\textwidth}
    C{0.16\textwidth}
    C{0.18\textwidth}
    Y@{}}
\toprule
Audit class & Observed count & Rank coverage & Interpretation \\
\midrule
\rowcolor{green!7}
Trace inventory & 64 files; 25,536 events & All rank groups & No non-finite values or execution errors \\
\rowcolor{green!7}
Global-position DSA backend & 3,648 events & All rank groups & Candidate positions retain global coordinates \\
\rowcolor{green!7}
Fused global indexer & 1,344 events & All rank groups & Global top-2,048 selection is exercised \\
\rowcolor{mindlabbluepale!20}
Update sequence & 2 backwards + 1 finalize + 1 step per rank & 32/32 ranks & \texttt{finalize\_model\_grads} precedes optimizer consumption \\
\bottomrule
\end{tabularx}
\end{table}

\begingroup
\looseness=-1
The 64 files contain 25,536 events. Across all rank groups, the audit records
3,648 global-position DSA backend events and 1,344 fused global-indexer events,
as well as both response backwards and model-gradient finalization. No event is
non-finite or erroneous; together, the traces cover the exact-2M operator
sequence and control flow across every recorded rank group in the job.
\par
\endgroup

Together, the traces establish complete distributed execution of the exact-2M
online workflow from global sparse selection through the finalized optimizer
step on all 32 ranks.

\subsection{Memory Accounting}

After model, adapter, optimizer, and input construction, Qwen resets the CUDA
peak counter for the measured probe. It reports allocated bytes in decimal GB,
excluding reserved blocks and host memory.

The predecessor GLM run records a capture-window
\texttt{max\_memory\_allocated} range of 112.571--145.148~GB per rank. The
counter is reset immediately before no-grad prefix capture and read before
full-GRPO response replay. The prompt-state representation stores 5.8125~GiB on CPU
per rank and stages only 72~MiB for an IndexShare layer or 88~MiB for an
index-computing layer, so the length-dependent state is not persistently
GPU-resident. The predecessor rank spread shows nonuniformity and motivates
placement and load-balance measurements. The capture window is a resource field
for the exact-2M operating point; rank traces and candidate latency describe
the complete transaction, while Table~\ref{tab:main-exact-results} marks
missing whole-step fields as \mbox{n/r}.

\Needspace{0.10\textheight}
\subsection{Receipt Matrix}

Table~\ref{tab:claim-matrix} distinguishes program execution, global forward
fidelity, and distributed updates.

\begin{table}[H]
\centering
\caption{End-to-end execution matrix. GLM closes the execution, global
response-forward, distributed-update, and optimizer-step stages in one
exact-2M online workflow.}
\label{tab:claim-matrix}
\scriptsize
\setlength{\tabcolsep}{3pt}
\renewcommand{\arraystretch}{1.18}
\begin{tabularx}{\textwidth}{@{}C{0.17\textwidth}
    C{0.19\textwidth}
    Y
    Y@{}}
\toprule
Validation level & Criterion & Qwen3.6-27B & GLM-5.2 \\
\midrule
\rowcolor{green!7}
Program execution & Requested stages, backwards, and optimizer steps complete & \textbf{Yes.} Exact 2.097M $G=2/G=8$ single steps and eight 4.25M prefix-reuse $G=8$ steps complete on eight H20 GPUs & \textbf{Yes.} One real exact-2M $G=2$ transaction completes policy rollout, mixed rewards, two 78-layer backwards, and one step on 32 H20 GPUs \\
\rowcolor{mindlabbluepale!20}
Global response forward & Response tokens use the intended prompt-wide operator & \textbf{Yes.} CP8 performs a global LSE/output merge with BF16 numerator accumulation & \textbf{Yes at exact 2M.} CP32 exchanges candidate positions and selected values, then composes global DSA output \\
\rowcolor{green!7}
Distributed update & Adapter gradients are finalized across the distributed ownership layout before the step & \textbf{Response-only closure.} $dQ$ is all-reduced; page-owner K/V contributions and replicated-adapter synchronization are tracked in the companion CP8 audit & \textbf{Yes at exact 2M.} Two backwards precede \texttt{finalize\_model\_grads} and one optimizer step on every rank \\
\rowcolor{green!7}
Full online transaction & Rollout, global response computation, two backwards, distributed gradient finalization, and the optimizer step execute in one workflow & \textbf{Response replay path.} Qwen supplies the declared response-only objective and grouped optimizer cycles; its short-context receipt covers sampled online control flow & \textbf{Yes at exact 2M.} Real rollout and rewards feed global DSA, two 78-layer backwards, gradient finalization, and one optimizer step on all 32 ranks \\
\bottomrule
\end{tabularx}
\end{table}

The matrix organizes the receipts by execution, forward composition, and update
ownership. Qwen's page-owner K/V path is recorded in the companion CP8
adapter-synchronization audit. The current GLM path performs global DSA
candidate/output composition and Megatron gradient finalization at exact 2M,
followed by one optimizer step on all 32 ranks. The same working transaction
provides a direct execution base for repeated updates, task metrics, and
longer-horizon training studies.

\section{Fixed-Budget Systems Lessons}
\label{tr:analysis}

The two implementations share a prompt-state graph boundary but expose
different bottleneck chains, determined by what each architecture must retain
and communicate afterward.

\subsection{Fixed-Budget Capacity Comes from Lifetime, Not Sparsity Alone}

Neither path avoids the long prompt forward, and neither path adds accelerators
as context grows within its reported envelope. Every prompt token still passes
through every decoder layer. The capacity gain comes from allowing prompt
attention scratch, dense FFN intermediates, MoE routes, expert-token
permutations, and adapter activations to die before response backward. Once the
prompt has been captured, the live autograd working set follows response length
rather than prompt length.

This explains why several plausible optimizations are insufficient on their
own. DSA reduces attention arithmetic but retains long-context indexing
~\citep{deepseekv32,bai2026indexcache}. MoE evaluates only selected experts but
expands tokens into routed rows and distributes a very large parameter set
~\citep{shazeer2017moe,lepikhin2020gshard,fedus2021switch}. QLoRA reduces
persistent trainable state, not activations~\citep{dettmers2023qlora}.
Activation checkpointing reduces saved tensors by trading storage for
recomputation~\citep{chen2016training,korthikanti2022recomputation}. In the
traced runs reported here, recomputation still re-created peak workspace at
the guarded boundary. The
working design composes all of these tools around the prompt-state graph
boundary shared by both execution paths under review.

\begin{tcolorbox}[
  float=t,
  colback=mindlabbluewash,
  colframe=mindlabblue,
  boxrule=0.55pt,
  arc=0pt,
  left=7pt,right=7pt,top=5pt,bottom=5pt]
\textbf{Observed numeric envelope.}
\begin{itemize}[leftmargin=1.5em,itemsep=1pt,topsep=3pt]
  \item \textbf{Fixed accelerator budgets:} Qwen uses eight H20 GPUs in its
  reported 2M and 4.25M paths; the GLM grouped 2M path uses 32 H20 GPUs.
  Device count is an input constraint, not the reported scaling axis.
  \item \textbf{Context lengths:} Qwen executes
  \(2{,}088{,}960+8{,}192=2{,}097{,}152=2^{21}\) positions per response,
  exactly \(8\times\) its stored native setting. GLM executes an exact
  2,097,152-token online prompt with an explicit factor-two YaRN extension
  over its 1,048,576-position native setting.
  \item \textbf{Configured group:} Qwen \(G=2\rightarrow8\) changes peak
  allocation by \(+0.208\)~GB (\(+0.213\%\)) while adding 1,586.445~s;
  one prompt capture is amortized across all serial members.
  \item \textbf{Measured-phase accelerator-hours:} elapsed time times allocated
  devices gives 11.553 and 15.078 H20-hours for Qwen 2M (G=2) and (G=8),
  48.334 for the 4.25M resident replay. The predecessor GLM grouped
  diagnostic accounts for 26.446 H20-hours; the exact-2M receipt is represented
  by its candidate timings and complete rank/group event trace.
  These device-time estimates define the fixed-device comparison axis.
  \item \textbf{Qwen capacity bracket:} Qwen's physical
  KV arithmetic adds 8.192 decimal GB per rank per one million additional
  global prompt positions. A train-block proxy passes at 4,538,368 and OOMs
  one 4,096-position chunk later, but a fuller path already OOMs in policy
  backward at 4,456,448.
  \item \textbf{GLM 2M operating point:} GLM retains 5.8125~GiB of
  CPU prompt state per rank and stages 72--88~MiB per layer on GPU. The recorded
  capture-window peak allocation ranges from 112.571 to 145.148~GB per rank,
  revealing a 32.577~GB rank spread and motivating better-balancing tests.
  The exact-2M transaction adds global DSA, response replay, gradient
  finalization, and the optimizer event to this state-accounting picture.
\end{itemize}
\end{tcolorbox}

\subsection{Physical Ownership Is Part of the Algorithm}

A context shard is useful only when its physical allocation is also sharded.
The first Qwen page implementation retained small slices whose parent chunks
remained allocated. The logical page table looked distributed while allocator
memory did not change. Copying selected pages into right-sized tensors made the
ownership statement true and moved the 2,088,960-position prefix peak into the feasible range.

The same principle applies to GLM CPU state. A page tensor, its position order,
and the consuming layer form one object. Restoring correct bytes in the wrong CP
order changes causal positions and sparse selection. Offload transfers operator
state and must preserve identity and order.

\subsection{Dense and MoE Move the Peak to Different Places}

For Qwen, every token evaluates the same dense FFN, and no expert routing state
crosses devices. Context-growing storage is concentrated in 16 full-attention
KV sets; 48 GDN boundaries remain fixed-size with prompt length. Page compaction
and context partitioning therefore target the dominant stored state.

For GLM, expert sparsity separates total parameters from activated parameters,
but full-sequence training still handles eight expert assignments per token.
At the final CP shard size, one expanded BF16 hidden buffer can be 6~GiB before
output, permutation, or LoRA work. Removing DSA scratch simply revealed the
next MoE allocation. Response-only replay succeeds because only a few suffix
rows are routed under autograd; the no-grad prompt routes remain transient.

The broader lesson is that sparsity transfers cost. DSA turns dense attention
into an index-selection and selected-value movement problem. MoE turns dense
FFN compute into expert residency and token communication. A training system
must implement the transferred problem, not only count fewer FLOPs.

\subsection{Context and Expert Parallelism Are Orthogonal}

Context parallelism partitions token history. Expert parallelism partitions
FFN parameters~\citep{liu2025moefolding}. Folding CP32 and EP32 onto the same 32 ranks is a useful
placement, but their collectives have different meanings. EP all-to-all sends
response rows to expert owners and combines expert outputs. It cannot turn 32
local sparse candidate sets into one global top-2048 set. A CP attention merge
cannot balance routed expert load. This distinction also appears in backward.
A globally composed attention forward and its backward ownership contract are
separate stages. In the inspected Qwen path, \(dQ\) is all-reduced, while
\(dK/dV\) and the corresponding adapter gradients remain local to page owners;
eight independent AdamW instances then step. The companion CP8 audit records
this adapter-synchronization boundary. GLM composes global DSA and calls
\texttt{finalize\_model\_grads} before its exact-2M optimizer step. Distributed
gradient ownership is therefore explicit for both paths.

\subsection{Forward and Update Ownership Contracts}

The Qwen and GLM paths occupy different positions across three validation
levels.

\begin{itemize}[leftmargin=1.6em]
  \item Both complete fixed-budget 2.097M runs. Qwen completes its supplied-
  response program; GLM completes a real policy rollout, reward computation,
  two response backwards, distributed gradient finalization, and one optimizer
  step on every worker.
  \item Qwen composes a global full-attention response forward over CP8. Its
  production numerator reduction uses BF16, so the claim is partition-correct
  forward semantics, not bitwise FP32 equality. GLM exchanges candidate
  positions and selected values across CP32 and composes global DSA at exact 2M.
  \item Qwen completes its stated response-only update objective, with
  page-owner K/V-adapter synchronization recorded in the companion CP8 audit.
  After both backwards in the exact-2M transaction, GLM calls
  \texttt{finalize\_model\_grads} and then applies one optimizer step on every
  rank.
\end{itemize}

The Qwen response-side backward has native/blockwise operator checks for both
full attention and GDN, together with chunked replay checks across the response
path. These checks support response-side operator and chunking fidelity. For GLM,
exact-2M traces record two full backwards, global DSA, gradient finalization,
and an optimizer step on every rank in the exact-2M receipt.

\subsection{Group Scaling Is a Scheduling Result}

Qwen \(G=8\) consumes nearly the same peak allocated memory as \(G=2\),
consistent with response branches being serialized after one prompt capture.
A fourfold group increase raises measured peak allocation by 0.208~GB, or
0.213\%, while post-prefix work grows by a factor of 3.93. The marginal time
across the six additional members is 264.408 seconds per member. Meanwhile,
sharing the 4,655-second prefix reduces mean total wall time per supplied
response by 67.4\%, from 2,599.390 to 848.153 seconds.

The implementation is parameterized by group cardinality rather than capped at
the observed \(G=2\) or \(G=8\). The serial loop consumes a configured member
list; live response autograd is bounded by the largest member while inputs,
scores, rewards, and reports accumulate with \(G\). Changing the group also
changes reward normalization and the GRPO estimator. The Qwen endpoints show
that group size is not the dominant live-autograd capacity axis in this design;
the exact-2M GLM receipt uses a nondegenerate \(G=2\) reward pair.

These single-run timings span different models, GPU counts, suffixes, and
numerical paths. They report terminal execution inside the specified resource
envelopes.

\subsection{Eight H20s Carry Qwen to a 4.25M Context Envelope}

Under the same eight-H20 budget used for the 2,097,152-position runs, Qwen
reaches 4.25M context, where 4.25M means exactly
\(4.25\times2^{20}=4,456,448\) positions: 4,448,256 prompt positions and an
8,192-position response. A resident \(G=8\) run completes one 1,086-chunk
prefix capture, all eight serial old/reference/policy branches, and all four
2,048-position response-backward blocks per member. The measured path takes
21,750.133 seconds from prefix start through the group and peaks at
82.960~GB per rank. This is a complete 4.25M response-replay run, not a
prefix-only or stage-1-forward result.

The resident reuse run extends that result to an 8-step curve. Each step
accumulates all eight group members before applying an optimizer update. Across
64 member replays, every rank records eight applied optimizer steps and a peak
of 83.894~GB. A prefix capture at this scale takes 17,729.8 seconds, compared
with about 4,051 seconds per post-prefix \(G=8\) cycle. Reusing the prefix for
eight cycles therefore avoids seven additional captures; using the measured
capture cost, this is an estimated 34.5 hours of avoided prefill.

A separate 1M real-DAPO-token prefix-comparison run evaluates the fresh and
resident execution modes. The recaptured-versus-resident GRPO loss difference is
\(+0.1236\%\) after step one and \(+0.0378\%\) after step two; mean absolute
policy-log-probability differences are 0.0137 and 0.0250. At steps four and
eight, the loss differences are \(-22.81\%\) and \(-9.33\%\), with mean
absolute log-probability differences of 3.136 and 3.774. The prompt and
response segments are drawn from real DAPO text, but the response segments are
not model-sampled completions and the rewards/advantages are synthetic. These
measurements provide a direct record of the two prefix-state execution modes
for the measured workload.

The remaining frontier numbers provide companion capacity measurements. A clean
prompt-adapted run before detached-prefix gradient-page
pruning OOMed in policy backward. A train-block proxy reaches 4,538,368 before
the next 4,096-position chunk OOMs at 4,542,464. None of these labels changes
the separate CP8 replica-finalization test: local optimizer application is
recorded; the companion CP8 audit isolates the cross-rank replicated-adapter
ownership path. See Appendix~\ref{app:qwen-frontier} for full details.

\subsection{The Fixed-Budget Operating Envelope}

The comparison axis is the fixed accelerator envelope:
4,456,448 Qwen positions on eight H20 GPUs and 2,097,152 GLM prompt tokens on
32 H20 GPUs. Adding devices is a valid scale-out strategy, but it is outside
these experiments. LongStraw instead trades GPU residency for compact physical
pages, CPU state, recomputation, serial replay, and phase-specific topology.
The GLM run reuses the same 32 devices first for TP8/PP4 rollout and then for
TP1/CP32/EP32 training.

The stored Qwen configuration has a native maximum position setting of 262,144;
its run at a context length of 2,097,152 is exactly \(8\times\) that
setting. The published \mbox{GLM-5.2} configuration uses 1,048,576; the exact-2M
online run uses a shared opt-in YaRN extension for rollout and training. The 4.25M
prefix-reuse run provides a repeated eight-step training curve, and a separate
1M real-DAPO-token prefix-comparison probe records fresh-versus-resident
behavior. Its response segments are real text but not model rollouts, and its
rewards/advantages are synthetic; task-quality evaluation is a separate axis
from these systems receipts.
The result is an accelerator-bounded systems operating envelope with
objective-scoped multi-step measurements. The exact-2M GLM receipt additionally
closes the online rollout-to-update path, while task-quality studies are a
separate evaluation axis reported alongside the systems receipts.

\section{Related Work}
\label{sec:related_work}

\subsection{Scale-Out Long Context and the Fixed-Budget Axis}

Prior work already establishes that million-token sequence processing is
possible. Ring Attention reports exact-attention training at 4.096M positions
for a 7B model on 32 A100 GPUs~\citep{liu2023ring}. DeepSpeed-Ulysses studies a
scale-out regime in which sequence length and device count grow together; its
experiments scale to 256 A100 GPUs and include a one-million-token sequence for
a 1.2B GPT model~\citep{jacobs2023ulysses}. ByteScale reports a 2M LLaMA-7B
case on 1,024 GPUs within a production cluster exceeding 12,000 GPUs
~\citep{ge2025bytescale}. USP then combines ring-style and all-to-all sequence
parallelism and analyzes its interaction with tensor parallelism, ZeRO,
recomputation, and offload~\citep{fang2024usp}.
DistFlashAttn adds load-balanced exact-attention scheduling and overlaps
peer-to-peer KV transfer with attention compute, while LoongTrain combines
head and context parallelism through 2D-Attention and a double-ring schedule
~\citep{li2023distflashattn,gu2024loongtrain}. Both remain scale-out methods
that widen device-level parallelism.

Ring Attention and ByteScale provide full-sequence scale-out reference points.
OOMB is a neighboring fixed-budget long-context system in the broader lineage
of state partitioning, offload, paging, and activation recomputation. It
combines chunk-recurrent training and on-the-fly recomputation with paged KV
cache and gradient management, asynchronous CPU offload, and dense or
page-level sparse attention for million-token full-parameter fine-tuning
~\citep{li2026out}. LongStraw instead studies shared-prefix, multi-response GRPO:
one prompt conditions several old/reference/policy branches, whose lifetime,
accumulation, and refresh must be coordinated across Qwen hybrid attention and
GLM MLA/DSA/MoE. The Qwen prototype transparently uses selected OOMB
\texttt{chunkoptim} cache and paged-attention kernels as low-level components,
but the systems method is not an algorithmic extension of OOMB. We cite OOMB
as conceptual and implementation-component lineage, not as an equivalent
training objective or an unmatched efficiency baseline.

Large-scale technical reports operate at a different industrial scale.
DeepSeek-V3 reports a 2,048-H800 training cluster and 2.788 million H800
GPU-hours for its full 671B-MoE training program~\citep{deepseekv3}.
LongCat-Flash reports a 560B MoE trained with infrastructure spanning tens of
thousands of accelerators, while GLM-5 reports a 744B MoE trained over 28.5
trillion tokens~\citep{longcatflash,glm5team2026}. These results are not
apples-to-apples baselines for LongStraw: model size, objective, hardware, sequence
semantics, and experimental scope all differ. They instead establish why our claim
is not ``first long context.'' LongStraw fixes the accelerator envelope at eight
H20 GPUs for Qwen and 32 H20 GPUs for GLM, then asks which state-lifetime and
ownership decisions make a GRPO-shaped execution path fit. We report a
fixed-budget operating envelope and compare accelerator-bounded GRPO execution
with the surrounding scale-out literature for context.

\subsection{Relationship to MinT}

MinT manages LoRA adapter revisions across rollout, update, export, evaluation,
and serving over resident base-model deployments; its training plane includes
distributed Megatron execution for dense and MoE models, parallelism-aware
adapter state, and MLA/DSA support~\citep{mindlab2026mint}. The current LongStraw
rerun interface directly reuses that managed control plane and resident
Megatron LoRA substrate, but changes the state boundary inside one update: it
captures architecture-specific prompt state without autograd, stages that state
under CP/EP ownership, and serially rebuilds response graphs. This architectural
lineage connects MinT's managed training plane to the complete exact-2M
LongStraw workflow. MinT's million-adapter result measures addressable policy-
catalog scale with bounded serving working sets, not long-context execution.
Our 2M and 4.25M results measure context positions, not adapter count or serving
concurrency.

\subsection{Memory-Efficient and Distributed Attention}

Exact attention does not require materializing the full quadratic score matrix.
Memory-efficient algorithms stream score blocks or improve IO locality while
preserving the softmax result~\citep{rabe2021self,dao2022flashattention}. Ring
Attention distributes sequence blocks over devices and composes attention as
the blocks circulate~\citep{liu2023ring}. DeepSpeed-Ulysses exchanges
sequence and attention-head partitions with all-to-all collectives, while USP
combines Ulysses-style and ring-style sequence parallelism
~\citep{jacobs2023ulysses,fang2024usp}. These methods address the core
attention computation. Our setting adds a training-specific boundary: prompt
state is retained across old, reference, and policy branches, while the suffix
is replayed under autograd. The central questions become physical page
ownership, state validity, and which distributed reductions are required in
both forward and backward.

PagedAttention makes KV allocation and page tables first-class serving-system
objects~\citep{kwon2023efficient}. In this report, pages cross a training
boundary. A logical page shard must own a compact physical allocation; a view
into a larger parent chunk does not release memory. Prompt pages are read-only
during grouped response replay, and their validity ends when an optimizer step
changes the adapted parameters.

\subsection{MLA, Sparse Attention, and Index Reuse}

MLA compresses KV state into a latent~\citep{deepseekv2}.
DeepSeek-V3.2 forms DeepSeek Sparse Attention (DSA) by adding a lightweight
top-k indexer to MLA~\citep{deepseekv32}. The GLM-5 report provides MLA/DSA
background~\citep{glm5team2026}; the released \mbox{GLM-5.2} configuration
file instantiates index producers and \mbox{IndexShare}
consumers~\citep{glm52config}.
IndexCache formalizes cross-layer sparse-index reuse~\citep{bai2026indexcache};
the released GLM configuration specifies this model's pattern.

Our focus is the resulting training-state requirements. A saved MLA latent page is
not sufficient when the sparse indexer also needs long-context keys. Index
reuse saves work but creates producer/consumer lifetime inside each response
forward. Local sparse selection is not global selection over a context-parallel
prompt. Distributed training must define candidate merge, selected-value
movement, and output composition.

\subsection{MoE Training and Multidimensional Parallelism}

Sparse MoE models distribute the full parameter set but activate only selected
experts per token~\citep{lepikhin2020gshard,fedus2021switch}.
Expert parallelism reduces per-rank parameter residency but introduces token
permutation, all-to-all dispatch, expert computation, and combine. DeepSeek-V3
links fine-grained MoE to cross-node overlap and memory-efficient
training~\citep{deepseekv3}. LongCat-Flash
similarly co-designs MoE layer structure, communication overlap, deterministic
kernels, and its EP/CP/PP layout~\citep{longcatflash}. MoE Parallel Folding
analyzes heterogeneous tensor, context, expert, data, and pipeline
mappings~\citep{liu2025moefolding}. Tutel selects MoE all-to-all implementations
and pipeline degree by workload and cluster scale; MegaBlocks maps
irregular expert-token work to block-sparse operations
~\citep{hwang2022tutel,gale2022megablocks}.
These systems optimize sparse kernels, communication, and parallel mappings;
LongStraw instead isolates long-lived prompt state from each short-lived native MoE
replay and makes the remaining distributed-gradient obligations explicit
before optimizer consumption.

The GLM path in this report uses CP32 and EP32 over the same ranks. This
placement is economical, but the parallel dimensions remain semantically
different. CP owns context and must preserve attention; EP owns experts and
routed-token computation. Neither collective can replace the other.

\subsection{Distributed Training State and Optimizer Sharding}

Megatron-LM established tensor model parallelism for large
Transformers~\citep{shoeybi2019megatron}; subsequent Megatron training systems
compose tensor, pipeline, and data parallelism at cluster scale
~\citep{narayanan2021megatron}. ZeRO and PyTorch FSDP instead shard
parameters, gradients, and optimizer state across data-parallel workers
~\citep{rajbhandari2020zero,zhao2023fsdp}. These systems reduce persistent
state and define ownership for standard layer graphs. LongStraw composes CP, EP,
and sharded optimizer machinery with custom prompt-state replay. That boundary
must still call the appropriate gradient finalization or selective reduction;
optimizer sharding cannot recover contributions that never reach a parameter
owner.

Heterogeneous-memory variants widen the placement space. ZeRO-Offload moves
selected model states from GPU to CPU memory, while ZeRO-Infinity can place
partitioned model states in CPU or NVMe memory
~\citep{ren2021zerooffload,rajbhandari2021zeroinfinity}. LongStraw's CPU prompt
pages are different state with a different lifetime, but they inherit the same
requirement that ownership and transfer timing be explicit.

\subsection[Activation Checkpointing and PEFT]{Activation Checkpointing and Parameter-Efficient Adaptation}

Activation checkpointing trades recomputation for lower retained activation
memory~\citep{chen2016training}. For the GLM response path, the useful boundary
is the complete decoder layer. Checkpointing attention alone leaves MoE router,
dispatch, selected-expert, and adapter intermediates alive. The long prompt is
not checkpointed for backward; it is evaluated without autograd and represented
by stored conditional state.
Selective activation recomputation instead retains the layer boundary while
recomputing only memory-heavy, relatively inexpensive attention operations
~\citep{korthikanti2022recomputation}. That complementary design reduces
redundant recompute; it does not replace the full-layer boundary required by our
native MoE replay.

LoRA and QLoRA reduce trainable parameter, gradient, and optimizer-state
storage~\citep{hu2021lora,dettmers2023qlora}. They do not remove response
activations or the need to synchronize replicated adapter gradients. In Qwen,
full-attention forward composition is global, but coherent CP8 adapter updates
still require an additional cross-rank gradient-finalization test in
Section~\ref{tr:qwen-path}. AdaLoRA allocates adapter rank under a parameter
budget, while DoRA separates weight magnitude from low-rank directional
updates~\citep{zhang2023adalora,liu2024dora}. Related work explores
hierarchical rank allocation, intra/inter-layer adapter sharing,
mixed-precision fidelity, and joint quantization with low-rank adapters
~\citep{zhou2025rankadaptor,zhou2025bslora,zhou2026balancing,zhou2026autoqra}.
GPTQ and SparseGPT are established post-training quantization and pruning
methods~\citep{frantar2022gptq,frantar2023sparsegpt}; global rank/sparsity
optimization and probabilistic layer quantization further change the base
model's storage and sensitivity profile~\citep{zhou2025largecompression,zhou-etal-2025-qpruner}.
Qwen uses NF4 QLoRA and GLM rank-8 LoRA; adapter design is not a LongStraw
contribution. LongLoRA combines LoRA with shifted sparse attention for efficient
long-context adaptation~\citep{chen2023longlora}; LongStraw instead preserves
the model's native prompt semantics and changes graph lifetime and placement.

\subsection{Adapter Serving and Inference Infrastructure}

Punica and S-LoRA establish multi-tenant batching, memory management, and
kernel paths for serving many adapters concurrently
~\citep{chen2023punica,sheng2023slora}. Dynamic operator selection likewise
treats adapter placement and operator reuse as serving-time systems problems
~\citep{zhou2025dynamic}. Budget-driven depth routing and dynamic low-rank
substitution address adaptive inference, where the objective is to reduce
latency or compute for a fixed request~\citep{zhou2026buddy,zhou2026deputy}.
LongStraw is adjacent but distinct: it targets the training-time prompt/response
boundary, preserves architecture-specific state across a grouped GRPO update,
and exposes the gradient-ownership conditions that serving systems do not need
to satisfy.

\subsection{GRPO Systems}

PPO alternates policy sampling with multiple optimization epochs over a clipped
surrogate objective~\citep{schulman2017ppo}. GRPO normalizes outcome rewards
within a response group and removes the learned critic used by PPO-style
training~\citep{shao2024deepseekmath}. DeepSeek-R1-Zero applies GRPO in a
reasoning-training pipeline that begins without supervised fine-tuning
~\citep{deepseek2025r1}. DAPO then extends the GRPO family with decoupled
clipping, dynamic sampling, token-level policy-gradient loss, and overlong
reward shaping~\citep{yu2025dapo}. These methods change optimization behavior,
not prompt-state lifetime.

DeepSeek-V3.2
describes additional stabilization mechanisms for scaled GRPO, including
off-policy sequence masking and preservation of MoE routing between inference
and training~\citep{deepseekv32}. LongCat-Flash-Thinking-2601 develops an
asynchronous system for long-tailed environment interaction and large-scale
agentic RL~\citep{longcatthinking2601}. DeepSpeed-Chat, HybridFlow, and OpenRLHF
address the broader RLHF execution problem, including model-role placement and
transitions among generation, scoring, and training; OpenRLHF assigns rollout
and actor/training engines distinct roles under Ray
~\citep{yao2023deepspeedchat,sheng2024hybridflow,hu2024openrlhf}. AReaL goes
further by decoupling rollout and training asynchronously and explicitly
managing data staleness~\citep{fu2025areal}.

The Qwen million-token capacity workload is narrower: supplied responses and
rewards isolate long-context policy backward, with a separate 192-token online
canary. GLM exercises the sampled-response, reward, and finalized update chain
at exact 2M by time-multiplexing rollout and training on the same 32 H20s.

\section{Conclusion}
\label{sec:conclusion}

LongStraw shows that long-context GRPO under a fixed GPU budget is a tensor-lifetime
and ownership problem, not a context-length race or a single-kernel change.
Long-context processing itself is established when a sufficiently large
accelerator fabric is available. The systems question here is how much
GRPO-shaped execution fits without adding accelerators. LongStraw fixes the
inventory at eight H20 GPUs for Qwen and 32 H20 GPUs for GLM, then makes the
interaction among model structure, prompt-state ownership, suffix replay, parallel
communication, gradient composition, and optimizer ordering explicit. The
shared mechanism is a no-grad prompt boundary followed by serial short-response
replay; the state on that boundary is architecture specific at every layer and
ownership boundary in both model families.

\begingroup
\looseness=-1
For the dense-hybrid Qwen model, compact physical KV pages and recurrent GDN
state make the prompt fit across CP8, and a global LSE/output reduction composes
the full-attention response forward. For GLM, CPU-resident MLA and indexer-key
pages, one-layer staging, complete-layer checkpointing, IndexShare
reconstruction, global DSA, and CP32/EP32 placement carry two 78-layer response
backwards through gradient finalization and one optimizer step. Temporal
resource reuse switches the same 32 H20s between TP8/PP4 rollout and
TP1/CP32/EP32 training. The GLM progression from full-graph failure through
prefix capture, layer-0 replay, all-layer integration, and grouped execution
shows that each dependency class must be resolved in turn.
\par
\endgroup

The durable result is a budget-conditioned, architecture-aware operating
envelope. Within the same eight-H20 envelope, Qwen completes
a 4.25M \(G=8\) response replay and eight consecutive \(G=8\) optimizer steps
comprising 64 member replays at a peak of 83.894~GB per rank. This reuse avoids
repeating a measured 17,729.8-second prefix capture on every cycle. The
4,538,368/4,542,464 train-block bracket exposes further capacity room.
Under the 32-H20 envelope, GLM completes its exact-2M online GRPO workflow end
to end: real policy rollout, mixed rewards, global cross-CP DSA,
two full 78-layer backwards, gradient finalization, and one optimizer step on
all 32 ranks, completing the vLLM-to-Megatron transaction.

Together, these results establish a training-ready path beyond two million
tokens under a fixed GPU budget. The next evaluation scales this working
transaction across repeated updates and task metrics, while the independent
Qwen CP8 replica-finalization check strengthens the companion dense-hybrid path.

\Needspace{1\baselineskip}
\section{Evaluation Scope}
\label{sec:limitations}

The report evaluates two complementary long-context GRPO paths under fixed
hardware. Qwen closes exact response-only execution and optimizer application at
2,097,152 positions for $G=2$ and $G=8$, and reuses one 4,456,448-position
prefix across eight $G=8$ optimizer steps. GLM closes the complete exact-2M
online transaction: a vLLM rollout, reward computation, global DSA, two
78-layer response backwards, distributed gradient finalization, and one
optimizer step on all 32 ranks. The following paragraphs define the objective,
ownership, and measurement fields used by these receipts.

\subsection{Qwen CP8 Gradient-Composition Boundary}

Qwen's response-only receipt composes the full-attention forward over all CP8
page partitions and all-reduces $dQ$. Page owners retain their $dK/dV$
contributions for the replicated K/V projection adapters; the run records the
corresponding local AdamW calls. This owner-composition rule is documented as a
separate CP8 adapter-synchronization audit, while the reported response-only
execution/update closure covers direct response gradients, replay, and prefix
reuse. GLM's exact-2M transaction uses the complementary global path:
cross-CP DSA is composed first, then \texttt{finalize\_model\_grads} runs on
every rank before the optimizer step.

\subsection{Objective and Position Configuration}

Long-context Qwen receipts use stored responses and deterministic rewards for
the declared response-only objective. Prompt state is treated as read-only
within a GRPO update, so each member rebuilds its response graph and releases
that graph after backward. The 4.25M route keeps the captured prefix across
eight explicit update cycles. A separate short-context Qwen integration receipt
exercises sampled responses and DAPO rewards under the same sampled control path
for the recorded online receipt.

The GLM receipt uses real policy sampling from the Tinker-managed vLLM sampler,
ground-truth DAPO reward, and $G=2$ responses with rewards $[-1,+1]$. Rollout
and Megatron replay share the same factor-two YaRN configuration over the
checkpoint's native 1,048,576-position setting, so position semantics remain
aligned through the complete exact-2M transaction.

\subsection{Distributed Update Path}

The two model families expose different ownership contracts. Qwen stores
full-attention KV pages by CP owner and combines response statistics globally;
its K/V-adapter synchronization is kept as a named companion audit. GLM stores
MLA and indexer-key pages on CPU, stages one layer at a time, routes response
rows through EP32, and finalizes the CP/EP-owned gradients before its single
optimizer update. These contracts motivate separate Qwen and GLM receipts
instead of one generic parallel recipe for both model families.

\subsection{Measurement Scope}

All headline measurements use fixed device inventories: eight H20 GPUs for the
Qwen paths and 32 H20 GPUs for the GLM path. Qwen records whole-run allocated
memory and wall time for its 2M and 4.25M receipts. The GLM evidence records the
predecessor capture-window allocation diagnostic and the exact-2M rollout
candidate latencies; the transaction trace records every response backward,
gradient finalization, and optimizer event. Tables retain these phase labels so
the numbers can be reproduced without turning a phase measurement into a
cross-system throughput ranking.

Together, these scopes describe a working, training-ready systems path beyond
two million tokens. The receipts also provide the direct starting point for
repeated-update curves, larger position sweeps, and the companion Qwen adapter
synchronization audit alongside the exact-2M GLM trace.

\begingroup
\renewcommand{\bibfont}{\fontsize{8.5}{10.2}\selectfont}
\bibliographystyle{plainnat}
\bibliography{references}
\endgroup

\begin{appendices}
\section{Model and Run Configuration}
\label{app:configuration}

The two tables below describe each architecture using the inspected model
snapshot and execution run used throughout this report. They separate
architecture facts from run parameters and update-ownership fields. Values are
drawn directly from the final run record; fields outside that record remain
unspecified rather than reconstructed from other notes.

\begin{table}[H]
\centering
\caption{Qwen configuration for the inspected snapshot and fixed eight-H20 runs.}
\label{tab:qwen-config}
\scriptsize
\renewcommand{\arraystretch}{1.08}
\begin{tabularx}{\textwidth}{@{}C{0.28\textwidth}Y@{}}
\toprule
Field & Value \\
\midrule
Decoder & 64 layers; hidden 5,120; intermediate 17,408 \\
Token mixers & 48 GDN; 16 full GQA; 24 query heads, 4 KV heads, head dimension 256 \\
Native position setting & 262,144 \\
Adaptation & NF4 QLoRA, rank 16, alpha 32; 116,727,808 trainable parameters \\
Parallel/storage & 8 H20, CP8, page size 64, compact GPU pages, lazy/pruned detached-prefix gradient pages \\
Prompt/response & 2,088,960 prompt tokens; 8,192 response input tokens per member; at most 8,191 scored \\
Replay blocks & 510 prompt chunks of 4,096; four response blocks of 2,048; stage-1 MLP microblock 4,096; backward MLP microblock 512 \\
Objective in run & Synthetic rewards; old/reference from same runner; \(\beta=0\); live ratio term unclipped \\
Update boundary & Response-only execution: global response forward, reduced \(dQ\), and local AdamW calls; page-owner K/V adapter synchronization is tracked in the companion CP8 audit \\
\bottomrule
\end{tabularx}
\end{table}

\begin{table}[!htbp]
\centering
\caption{GLM configuration for the exact-2M online transaction and inspected runtime.}
\label{tab:glm-config}
\scriptsize
\begin{tabularx}{\textwidth}{@{}C{0.28\textwidth}Y@{}}
\toprule
Field & Value \\
\midrule
Decoder & 78 layers; hidden 6,144; 64 MLA query heads; query/KV latent widths 2,048/512 \\
Sparse attention & 32 indexer heads of dimension 128; top-2,048; 21 compute layers and 57 IndexShare consumers \\
Feed-forward & First 3 dense; remaining 75 MoE; 256 routed experts, top-8, one shared expert; expert intermediate 2,048 \\
Position setting & Native 1,048,576; opt-in YaRN factor 2.000015 with \texttt{max\_position\_embeddings=2097168} for rollout and training \\
Adaptation & LoRA rank 8; attention projection, dense/routed/shared FFN, and output-head target families \\
Frozen objects & Base weights, embeddings, norms, router parameters, and DSA indexer parameters; router expert-bias transition not measured \\
Parallel/storage & 32 H20 reused by phase; rollout TP8/PP4; training TP1/CP32/EP32/ETP1/PP1; CPU prefix pages; page size 64 \\
Prompt/response & Exactly 2,097,152 prompt tokens; two five-token completions, each with four scored log-probabilities \\
Objective in run & Rewards and advantages \([-1,+1]\); \(\epsilon=0.2\), \(\beta=0\); old snapshot is reference; one lower and one upper clip event \\
Update boundary & Global cross-CP DSA; two 78-layer response-only backwards; \texttt{finalize\_model\_grads}; one optimizer step on all 32 ranks \\
\bottomrule
\end{tabularx}
\end{table}

The run artifact verifies its packaged files and records the phase fields listed
in Table~\ref{tab:glm-config}. The table keeps candidate timing and capture
memory distinct from the full rank trace that closes the exact-2M transaction.

\section{GLM Page Mapping and State-Size Derivation}
\label{app:glm-pages}

Megatron zigzag context parallelism divides the prompt into \(2C\) contiguous
chunks for \(C=32\). With 32,768 global pages and page size 64, each chunk has
512 pages. Rank \(r\) owns chunks \(r\) and \(63-r\):
\begin{equation}
\mathcal{P}_r=
\{512r,\ldots,512r+511\}\cup
\{512(63-r),\ldots,512(63-r)+511\}.
\label{eq:glm-zigzag-pages}
\end{equation}
Each rank owns 1,024 pages (65,536 prompt tokens); all 32 ranks have logged
endpoint samples.

The page tensor shapes come from the live replay implementation and a prior
shape trace. An MLA page is BF16 \([1,64,1,576]\), where 576 combines the latent and
rotary content used by the absorbed path. A DSA indexer-key page is BF16
\([1,64,128]\). Materializing a complete local layer produces
\([1,65{,}536,1,576]\); an index-computing layer also materializes
\([1,65{,}536,128]\).

The following numbers are derived from those shapes, not measured memory
peaks. One local MLA layer occupies 72~MiB. One local DSA index-key set occupies
16~MiB. Across 78 MLA states and 21 index-key states, CPU prefix storage is
\begin{equation}
78\times72~\mathrm{MiB}+21\times16~\mathrm{MiB}
=5{,}952~\mathrm{MiB}=5.8125~\mathrm{GiB/rank}.
\end{equation}
The collective CPU payload is 186~GiB. One staged shared-index layer needs a
72~MiB prompt payload before response work; one index-computing layer needs
88~MiB. These figures exclude page metadata, pinned transfer buffers, model
weights, response tensors, and allocator overhead.

\section{Representative GLM Trace Formats}
\label{app:trace-contract}

The predecessor policy trace begins with response hidden shape \([2,1,6144]\). A compute
layer records both state components, 1,024 pages, local prefix 65,536, total
local length 65,538, top-k shape \([1,2,2048]\), CPU offload, whole-layer
checkpointing, and backend \texttt{runtime\_unfused\_absorbed}. An IndexShare
consumer records MLA state only and consumes the per-forward selection holder
published by its producer.

Output logits have shape \([1,2,154880]\). Backward runs from layer 77 to 0.
Layer and attention-projection gradients are \([2,1,6144]\); sparse-attention
gradients are \([2,1,16384]\). Policy traces contain 396 JSONL events; old traces
contain 160. These are host events, not kernel profiles or numerical gradient
dumps.

In the current exact-2M run, gradients are disabled for 1,394 parameters during
capture and 99 state hooks
cover 78 MLA states plus 21 DSA index-key states. One shared RoPE cache replaces
98 repeated copies. Complete
decoder layers use reentrant checkpointing with RNG preservation; saved sparse
tensors may be moved to CPU. The exact-2M audit contains 64 rank/group files
and 25,536 events, including 3,648 global-position DSA events and 1,344 fused
global-indexer events. It records both backwards and gradient finalization on
all ranks without non-finite values or execution errors.

\section{Distributed-Gradient Validation}
\label{app:gradient-audit}

\paragraph{Qwen.}
The distributed attention backward all-reduces \(dQ\) and returns page-owner
\(dK/dV\). The latter are correct gradients of sharded KV tensors, but their
projection adapters are replicated. The probe creates one AdamW instance per
rank without DDP or a parameter-gradient reducer. The required selective K/V
and upstream-hidden composition is absent from the historical path.

\paragraph{GLM.}
The current group loop executes two backward calls, then invokes
\mbox{\texttt{finalize\_model\_grads}}, the optimizer step, and gradient clear in
that order. All 32 ranks record this sequence at exact 2M. This closes the
missing-finalization defect in the predecessor run.

\section{Detailed GLM Capacity Progression}
\label{app:glm-progression}

The conventional 2,097,152-position full-sequence attempt exposed several independent peaks.
The DSA score scratch \([8192,2{,}097{,}152]\) in FP32 is approximately
64~GiB. After restricting that path, expert-LoRA scale/add work reached
7.80~GiB and an FC2 matmul allocation reached 11.70~GiB. Smaller MoE chunks
moved the failure into expert-output concatenation; at chunk size 65,536 the
requested concatenate allocation was 19.37~GiB. These failures show why one
kernel optimization did not resolve the full graph.

The corresponding capacity milestones were:
\begin{enumerate}[leftmargin=2.0em]
  \item all-layer no-grad prompt capture at 128K, 256K, 512K, 1M, and 2.097M;
  the final 2,097,152-position prefix-only capture took 675.657~s;
  \item layer-0 2.097M capture, two local backwards, and an optimizer-call test in
  738.579~s;
  \item all-layer CP32 tests at 32K and 64K after IndexShare, CPU page, shared
  RoPE, and checkpoint fixes;
  \item a 2.097M \(G=1\) all-78-layer test in 2042.975~s; and
  \item the predecessor \(G=2\) rank-complete transaction in 2975.138~s;
  \item 32K/64K online integration canaries; and
  \item the current exact-2M policy rollout and finalized \(G=2\) update.
\end{enumerate}
The sequence is monotonic in execution coverage, not in elapsed time. The
CP-local DSA and skipped-finalization boundaries apply only to the predecessor
milestones; the current exact-2M transaction observes the global operator and
gradient finalization.

\section{Qwen 4.25M Replay within Eight H20s}
\label{app:qwen-frontier}

The Qwen investigation also reached a 4.25M context frontier, where
4.25M means \(4.25\times 2^{20}=4,456,448\) exact positions. The resident
response-replay run captures 4,448,256 prompt positions once, then completes
all eight old/reference/policy branches for 8,192 response positions. Every
policy branch finishes four 2,048-position backward blocks. The group takes
4052.920~s after a 17697.213~s inferred prefix interval, or 21750.133~s from
prefix start through \(G=8\), at 82.960~GB per rank. Its old resident command
interface intentionally skips the optimizer, so it is a complete replay run
rather than an optimizer-step run.

\begin{table}[!htbp]
\centering
\caption{Qwen 4.25M replay within eight H20s (4,456,448 exact positions). Train-block probes are capacity checks. The
resident row completes every old/reference/policy branch and all four backward
blocks for each of eight members, but its old command interface skips the
optimizer. The resident prefix-reuse row completes eight \(G=8\)
optimizer steps (64 member replays) from one captured cache. Completed-run peaks
use PyTorch allocator bytes in decimal GB; the \(^*\) approximate
full-run OOM value is sampled process memory and is not directly comparable.
The companion prefix-comparison experiment is reported separately.}
\label{tab:frontier-results}
\footnotesize
\begin{tabular}{cccc}
\toprule
Context length & Run type & Result & Reported GB \\
\midrule
\rowcolor{gray!4}
4,194,304 & train-block proxy & fits & 136.717 \\
\rowcolor{gray!4}
4,456,448 & train-block proxy & fits & 143.163 \\
\rowcolor{gray!4}
4,538,368 & train-block proxy & fits & 145.176 \\
\rowcolor{red!6}
4,542,464 & train-block proxy & OOM & \(\approx 145.277\) \\
\rowcolor{red!6}
4,456,448 & unpruned full run & OOM in policy backward & \(\approx 142.293^*\) \\
\rowcolor{yellow!10}
4,456,448 & resident response replay, \(G=8\) & completed, all 8 members & 82.960 \\
\rowcolor{yellow!10}
4,456,448 & resident prefix reuse, \(G=8\) & completed, 8 optimizer steps & 83.894 \\
\rowcolor{yellow!10}
4,456,448 & batched old/reference, \(G=8\) & completed, small speedup & 135.128 \\
\bottomrule
\end{tabular}
\end{table}

A separate resident prefix-reuse run supplies the multi-step measurements.
It completes eight \(G=8\) accumulation cycles and eight optimizer steps: 64
member replays in total. Every rank records
\texttt{optimizer\_step\_applied=true}, \texttt{prefix\_stale=false}, and
\texttt{prefix\_frozen\_response\_only=true}; peak allocation rises from
82.960~GB on the first cycle to 83.894~GB thereafter. These fields record the
runner's intentional permission to reuse the cache; they do not disable the
adapters or make a pre-update cache parameter-invariant. The companion 1M
prefix-comparison experiment is documented in Section~\ref{tr:analysis}.

Before detached-prefix gradient-page pruning, a clean prompt-adapted run
completed prefix capture, old/reference scoring, and policy stage-1 forward,
then OOMed in response backward. The pruning is exact for LongStraw's detached-prefix
objective because prompt K/V pages still participate in attention and \(dQ\),
while their own unused \(dK/dV\) storage is omitted from the live graph.

The linear storage slope follows directly from the Qwen model structure. For
16 full-attention layers with four KV heads of dimension 256 in BF16, sharded
over CP8, one million additional global context positions add
\(10^6\mathbin{\times}16\mathbin{\times}2\mathbin{\times}4
\mathbin{\times}256\mathbin{\times}2/8=8.192\) decimal GB of KV storage per
rank. This arithmetic explains the above-4M capacity potential; replay scratch,
score buffers, and autograd state determine the complete-path headroom.

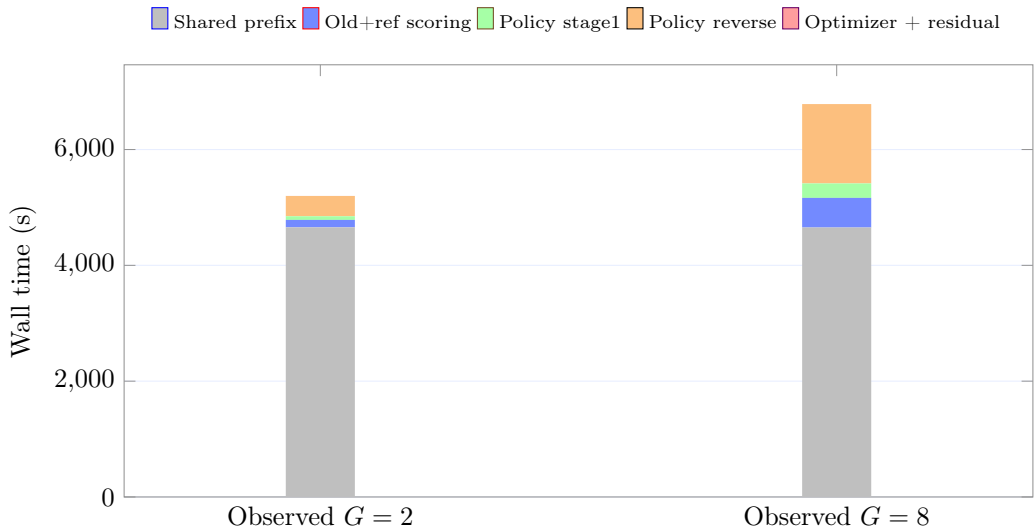
\begin{figure}[H]
\centering
\begin{tikzpicture}
\begin{axis}[
    ybar stacked,
    width=0.82\textwidth,
    height=7.299cm,
    ymin=0,
    ylabel={Wall time (s)},
    symbolic x coords={Observed \(G=2\),Observed \(G=8\)},
    xtick=data,
    bar width=26pt,
    enlarge x limits=0.38,
    legend style={at={(0.5,1.05)}, anchor=south, legend columns=5, draw=none, font=\scriptsize},
    ymajorgrids=true,
    grid style={draw=mindlabgrid},
    axis line style={draw=mindlabfg!45},
    tick style={draw=mindlabfg!45},
    every axis plot/.append style={draw=none}
]
\addplot+[fill=gray!50] coordinates {(Observed \(G=2\),4656.225) (Observed \(G=8\),4653.420)};
\addplot+[fill=mindlabblue!55] coordinates {(Observed \(G=2\),128.608) (Observed \(G=8\),511.800)};
\addplot+[fill=green!35] coordinates {(Observed \(G=2\),62.671) (Observed \(G=8\),249.600)};
\addplot+[fill=BurntOrange!55] coordinates {(Observed \(G=2\),350.800) (Observed \(G=8\),1370.224)};
\addplot+[fill=red!38] coordinates {(Observed \(G=2\),0.476) (Observed \(G=8\),0.181)};
\legend{Shared prefix, Old+ref scoring, Policy stage1, Policy reverse, Optimizer + residual}
\end{axis}
\end{tikzpicture}
\caption{\textbf{Qwen group accounting within eight H20s.} The \(G=2\) and
\(G=8\) bars are serial-loop anchors, not group-size limits. Four times as many
members add 1,586.445~s but only 0.208~GB (0.213\%) at peak because the shared
prefix dominates residency; the final segment is a phase-sum residual, not a
pure optimizer timer.}
\label{fig:time-breakdown}
\end{figure}

Batched old/reference scoring is a useful negative result. At 4K it reduces a
one-GPU \(G=8\) total from 17.529 to 14.528 seconds. At 4,456,448 context it changes the
post-prefix time from 4051.240 to 4024.290 seconds, only 0.7\%, while peak
memory rises from 83.894 to 135.128~GB. Window/sparse attention, a 4,096-token
response block, metadata batching, an alternative GDN backend, and a naive
query split remain diagnostic or rejected variants rather than primary
results in this report.

\begin{figure}[H]
\centering
\begin{tikzpicture}
\begin{axis}[
    width=0.92\textwidth,
    height=9.8cm,
    xlabel={Context positions (millions)},
    ylabel={Peak memory per rank (GB)},
    xmin=1.8, xmax=4.7,
    ymin=50, ymax=154,
    xtick={2.097,2.5,3.0,3.5,4.0,4.5},
    ymajorgrids=true,
    xmajorgrids=false,
    grid style={draw=mindlabgrid},
    axis line style={draw=mindlabfg!45},
    tick style={draw=mindlabfg!45},
    legend style={
        at={(0.5,1.035)},
        anchor=south,
        legend columns=3,
        draw=none,
        font=\footnotesize,
        cells={anchor=west},
        /tikz/every even column/.append style={column sep=5pt}
    }
]
\addplot+[only marks, mark=*, mark size=2.7pt, color=mindlabblue] coordinates {(2.088960,58.655)};
\addlegendentry{Prefix-only pass}
\addplot+[only marks, mark=square*, mark size=2.7pt, color=green!45!black] coordinates {(2.097152,97.503)};
\addlegendentry{Conditional-response run}
\addplot+[
    only marks,
    mark=triangle*,
    mark size=4.0pt,
    color=mindlabfg!75,
    mark options={fill=white, line width=0.75pt, xshift=-1.8pt}
] coordinates {(4.194304,136.717) (4.456448,143.163) (4.538368,145.176)};
\addlegendentry{Train-block proxy pass}
\addplot+[
    only marks,
    mark=x,
    mark size=4.0pt,
    color=red!70!black,
    line width=1.1pt,
    mark options={xshift=1.8pt}
] coordinates {(4.542464,145.277)};
\addlegendentry{Train-block OOM}
\addplot+[only marks, mark=diamond*, mark size=3.0pt, color=BurntOrange!80!black] coordinates {(4.456448,82.960) (4.456448,83.894)};
\addlegendentry{Replay / 8-step pass}
\draw[dashed, mindlabfg!50] (axis cs:2.0,150.755) -- (axis cs:4.65,150.755);
\node[
    anchor=north west,
    font=\footnotesize,
    text=mindlabfg!75,
    fill=white,
    inner sep=1.5pt
] at (axis cs:2.03,149.4) {H20-3e: 143,771 MiB = 140.401 GiB = 150.755 GB};

\node[
    anchor=north east,
    align=left,
    font=\footnotesize,
    fill=white,
    draw=mindlabfg!22,
    rounded corners=1pt,
    inner sep=2.5pt
] (pair425) at (axis cs:3.73,122) {%
    \textcolor{mindlabfg!75}{\(\blacktriangle\)} 4,456,448 proxy fits};
\draw[mindlabfg!65, line width=0.5pt]
    (pair425.north east) to[out=24,in=205] (axis cs:4.456448,143.163);

\node[
    anchor=north east,
    align=left,
    font=\footnotesize,
    fill=white,
    draw=mindlabfg!22,
    rounded corners=1pt,
    inner sep=2.5pt
] (pair454) at (axis cs:4.63,122) {%
    \textcolor{mindlabfg!75}{\(\blacktriangle\)} 4,538,368 proxy fits\\[-1pt]
    \textcolor{red!70!black}{\(\times\)} 4,542,464 proxy OOM};
\draw[mindlabfg!65, line width=0.5pt]
    ([xshift=-5pt]pair454.north east) to[out=82,in=275] (axis cs:4.538368,145.176);
\draw[red!70!black, line width=0.5pt]
    (pair454.north east) to[out=87,in=280] (axis cs:4.542464,145.277);
\end{axis}
\end{tikzpicture}
\caption{\textbf{Qwen 4.25M replay within eight H20s.} Here 4.25M denotes
4,456,448 exact positions. At this scale, the resident \(G=8\) path completes
all response replays at 82.960~GB, and the resident prefix-reuse variant
completes eight optimizer steps at 83.894~GB. A train-block proxy reaches
4,538,368 and OOMs one 4,096-position chunk later at 4,542,464. The clean
prompt-adapted path before detached-prefix gradient-page pruning OOMs in policy
backward at 4,456,448; it remains in
Table~\ref{tab:frontier-results}, but its sampled process-memory reading is omitted
here because it is not comparable with allocator peaks. The 8-step result
validates cross-step reuse and its efficiency; the companion prefix-comparison
experiment is documented in Section~\ref{tr:analysis}.}
\label{fig:memory-frontier}
\end{figure}
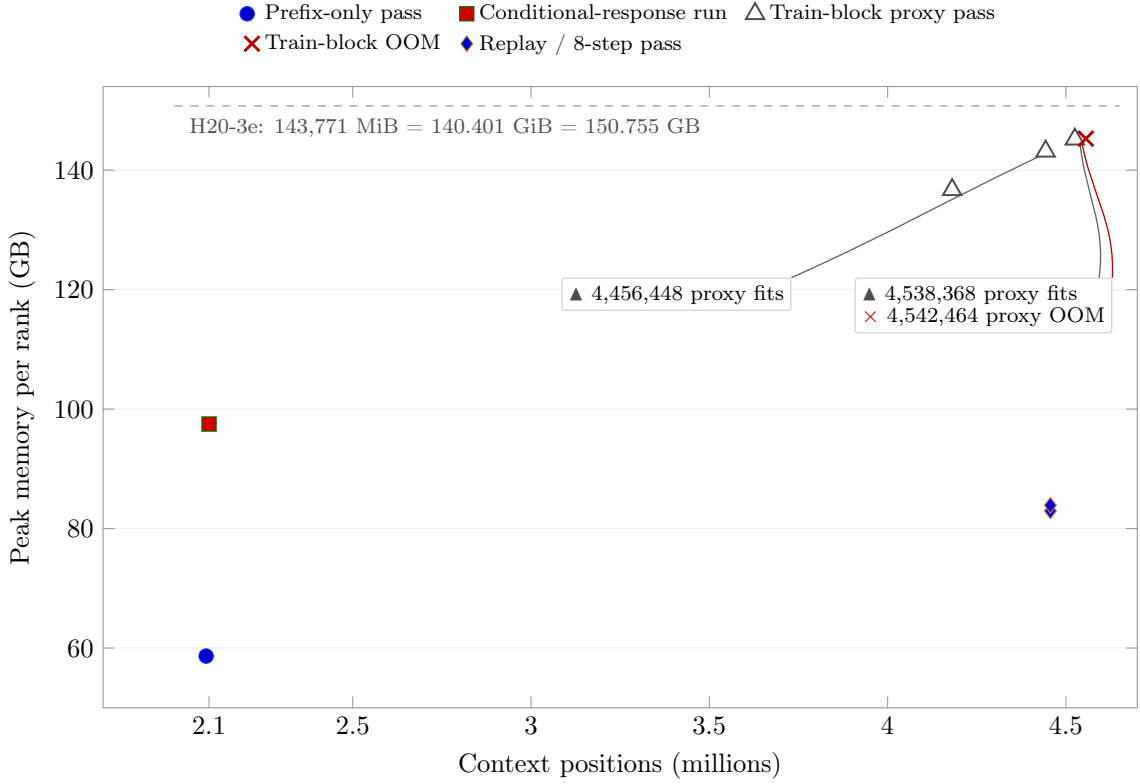

Figure~\ref{fig:memory-frontier} separates prefix-only capacity, train-block
diagnostics, and complete replay or optimizer-step runs. Its line connects
only the local train-block bracket; it is not a fitted memory law. The
low-memory 4.25M points use the detached-prefix objectives described above,
whereas the high-memory points retain the prompt-adapted block path.

Read vertically, the prefix-only and conditional-response points establish
different execution scopes; their lower memory does not imply that a
prompt-adapted training graph fits. The triangle sequence is a train-block
diagnostic whose final pass/OOM pair differs by one 4,096-token chunk. The
replay diamonds represent completed suffix work, while the conditional point
records a response path without an optimizer step.

\Needspace{2\baselineskip}
No line
connects the completed replay points because there is no matched sweep over
context length, objective, and measurement window. The allocator readings are
also not directly comparable when their collection windows differ;
Table~\ref{tab:frontier-results} remains the scope key for every plotted point.

\section{Acknowledgements}
\label{app:acknowledgements}

\begingroup
\footnotesize
We thank the following MindLab members for their support and contributions to
the broader research environment: Theo Li, Song Cao, Wenbin Wang, Fancy Kong,
Regis Ye, Charles Huang, Murphy Zhuang, Josh Ying, Anya Zhang, Alyssa, Ray Li,
Logan Liu, Xiang Liu, Yuhan Zhan, Kaixuan Fan, Mutian Hong, Zhuoran Shen,
Hua Jiang, Wenxi Qu, Yuxin Lu, Neo Liu, Hera Feng, Aaron Guan,
Fan Lin, Guoshuai Han, Xinyue Zhu, Chengdong Xu, Jingwei Cao, Smith Li, Kun Li,
Jianbo Wu, Yuyi Jiang, Sueky Zhang, Kairus Liu, Zhihui Li, Wei
Zhao, Anson Qiu, Hongquan Gu, Peixuan Hua, Nora Jiang, Ada Zhou, Qiuyu Jin,
Ruijia Zhang, Arthur Fu, Maxwell Yao, Jiayi Lin, Runze Lv, Hailee Hou, Miles
Jiang, Ya Zhang, Danney Zeng, Vin Bo, and Jason Zhang. We also thank the NVIDIA
team for its support.
\endgroup

\end{appendices}

\end{document}